\documentclass[pdflatex,sn-mathphys]{sn-jnl}

\jyear{2026}%
\usepackage[table]{xcolor}
\usepackage{xurl}
\usepackage[globalcitecopy]{bibunits}
\usepackage{hyperref}

\hypersetup{
    colorlinks=True,
    citecolor=blue,
    linkcolor=blue,
    urlcolor=blue,
    allcolors=blue
}

\makeatletter
\AtBeginDocument{\def\theHALG@line{\thealgorithm.\arabic{ALG@line}}}
\makeatother

\usepackage[nameinlink]{cleveref}
\crefname{figure}{Fig.}{Figs.}
\Crefname{figure}{Figure}{Figures}

\usepackage{xr-hyper}
\theoremstyle{thmstyleone}%

\theoremstyle{thmstyletwo}%

\theoremstyle{thmstylethree}%

\usepackage{subcaption}
\usepackage[section]{placeins}
\usepackage{tabularx}
\usepackage{makecell}
\usepackage{multirow}
\usepackage{float}

\usepackage{CJKutf8}
\usepackage{csquotes}

\usepackage{amsmath}
\usepackage{amssymb}
\usepackage{bm}
\usepackage{longtable}
\usepackage{totcount}
\usepackage{booktabs}

\usepackage{fancyhdr}

\usepackage{etoolbox}
\usepackage{xpatch}

\makeatletter
\patchcmd{\ps@headings}
  {\hbox to \hsize{\hfill Springer Nature 2021 \LaTeX\ template\hfill}}
  {\hbox to \hsize{}}{}{}
\patchcmd{\ps@titlepage}
  {\hbox to \hsize{\hfill Springer Nature 2021 \LaTeX\ template\hfill}}
  {\hbox to \hsize{}}{}{}
\xpatchcmd*{\shipout}{Springer Nature 2021 \LaTeX\ template}{}{}{}
\xpatchcmd*{\@outputpage}{Springer Nature 2021 \LaTeX\ template}{}{}{}
\xpatchcmd*{\ps@headings}{Springer Nature 2021 \LaTeX\ template}{}{}{}
\xpatchcmd*{\ps@myheadings}{Springer Nature 2021 \LaTeX\ template}{}{}{}
\makeatother

\regtotcounter{section}
\regtotcounter{table}
\regtotcounter{figure}
\begin{document}
\makeatletter
\immediate\write\@auxout{\string\bibdata{main}}
\makeatother
\addtocontents{toc}{\protect\setcounter{tocdepth}{-1}}
\makeatletter
\def\@extra@binfo{@main}
\def\@extra@b@citeb{@main}
\makeatother
\begin{bibunit}[sn-mathphys]

\title{3DTCR: Vortex-following generative reconstruction improves tropical cyclone intensity prediction from global AI weather forecasts}

\pagestyle{plain}

\author[1]{\fnm{Jun} \sur{Liu}}\email{junliu23@m.fudan.edu.cn}
\author*[1]{\fnm{Xiaohui} \sur{Zhong}}\email{zhong\_xiaohui@fudan.edu.cn}

\author[3]{\fnm{Kai} \sur{Zheng}}\email{kaizheng.work@gmail.com}

\author[1]{\fnm{Jiarui} \sur{Li}}\email{jrli24@m.fudan.edu.cn}

\author[1]{\fnm{Yifei} \sur{Li}}\email{liyifei24@m.fudan.edu.cn}

\author[2]{\fnm{Tao} \sur{Zhou}}\email{tzhou22@m.fudan.edu.cn}
\author[1]{\fnm{Wenwen} \sur{Zhang}}\email{wwzhang26@m.fudan.edu.cn}

\author*[1,3]{\fnm{Hao} \sur{Li}}\email{lihao\_lh@fudan.edu.cn}

\affil[1]{\orgdiv{Artificial Intelligence Innovation and Incubation Institute}, \orgname{Fudan University}, \orgaddress{\city{Shanghai}, \postcode{200433}, \country{China}}}

\affil[2]{\orgdiv{Department of Atmospheric and Oceanic Sciences, Institute of Atmospheric Sciences}, \orgname{Fudan University}, \orgaddress{\city{Shanghai}, \postcode{200438}, \country{China}}}

\affil[3]{\orgname{Shanghai Academy of Artificial Intelligence for Science}, \orgaddress{\city{Shanghai}, \postcode{200232}, \country{China}}}

\abstract{Global numerical weather prediction (NWP) and artificial intelligence (AI) models often underestimate tropical cyclone (TC) intensity, partly because coarse grids underresolve fine-scale three-dimensional inner-core structures that influence TC intensity evolution. Kilometer-scale NWP simulations can better resolve these structures but are too costly for operational use, which also makes high-resolution TC data limited. Here we construct a 3-km, TC-centered dataset for 2018--2024 using the Weather Research and Forecasting (WRF) model and develop 3DTCR, a vortex-following generative framework that reconstructs three-dimensional TC structure from FuXi global forecasts. 3DTCR combines conditional flow matching (CFM) with two-stage training to learn a coarse-to-fine mapping from center-aligned reanalysis simulation pairs and then adapt it to forecast inputs through latent space alignment. Evaluated on 2024 western North Pacific (WNP) forecasts, 3DTCR reduces the root mean square error (RMSE) and the magnitude of the mean bias error (MBE) in maximum 10-m wind speed by 36.5\% and 60.7\% relative to FuXi, and by 8.9\% and 28.2\% relative to the European Centre for Medium-Range Weather Forecasts high-resolution system (ECMWF-HRES), respectively. Case studies suggest that reconstructions capture wind gradients between the eye and eyewall, vertical structure of horizontal winds, and vortex asymmetries, while inheriting the FuXi-predicted TC tracks. 3DTCR generates 5-day TC-centered atmospheric fields in approximately 3 minutes on one GPU, whereas a full-basin 3-km WRF simulation takes about 400 hours on 32 CPU cores. These results demonstrate that such a vortex-following generative reconstruction can add localized three-dimensional detail to global forecasts and improve TC intensity prediction without basin-wide kilometer-scale numerical integration.
}



\maketitle
\pagestyle{plain}

\section{Introduction}
\label{sec:intro}

Tropical cyclones (TCs) are among the most destructive weather hazards, making reliable TC forecasts essential for disaster preparedness and risk mitigation \cite{zhang2009tropical,mendelsohn2012impact,peduzzi2012global,kossin2020global,radfar2026synergistic}.
Global numerical weather prediction (NWP) and artificial intelligence (AI) systems now provide skillful medium-range forecasts of large-scale atmospheric evolution \cite{pathak2022fourcastnet,bi2023accurate,chen2023fuxi,lam2023learning,kochkov2024neural,shi2025comparison}, yet coarse global systems in both paradigms frequently underestimate TC intensity \cite{emanuel2018100,demaria2024evaluation,zhang2026physics}.
TC intensity depends on interactions between the large-scale environment and compact inner-core processes, including sharp radial wind gradients, eyewall organization, moisture transport, and secondary circulations. The three-dimensional organization of these processes regulates momentum and moisture transport through the vortex and helps determine whether and how rapidly a TC strengthens or weakens \cite{li2023recent,roberts2020impact,emanuel2017role}. However, these features are smoothed in the coarse-resolution reanalyses and forecasts \cite{dulac2024assessing}. Consequently, improving intensity prediction may require information beyond a corrected scalar maximum wind speed, including the evolving three-dimensional vortex structure associated with intensification and weakening \cite{chen2025meteorologically,bhatia2022potential}.

Existing approaches trade structural fidelity against computational efficiency.
Sequence-based AI models combine historical best-track records with environmental fields to improve scalar intensity forecasts \cite{huang2025benchmark,wang2025global}, but they do not explicitly represent the three-dimensional vortex structure or the physical processes governing intensity change \cite{wang2025advancing,bhatia2022potential,kossin2020global}.
High-resolution NWP and moving nested grids can resolve eyewall dynamics and secondary circulations \cite{roberts2020impact}, but kilometer-scale integrations remain too computationally expensive for routine operational forecasting and large ensembles \cite{haghroosta2014efficiency,moon2021does,lean2024hectometric}. 
This cost also limits the availability of high-resolution TC datasets from which AI models can learn inner-core structure and multilevel physical organization.
Global AI-based weather forecasting models are significantly less expensive to run, but they are commonly trained on coarse reanalyses such as ERA5, in which sharp inner-core gradients and extreme winds are already attenuated \cite{dulac2024assessing,bloemendaal2019global}.
These trade-offs motivate a hybrid strategy that retains the large-scale guidance and efficiency of AI global forecasts while reconstructing high-resolution fields only around the moving TC.

In recent years, AI-based downscaling has shown promise for recovering fine-scale TC wind, rainfall, and multivariable fields from coarse inputs \cite{lopez2025dynamical,chen2024deep,vosper2023deep,jing2024tc,niu2026intelligent,bano2025regional}, but several challenges limit its application to intensity forecasting.
First, fixed-domain methods are sensitive to track error \cite{mardani2025residual}.
When the predicted and observed TCs are displaced, pixel-wise losses penalize a structurally realistic but displaced vortex at both locations.
This \enquote{double penalty} favors spatially averaged, overly smooth reconstructions \cite{subich2025fixing,wang2022customized}.
Second, many TC applications reconstruct individual surface variables, such as near-surface wind or rainfall \cite{sun2024key,chen2025constructing}, whereas intensity evolution depends on coupled, multilevel kinematic and thermodynamic structure.
Third, extreme inner-core winds form only a small fraction of global or basin-scale data, so deterministic models trained with pixel-wise losses can be dominated by weak and moderate conditions and suppress rare extremes \cite{sun2025can,subich2025fixing}.
Probabilistic generative models can restore variability and fine-scale detail \cite{jing2024tc,zhong2024fuxi,lockwood2024generative,niu2026intelligent,price2025probabilistic}, but iterative diffusion sampling can be computationally costly \cite{kornilov2024optimal,lipman2022flow,esser2024scaling,fotiadis2025adaptive}.
Flow matching offers a more direct generative trajectory and a promising alternative for efficient high-resolution reconstruction \cite{lipman2022flow,kornilov2024optimal,fotiadis2025adaptive,couairon2026archesweathergen}.
The remaining challenge is to combine this efficiency with a TC-following, multilevel representation that is robust to displacement and forecast domain shifts while preserving extreme inner-core structure.

Here, we construct a 3-km, TC-centered dataset spanning 2018--2024 with the Weather Research and Forecasting (WRF) model and develop 3DTCR, a vortex-following generative framework that reconstructs three-dimensional TC structure from global AI forecasts.
The dataset consists of independent $10^\circ \times 10^\circ$ simulations centered on historical TC best-track positions, with simulations from 2018--2023 used for training and 2024 reserved for independent evaluation.
Each sample contains 21 surface and pressure level fields that describe the kinematic and thermodynamic structure of a vortex.
3DTCR couples TC tracking with conditional flow matching (CFM), defining a moving reconstruction domain around the position provided by the driving global forecast rather than a fixed geographical domain.
Training proceeds in two stages.
Pretraining learns a coarse-to-fine mapping from spatially aligned ERA5-WRF pairs, and supervised fine-tuning (SFT) adapts this mapping to FuXi forecasts through center-aligned sampling, lead time conditioning, and latent space alignment.
This two-stage strategy is designed to reduce spatial pattern and domain shifts between the forecasts and high-resolution targets while recovering intensity relevant inner-core structures that are smoothed at coarse resolution.
As 3DTCR inherits the vortex position from the driving forecast, the generative model can focus on reconstructing structure and intensity instead of simultaneously correcting track displacement and recovering fine-scale detail.

We evaluate 3DTCR on 2242 forecasts from the independent 2024 western North Pacific (WNP) testing dataset at lead times of 6--120 h, using the International Best Track Archive for Climate Stewardship (IBTrACS) as the reference \cite{knapp2010international}.
Across these lead times, 3DTCR reduces the root mean square error (RMSE) and the magnitude of the mean bias error (MBE) in maximum 10-m wind speed by 36.5\% and 60.7\% relative to FuXi, and by 8.9\% and 28.2\% compared with the high-resolution forecasts from the European Centre for Medium-Range Weather Forecasts (ECMWF-HRES).
Case studies suggest that 3DTCR can recover key features, such as wind gradients between the eye and the eyewall, vertical wind structure, and asymmetries, while the reconstructed fields inherit the TC tracks from FuXi.
3DTCR generates 5-day TC-centered fields in approximately 3 minutes on a single NVIDIA A100 GPU, whereas a full-basin 3-km WRF simulation requires about 400 hours on 32 CPU cores.
These results indicate that vortex-following generative reconstruction can efficiently bridge global atmospheric forecasts and localized, high-resolution vortex structure and is associated with improved TC intensity estimates without basin-wide kilometer-scale integration.


\section{Results}
\label{sec:results}

We evaluate 3DTCR from five perspectives.
We first assess the suitability of the TC-centered WRF 3-km dataset as a reconstruction target.
We then determine whether 3DTCR improves forecasts of maximum 10-m wind speed (WS10M) relative to FuXi and ECMWF-HRES without changing the track provided by the driving forecasts.
Case studies and spectral analyses examine whether these improvements reflect coherent three-dimensional, multiscale vortex structure rather than uniform amplification of the wind field.
Finally, ablation studies and runtime comparisons quantify the contributions of the training strategy and the computational costs of TC-centered reconstruction.

Unless otherwise stated, quantitative evaluations are based on the independent 2024 WNP testing dataset.
IBTrACS provides the reference for TC intensity and position, whereas the WRF 3-km simulations serve as the high-resolution reference for spatial structure.
Table~\ref{tab:method_comparison} summarizes the model variants.
The 3DTCR-Pre variants use weights pretrained on ERA5-WRF pairs and take either ERA5 or FuXi as input at inference, without forecast domain adaptation.
The supervised fine-tuning (SFT) variants initialize from these weights and are fine-tuned either on FuXi alone, without maximum mean discrepancy (MMD), or on ERA5 and FuXi with MMD-based latent space alignment.
The end-to-end (E2E) variant uses the same training data and objective but starts from random initialization, thereby isolating the contribution of pretraining.
3DTCR denotes the full 3DTCR-SFT (FuXi, w/ MMD) model.
Table~\ref{tab:method_comparison}, Section~\ref{Materials_and_methods}, and Supplementary Section~\ref{app:ablation_framework} provide complete definitions of the variants and their experimental roles.

\begin{table}[!htp]
\centering
\caption{\textbf{Configurations and experimental roles of the 3DTCR variants.} All variants use the same reconstruction backbone and WRF 3-km target. The data source in parentheses denotes the input used at inference. SFT denotes supervised fine-tuning, and MMD denotes maximum mean discrepancy.}
\label{tab:method_comparison}
\scriptsize
\renewcommand{\arraystretch}{1.12}
\setlength{\tabcolsep}{1.5pt}
\begin{tabularx}{\textwidth}{@{}lc>{\centering\arraybackslash}p{2.2cm}>{\raggedright\arraybackslash}X@{}}
\toprule
\textbf{Method} & \textbf{Strategy} & \textbf{Loss function} & \textbf{Description} \\
\midrule
\multicolumn{4}{l}{\textbf{Baselines}} \\
\midrule
FuXi & - & - & Global AI weather forecast \\
\addlinespace[1pt]
ERA5 & - & - & Reanalysis data \\
\addlinespace[1pt]
ECMWF-HRES & - & - & Operational NWP forecast \\
\midrule
\multicolumn{4}{l}{\textbf{Ours}} \\
\midrule
3DTCR-Pre (ERA5) & Pretraining & $\mathrm{L}_{\text{RF}}$ & Pretrained on ERA5, evaluated with ERA5 input \\
\addlinespace[2pt]
3DTCR-Pre (FuXi) & Pretraining & $\mathrm{L}_{\text{RF}}$ & Pretrained on ERA5, evaluated with FuXi input without fine-tuning \\
3DTCR-E2E (FuXi, w/ MMD) & One-stage & $\mathrm{L}_{\text{RF}}+\lambda\mathrm{L}_{\text{MMD}}$ & Trained from random initialization on ERA5 and FuXi with MMD \\
\addlinespace[2pt]
3DTCR-SFT (FuXi, w/o MMD) & SFT & $\mathrm{L}_{\text{RF}}$ & Pretrained on ERA5, then fine-tuned on FuXi without MMD \\

\addlinespace[2pt]
3DTCR-SFT (FuXi, w/ MMD) & SFT+MMD & $\mathrm{L}_{\text{RF}}+\lambda\mathrm{L}_{\text{MMD}}$ & Pretrained, then fine-tuned on ERA5 and FuXi with MMD (full model) \\
\addlinespace[2pt]

\midrule
\multicolumn{4}{l}{\textbf{Target}} \\
\midrule
WRF 3-km & - & - & High-resolution reconstruction reference target \\
\bottomrule
\end{tabularx}
\end{table}
\FloatBarrier

\subsection{Validation of the 3-km TC-centered reference dataset}

We first assess whether the WRF 3-km simulations provide suitable targets for learning intensity-related inner-core structure.
Each simulation is generated in an independent $10^{\circ} \times 10^{\circ}$ domain centered on the corresponding IBTrACS TC position (Supplementary Section~\ref{WRF Simulation and Configuration Parameters} and Supplementary Table~\ref{tab:wrf_config}).
By increasing the representation of intense inner-core winds while retaining the surrounding circulation, TC-centered sampling mitigates the long-tailed wind-speed distribution found in global and basin-scale datasets (Supplementary Section~\ref{sec:ws10m_frequency_multiscale}).
Therefore, the combination of 3-km resolution and TC-centered sampling provides targets that represent sharp radial gradients and local wind maxima associated with TC intensity.

\begin{figure}[!htbp]
    \centering
    \includegraphics[width=0.90\linewidth]{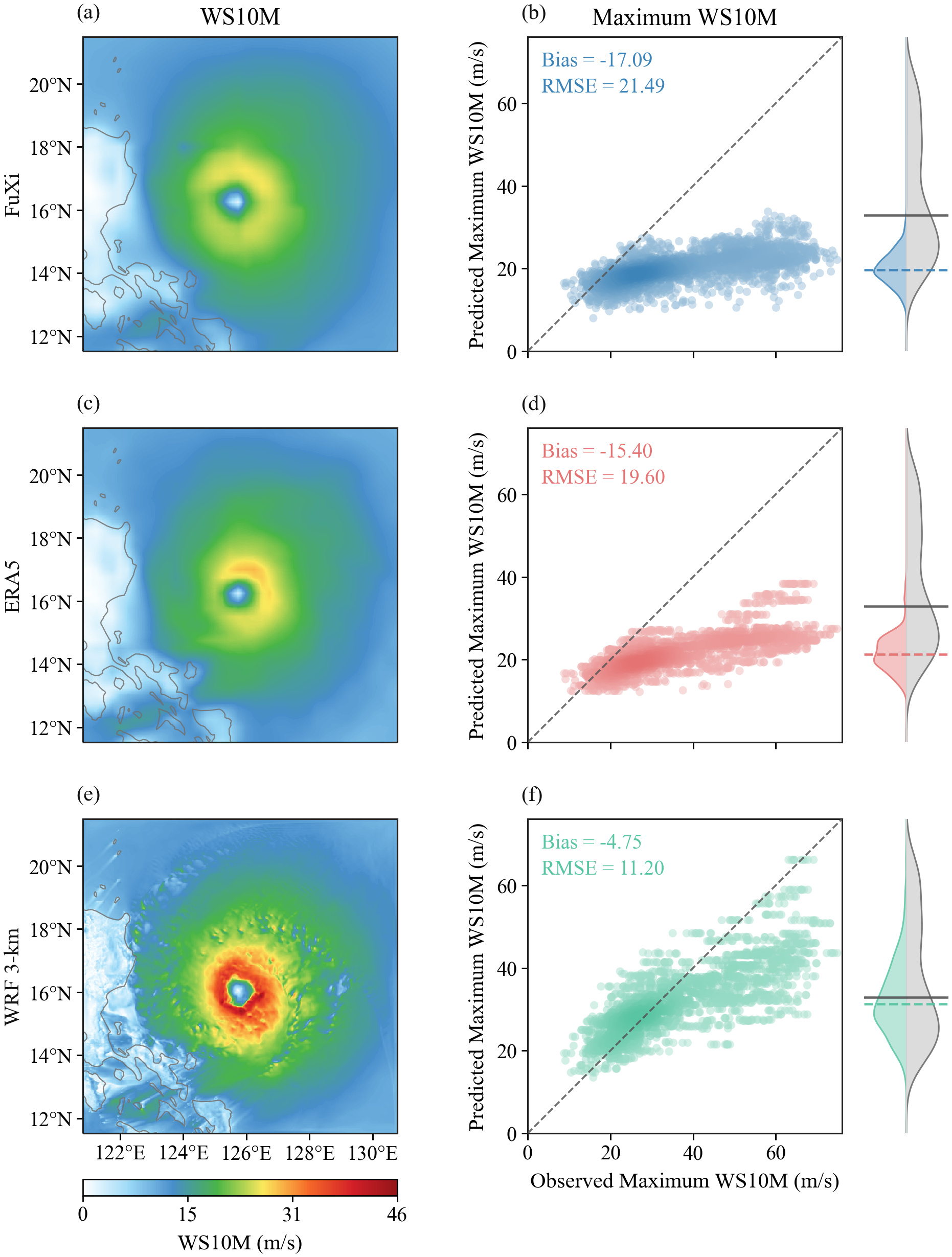}
    \caption{\textbf{Comparison of maximum 10-m wind speed (WS10M) among FuXi, ERA5, and WRF 3-km.} Left panels show the spatial distribution of WS10M for Typhoon DOKSURI at 12 UTC 24 July 2023. Middle panels compare maximum WS10M with IBTrACS observations for 2242 forecasts covering forecast lead times of 6--120 hours in the 2024 WNP testing set. Right panels show kernel density estimates of the intensity distributions. Solid curves denote IBTrACS, and dashed curves denote intensity from ERA5, FuXi, and WRF. Repeated observations arise from 5-kt best-track increments and from forecasts with different initialization times that verify against the same observation. A random jitter of ±2 m/s is applied to the observations for visualization only. All metrics are computed using the original values.}
\label{fig:combined_ws10m_fuxi_era5_wrf3km}
\end{figure}

\cref{fig:combined_ws10m_fuxi_era5_wrf3km} presents spatial maps of WS10M fields from FuXi, ERA5, and WRF for Typhoon DOKSURI at 12 UTC 24 July 2023, as well as comparisons of maximum WS10M against IBTrACS best-track intensity estimates for the testing dataset.
FuXi and ERA5 capture the vortex but smooth the eyewall and localized wind maxima.
In contrast, the WRF 3-km simulation resolves a narrower eye, sharper radial gradients, and more localized high-wind features than FuXi and ERA5.
The comparison of 25-km and 3-km WRF simulations in Supplementary Fig.~\ref{fig:DOKSURI} further shows that the finer grid resolves more detailed inner-core structure in the 10-m wind components, mean sea level pressure (MSL), and WS10M.

Across the 2242 forecasts in the testing dataset, FuXi and ERA5 have WS10M RMSE values of $21.49$ and $19.60$~m/s, respectively, and MBE values of $-17.09$ and $-15.40$~m/s (\cref{fig:combined_ws10m_fuxi_era5_wrf3km}).
The corresponding WRF 3-km RMSE and MBE values are 11.20~m/s and $-4.75$~m/s.
Thus, maximum WS10M from the WRF 3-km simulations shows smaller discrepancies from IBTrACS than either FuXi or ERA5.
Although the residual negative intensity bias shown in \cref{fig:combined_ws10m_fuxi_era5_wrf3km} may propagate into models trained on these simulations, the results support the use of WRF 3-km fields as reference targets for high-resolution reconstruction. Comparisons of spatial structure across resolutions in Supplementary Section~\ref{Validation of the Simulated TC Dataset: A Case Study of Typhoon DOKSURI} provide further support for this use.


\subsection{Tropical cyclone intensity forecast performance}

In this subsection, we evaluate whether reconstructing the three-dimensional vortex improves maximum WS10M forecasts.
Because the reconstruction window is centered on the position supplied by the upstream forecast, 3DTCR neither predicts nor corrects the TC tracks. Consequently, the reconstructed fields inherit the track of the corresponding input source (Supplementary Fig.~\ref{fig:track_error}), and the intensity differences reported below arise from changes in the vortex structure rather than relocation of the TC center.

\begin{figure}[!htbp]
    \centering
    \includegraphics[width=\linewidth]{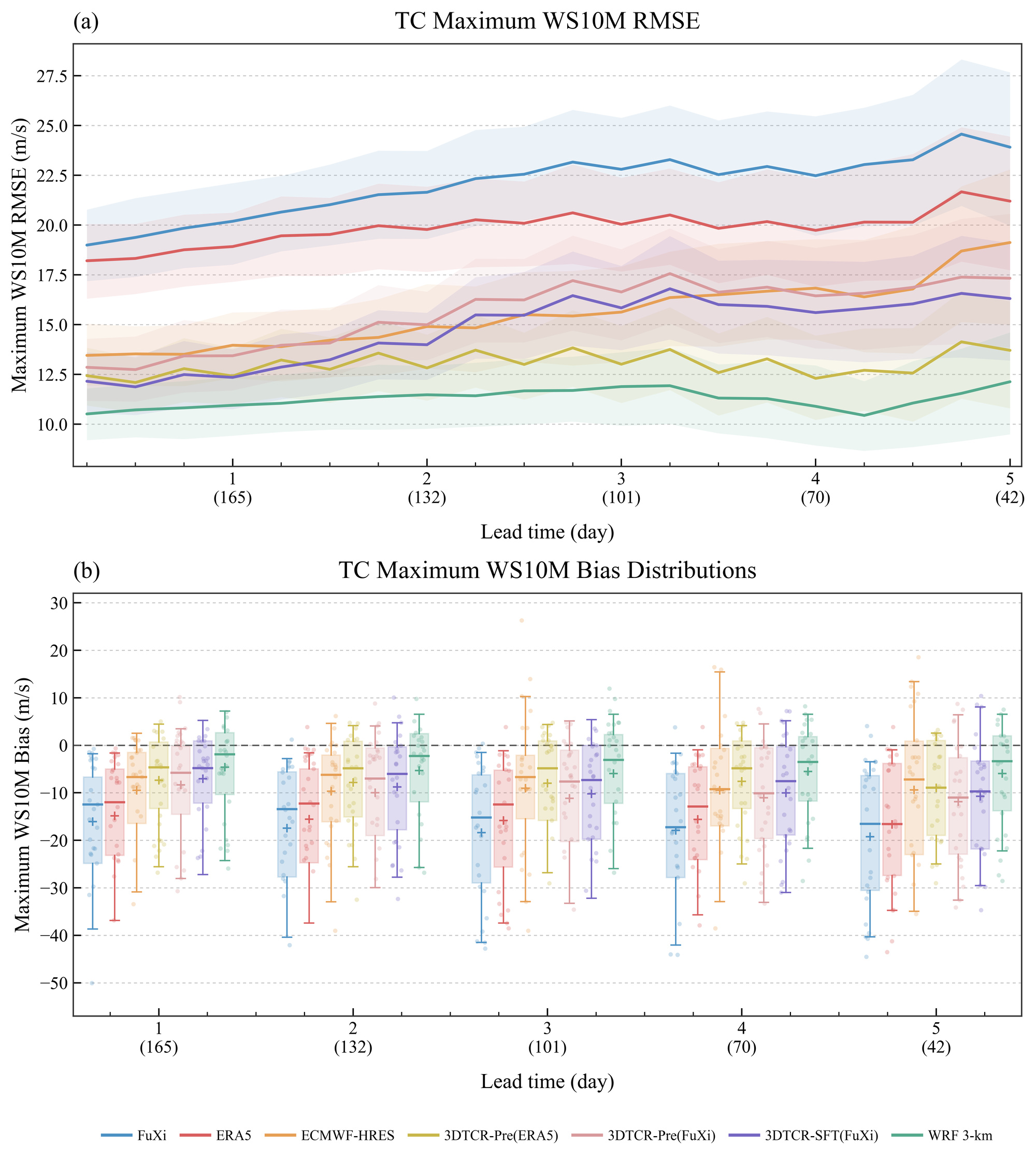}
     \caption{\textbf{Forecast performance for maximum tropical cyclone (TC) 10-m wind speed (WS10M).} (a) TC maximum WS10M RMSE relative to IBTrACS best-track data. Shading denotes 95\% confidence intervals obtained from 1000 bootstrap resamples. (b) Distributions of forecast bias, defined as the predicted minus the best-track maximum WS10M. Boxes indicate the interquartile range, center lines denote the median, plus signs denote the mean, and whiskers extend from the 5th to the 95th percentiles. Points show a random subset of forecast samples. Numbers in parentheses indicate the sample size at each lead time.}
    \label{fig:wind_evaluation}
\end{figure}

As shown in \cref{fig:combined_ws10m_fuxi_era5_wrf3km}, FuXi and ERA5 increasingly underestimate maximum WS10M as observed TC intensity increases.
This underestimation persists across forecast lead times (\cref{fig:wind_evaluation}).
Specifically, FuXi RMSE increases from 18.99 m/s at 6 h to 23.91 m/s at 120 h and exceeds 20 m/s at all lead times from 24 to 120 h, whereas 3DTCR RMSE remains below 17 m/s.
Supplementary Table~\ref{tab:forecast_comparison} summarizes RMSE at 24-h intervals for the baselines, 3DTCR variants, and WRF reference, and Supplementary Fig.~\ref{figures/scatter_distribution_adjusted.png} provides comparisons among all evaluated methods, including all 3DTCR variants.
Across all lead times, 3DTCR reduces RMSE and the magnitude of MBE by 36.5\% and 60.7\%, respectively, relative to FuXi, and reduces RMSE by 30.4\% relative to ERA5. Compared to ECMWF-HRES, 3DTCR reduces RMSE and the magnitude of MBE by 8.9\% and 28.2\%, respectively.
These results indicate that vortex-following reconstruction partially corrects the intensity underestimation from the coarse-resolution input, with improvements maintained throughout the 5-day forecast period.

These improvements are consistent with the intended role of the center-aligned training strategy.
At longer forecast lead times, displacement between the predicted and reference TC positions introduces a pixel-wise \enquote{double penalty} that can favor overly smooth reconstructions (Supplementary Information Section~\ref{sec:ws10m_4src_displacement}).
Center-aligned sampling reduces the positional mismatch presented to the reconstruction model, as it does not correct errors in the TC track from the input FuXi forecast.
Separating the tracking from structural reconstruction allows the generative model to focus on recovering fine-scale inner-core wind structure.

\subsection{Reconstruction of three-dimensional inner-core structures}

We use Super Typhoon KONG-REY (2024) to determine whether improved maximum WS10M is accompanied by coherent changes in multivariable and multilevel vortex structure.
In late October 2024, KONG-REY rapidly intensified, reaching a maximum sustained wind speed of 72 m/s in the Joint Typhoon Warning Center (JTWC) best-track analysis before making landfall in Taitung, eastern Taiwan.

\subsubsection{Horizontal and vertical vortex structure}

\begin{figure}[!htbp]
\centering
\includegraphics[width=1.0\linewidth]{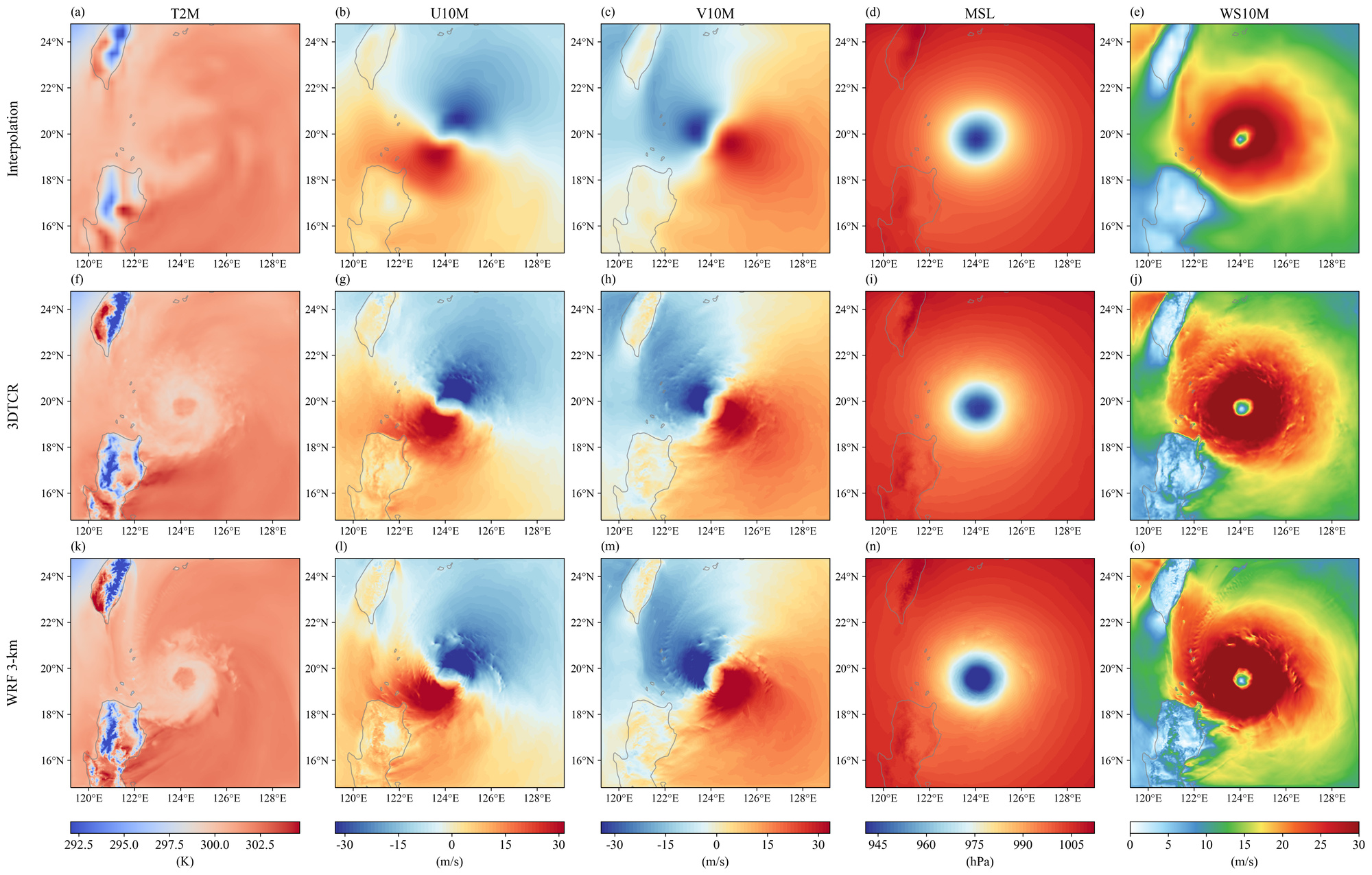}
\caption{\textbf{Near-surface structure of Typhoon KONG-REY at a forecast lead time of 18 h.} The forecast was initialized at 12 UTC on 29 October 2024. Columns show 2-m temperature (T2M), 10-m u and v wind components (U10M and V10M), mean sea level pressure (MSL), and 10-m wind speed (WS10M). Rows show bilinearly interpolated FuXi fields, the 3DTCR reconstruction, and the WRF 3-km reference.}
\label{figures/KONG-REY_T2M_U10_V10_MSL_WS10.jpg}
\end{figure}

\textbf{Spatial structure of near-surface fields.}
\cref{figures/KONG-REY_T2M_U10_V10_MSL_WS10.jpg} compares bilinearly interpolated FuXi fields, the 3DTCR reconstruction, and the WRF 3-km reference at a lead time of 18 h from the 12 UTC 29 October 2024 initialization.
The interpolated FuXi fields preserve the large-scale environment but exhibit broad, smoothed gradients around the vortex.
In contrast, 3DTCR recovers finer-scale variability in 2-m temperature (T2M) and the 10-m u and v wind components, sharpens the transition between the eye and eyewall, and produces an asymmetric WS10M pattern that more closely resembles the WRF 3-km reference.
The large-scale pressure pattern and vortex position remain consistent with the FuXi input.

Similar patterns occur for KONG-REY at 96 h and for Typhoons YAGI and YINXING at lead times from 18 to 108 h (Supplementary Figs.~\ref{figures/KONG-REY_Init2024102612_LeadStep16.jpg},~\ref{figures/YAGI_Init2024090312_LeadStep3.jpg},~\ref{figures/YAGI_Init2024090200_LeadStep18.jpg}, and~\ref{figures/YINXING_Init2024110412_LeadStep12.jpg}).
Across these cases, 3DTCR produces sharper wind gradients between the eye and eyewall and greater azimuthal variability than the interpolated FuXi fields.

\textbf{Vertical wind structure.}
To evaluate structure above the surface, we examine the latitudinal and longitudinal wind speed cross-sections of KONG-REY.
The reconstructed fields contain a weak-wind eye flanked by deep high-wind cores and retain the asymmetric inner-core structure (Supplementary Fig.~\ref{fig:KONG-REY_vertical_cross_sections_all_2024-10-30T06-00-00_black_contour_k.png}).
We further decompose the horizontal wind into radial and tangential components relative to the TC center and calculate azimuthal means at each radius and pressure level (Supplementary Fig.~\ref{fig:KONG-REY_fuxi_wrf_tangential_radial_winds_diff.png}).
The mean tangential wind shows a pronounced eyewall maximum extending through much of the troposphere, whereas the mean radial wind shows low-level inflow and upper-level outflow of the secondary circulation.
Together with the inner-core asymmetries, these features indicate that 3DTCR reconstructs an organized three-dimensional circulation rather than simply increasing the near-surface wind field.

\textbf{Multilevel and temporal consistency.}
The reconstructed geopotential and wind fields preserve a vertically coherent vortex at 500 hPa, 850 hPa, and the surface (Supplementary Figs.~\ref{fig:3d_reconstruction},~\ref{fig:KONG-REY_circulation_500.png},~\ref{fig:KONG-REY_circulation_850.png}, and~\ref{fig:KONG-REY_circulation_surface.png}).
In the selected case, the reconstructed 10-m surface wind field generally follows the spiral cloud bands visible in the satellite image, particularly in the inflow regions (see Supplementary Fig.~\ref{fig:KONG-REY_Cloud_Wind_barbs.png}).
This comparison provides qualitative, rather than quantitative, support for the physical consistency of the reconstructed surface circulation.

The predicted sequence also shows the temporal evolution of MSL and 10-m wind vectors, including contraction of the wind field during TC intensification (Supplementary Fig.~\ref{fig:KONG-REY_2024-10-29_06-00-00_tc_track_add_msl_barbs.png}).
Typhoon YAGI exhibits similar vertical wind distributions, circulation patterns, cloud-relative surface winds, and temporal evolution (Supplementary Figs.~\ref{fig:YAGI_vertical_cross_sections_all_2024-09-04T06-00-00_black_contour_k.png},~\ref{fig:YAGI_fuxi_wrf_tangential_radial_winds_diff.png},~\ref{fig:3d_reconstruction_yagi.png},~\ref{fig:YAGI_circulation_500.png},~\ref{fig:YAGI_circulation_850.png},~\ref{fig:YAGI_circulation_surface.png},~\ref{fig:YAGI_Cloud_Wind_barbs.png}, and~\ref{fig:YAGI_2024-09-04_00-00-00_tc_track_add_msl_barbs.png}).

\textbf{Transfer to other basins.}
To examine transfer beyond the WNP training domain, we apply 3DTCR without retraining to representative TCs in the eastern Pacific and North Atlantic.
3DTCR also produces localized inner-core structures for Tropical Storm JOHN in the eastern Pacific and Hurricane MILTON in the North Atlantic, neither of which is included in training (Supplementary Section~\ref{sec:cross_basin_generalization}).
As WRF 3-km reference fields are not generated for these basins, these cases provide qualitative evidence of transfer but do not establish quantitative cross-basin generalization.

\subsubsection{Spectral and harmonic analysis}

\textbf{Wind-speed distributions and spectra.}
Furthermore, we use wind speed probability density functions (PDFs) and spectral diagnostics to assess variability across spatial scales.
PDFs characterize the distribution of wind speed, including its high-wind tail, whereas energy spectra quantify variability by spatial scale.
High wavenumbers represent smaller-scale structures, where numerical smoothing can lead to excessive attenuation of energy and loss of fine-scale variability.

\begin{figure}[!htbp]
    \centering
    \includegraphics[width=\linewidth]{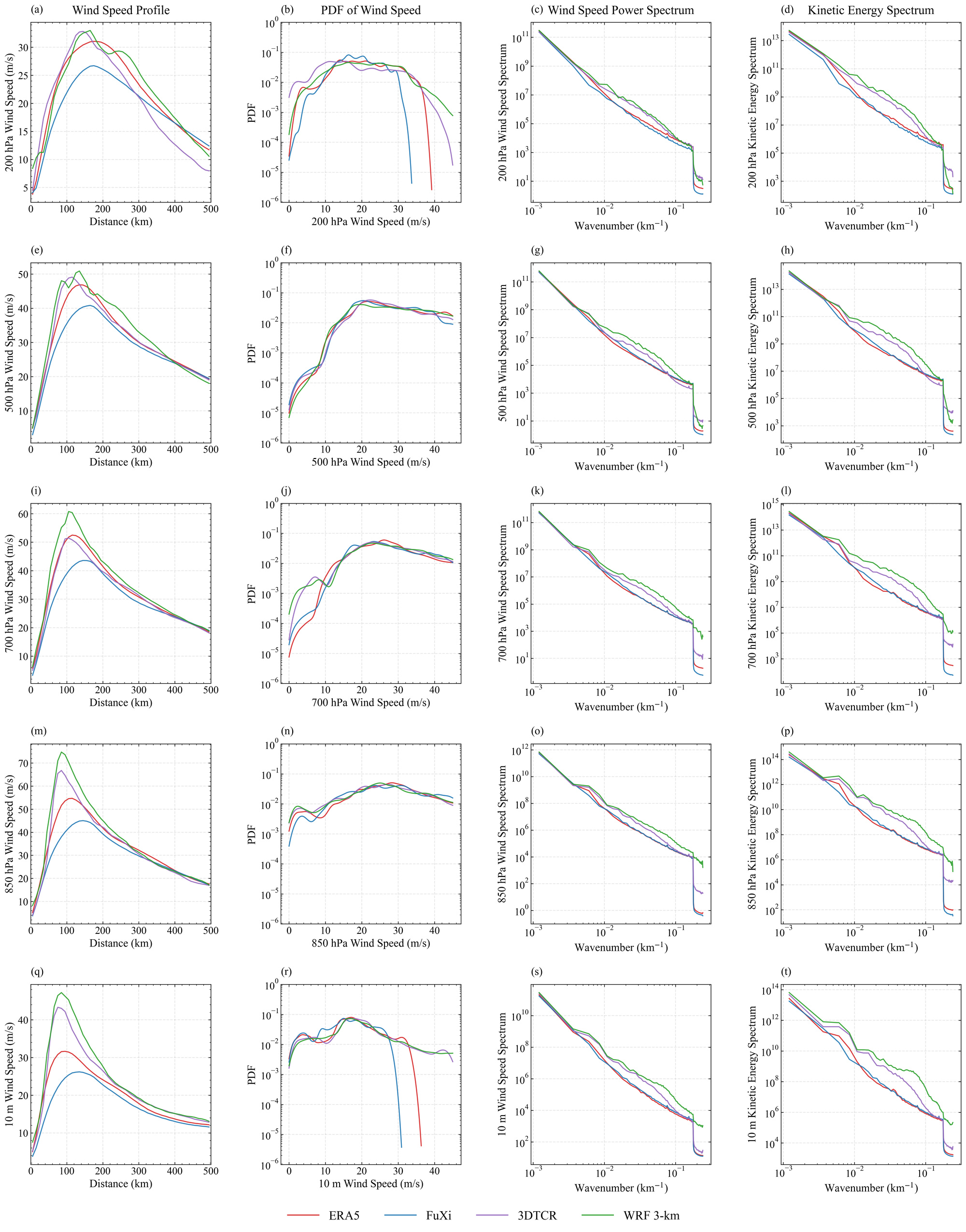}
    \caption{\textbf{Multilevel wind-speed distributions and spectra for Typhoon KONG-REY.}
    From top to bottom, rows show winds at 200 hPa, 500 hPa, 700 hPa, 850 hPa, and at 10 m.
    From left to right, columns show radial wind speed profiles, probability density functions, wind speed power spectra, and kinetic energy spectra. Results are shown for ERA5 (red), FuXi (blue), 3DTCR (purple), and WRF 3-km (green). Both spectra are plotted on logarithmic axes.}
    \label{KONG-REY_comprehensive_multi_level_ws_energy.png}
\end{figure}

For KONG-REY, \cref{KONG-REY_comprehensive_multi_level_ws_energy.png} compares wind speed profiles, PDFs, wind speed spectra, and kinetic energy spectra at 200, 500, 700, and 850~hPa and at 10 m.
The coarse-resolution inputs yield narrower PDFs and underestimate the occurrence of high wind speeds.
The 3DTCR distribution is broader and more closely approaches the WRF 3-km reference, suggesting improved representation of intense inner-core winds.
At low wavenumbers, all datasets have similar spectral characteristics, consistent with their representation of the large-scale circulation.
However, at high wavenumbers, energy decays more rapidly in FuXi and ERA5 than in WRF 3-km.
In contrast, the 3DTCR spectra retain more high-wavenumber variability and more closely follow the WRF 3-km reference, indicating partial recovery of the small-scale variability suppressed in the coarse-resolution inputs.
Typhoon YAGI (Supplementary Fig.~\ref{fig:YAGI_comprehensive_multi_level_ws_energy.png}) shows the same qualitative pattern.
Together, these case studies indicate that 3DTCR recovers variability across a broader range of spatial scales.

\textbf{Azimuthal harmonic analysis.}
We transform the WS10M field into TC-centered polar coordinates and apply Fourier decomposition along the azimuthal direction to characterize vortex asymmetry (Supplementary Figs.~\ref{fig:WRF3km_KONG-REY_wave_decomposition_5x4.png} and~\ref{fig:KONG-REY_wind_speed_wave_decomposition_20.png}).
Wavenumber 0 represents the azimuthally symmetric mean vortex, whereas higher-order harmonics (wavenumbers $\geq 1$) characterize departures from axisymmetry and represent increasingly complex asymmetric vortex structures.
The interpolation baseline is dominated by wavenumber 0 and has relatively weak higher-order components.
In contrast, 3DTCR produces stronger selected asymmetric components and a synthesis of wavenumbers 0--20 that is closer to the WRF 3-km reference.
Typhoon YAGI shows a similar pattern (Supplementary Figs.~\ref{fig:WRF3km_YAGI_wave_decomposition_5x4.png} and~\ref{fig:YAGI_wind_speed_wave_decomposition_20.png}).
These case studies suggest that 3DTCR reconstructs both the dominant symmetric vortex and aspects of the asymmetric structure that are attenuated in the interpolated coarse-resolution fields.

\subsection{Two-stage training and domain-aligned fine-tuning improve extreme wind reconstruction}

As forecast errors accumulate with lead time, operational forecasts differ from reanalyses and show spatial displacement, intensity bias, and structural smoothing (Supplementary Section~\ref{sec:ws10m_4src_displacement}).
These differences are particularly consequential for intense inner-core winds.
Although generative models can alleviate the smoothing associated with deterministic regression, direct training on forecast fields requires the model to correct forecast biases while simultaneously reconstructing fine-scale structure.

In this subsection, we use ablation experiments to evaluate the contributions of two design choices: two-stage training and MMD-based latent space alignment.
Table~\ref{tab:method_comparison} summarizes the model configurations, objectives, and inference inputs.
Supplementary Section~\ref{app:ablation_framework} describes the complete experimental design.
Supplementary Sections~\ref{app:training_objectives} and \ref{app:mmd_application} provide the two-stage training objectives and implementation of MMD alignment, respectively.
Since the primary objective is to recover rare extreme winds, we evaluate the critical success index (CSI) at WS10M thresholds from 13.9 to 32.7~m/s.

\begin{figure}[!htbp]
    \centering
    \includegraphics[width=\textwidth]{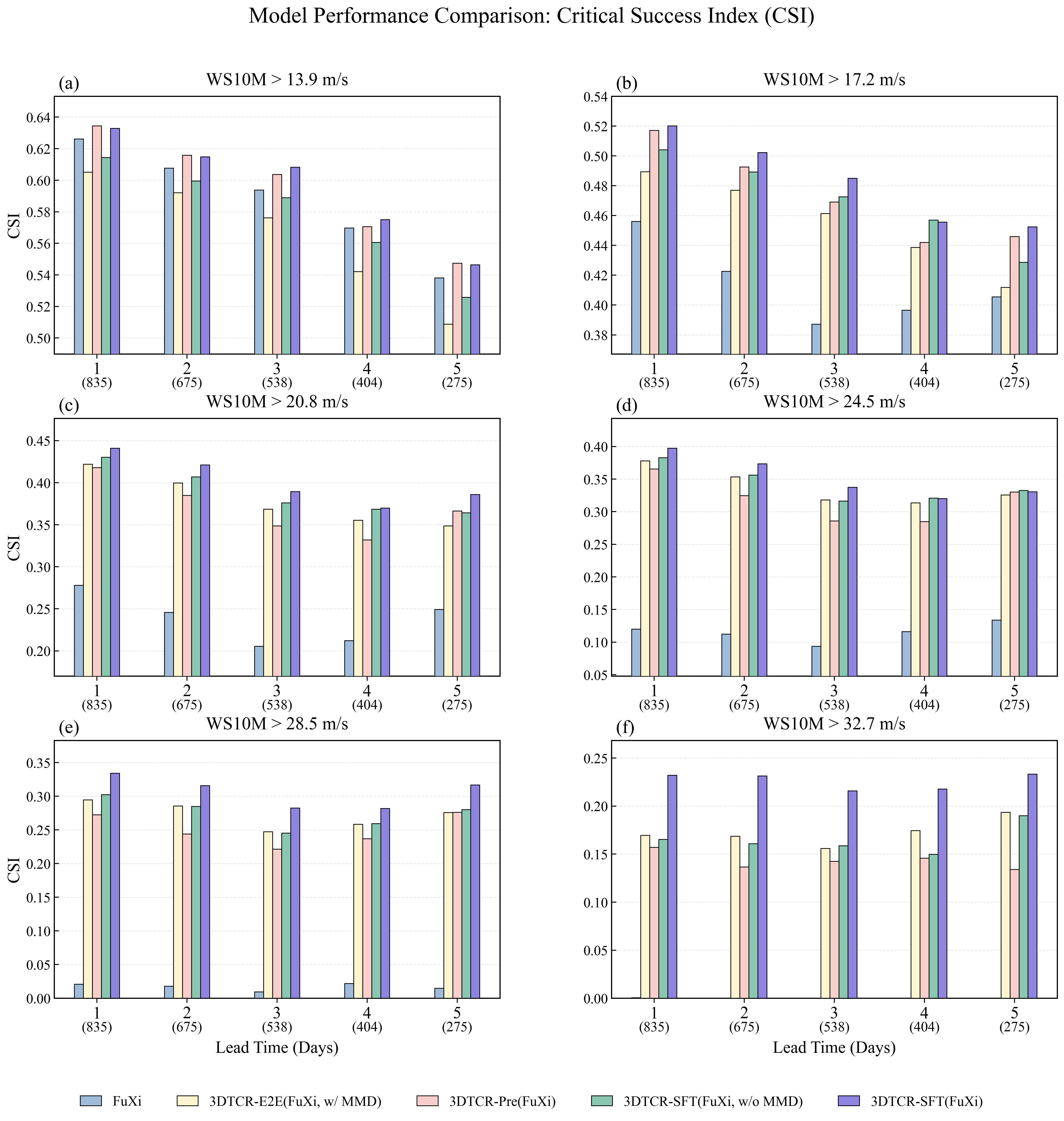}
    \caption{
\textbf{Ablation analysis of 3DTCR training strategies.} Critical success index (CSI) for 10-m wind speed (WS10M) exceeding six thresholds from 13.9 to 32.7 m/s at lead times of 1--5 days (panels \textbf{a}--\textbf{f}). Bar groups compare FuXi with 3DTCR variants trained with pretraining (3DTCR-Pre), supervised fine-tuning (3DTCR-SFT), and end-to-end training (3DTCR-E2E). Numbers in parentheses denote the sample size at each lead day. Parenthetical labels denote the inference input source. For example, 3DTCR-Pre (FuXi) denotes the pretrained model applied to FuXi inputs at inference. The labels \enquote{w/ MMD} and \enquote{w/o MMD} indicate whether MMD-based latent space alignment is applied.}
\label{figures/experiment_fuxifinetune_evalpaperv3_vlpips_changecsiv6_sample_interval_1_csi_daily.png}
\end{figure}

As shown in \cref{figures/experiment_fuxifinetune_evalpaperv3_vlpips_changecsiv6_sample_interval_1_csi_daily.png}, MMD-based latent space alignment improves extreme wind reconstruction across the evaluated thresholds.
The full 3DTCR-SFT model has higher CSI scores than the corresponding variant without MMD, indicating more accurate identification of high-wind regions.
Supplementary Fig.~\ref{figures/scatter_distribution_adjusted.png} compares the predicted and observed maximum wind speeds.
The 1:1 line denotes perfect agreement, with points above and below the line indicating overestimation and underestimation, respectively.
The full model also has lower bias and clusters more tightly around the 1:1 line than the variant without MMD.
These comparisons are consistent with MMD-based latent alignment reducing the effect of the distribution mismatch between FuXi forecasts and ERA5 reanalysis on the reconstruction.
The implementation of MMD alignment at the U-Net bottleneck is described in Supplementary Section~\ref{app:mmd_application}.

The comparison between 3DTCR-E2E and the full 3DTCR model further demonstrates the benefit of the two-stage training strategy, which separates structural learning from forecast-domain adaptation (purple bars). Despite using the same multi-task objective, the E2E model performs less well at the most extreme threshold ($\mathrm{WS10M} > 32.7$~m/s;~\cref{figures/experiment_fuxifinetune_evalpaperv3_vlpips_changecsiv6_sample_interval_1_csi_daily.png}).
This pattern is consistent with the challenge of learning two coupled corrections from random initialization: an amplitude correction for systematically underestimated extreme winds and a structural correction for the growing mismatch between FuXi forecasts and WRF 3-km targets.

A two-stage training strategy is used to separate these two tasks.
Pretraining first learns the coarse-to-fine mapping from spatially aligned ERA5-WRF pairs.
Since the WRF targets are produced by dynamical downscaling from ERA5, these input and target pairs retain a relatively strong spatial consistency.
Center-aligned fine-tuning then adapts the learned mapping to FuXi forecasts with lead time conditioning and MMD-based latent alignment.
This task decomposition reduces the need to learn forecast bias correction and fine-scale reconstruction simultaneously. 
Consistent with this interpretation, the scatter plot (Supplementary Fig.~\ref{figures/scatter_distribution_adjusted.png}) shows that the E2E variant has larger bias and RMSE than the full two-stage model, particularly at high wind speeds.
For Typhoon AMPIL, the full model also produces more coherent inner-core spatial structure and fields that more closely resemble the WRF 3-km reference than either the E2E model or the SFT variant without MMD (Supplementary Fig.~\ref{fig:ampil_ablation_multivariable_reconstruction}). 
Together, these results support two-stage training with latent space alignment for reconstructing rare extreme wind structures.

\subsection{Vortex-following reconstruction reduces computational cost}
\label{Computational Efficiency}

Basin-wide TC simulations are computationally expensive at 3-km spatial resolution.
Full-basin coverage is helpful, but the fine-scale structures most relevant to intensity are concentrated near the TC inner core.
The rationale for targeting this region and the associated trade-off between basin-wide coverage and inner-core fidelity are discussed in Supplementary Section~\ref{sec:Computational Cost Comparison and the Necessity of vortex-following reconstruction}.

The vortex-following design restricts high-resolution reconstruction to
a \ensuremath{10^{\circ} \times 10^{\circ}} window comprising
\ensuremath{370 \times 370} grid points, while retaining the global model forecasts to represent the large-scale environment. 
As shown in Supplementary Table~\ref{tab:computational_cost}, a 5-day, full-basin WRF simulation at approximately 3-km resolution (1852 $\times$ 2592 grid points) requires approximately 400 hours on 32 CPU cores.
Restricting WRF to the TC-centered domain with 370 $\times$ 370 grid points reduces the runtime to approximately 12 hours on the same CPU configuration.
3DTCR generates the corresponding 5-day sequences of TC-centered fields in approximately 3 minutes on a single NVIDIA A100 GPU.
This substantial runtime difference highlights the practical computational advantage of targeted generative reconstruction as a complement to high-resolution numerical simulation.

\section{Discussion}

Accurate forecasting of TC intensity remains challenging because intensity change depends on localized inner-core processes that global NWP and AI systems resolve incompletely.
3DTCR addresses this limitation by reconstructing three-dimensional TC fields at 3-km resolution within a moving $10^{\circ}\times10^{\circ}$ window.
Evaluated over the independent 2024 WNP testing dataset, 3DTCR reduces the RMSE and the magnitude of the MBE for maximum WS10M by 36.5\% and 60.7\% relative to FuXi, and by 8.9\% and 28.2\% compared to ECMWF-HRES.
Case studies further suggest that the reconstructed fields contain sharper wind gradients between the eye and eyewall, coherent vertical structure of horizontal winds, and vortex asymmetries.
These results demonstrate that 3DTCR improves TC intensity forecasts by reconstructing the three-dimensional vortex rather than correcting a scalar intensity alone.

The TC-centered WRF dataset and two-stage training strategy are two design elements contributing to these improvements. 
TC-centered sampling increases the representation of intense winds and sharp spatial gradients in the training distribution, which occupy the sparsely sampled tail of global and basin-scale datasets.
Pretraining on spatially aligned ERA5-WRF pairs first establishes a coarse-to-fine mapping.
Center-aligned fine-tuning then adapts this mapping to FuXi forecasts through lead time conditioning and MMD-based latent space alignment.
This separation addresses a major challenge in forecast-based reconstruction.
When the predicted and reference TCs are displaced, a pixel-wise objective penalizes a realistic TC at both locations. 
This \enquote{double penalty} favors spatially averaged, overly smooth fields \cite{ebert2013progress,subich2025fixing,wang2022customized}.
Center alignment reduces the positional mismatch presented to the reconstruction model without correcting the TC track provided by the driving forecasts.
In the ablation experiments, the two-stage model performs better than the model trained from random initialization at the most extreme wind speed threshold, and latent space alignment improves the CSI scores across the evaluated thresholds.
These findings are consistent with separate structural pretraining and forecast domain adaptation being useful for recovering rare extreme wind structures.

3DTCR also provides an alternative to high-resolution reconstruction over a fixed geographic domain.
Most regional AI forecasting and downscaling approaches restrict enhanced resolution to a predefined domain \cite{nipen2026regional,wijnands2025comparison,bano2025regional,mardani2025residual}.
Such a domain is not necessarily well suited to a compact weather system that travels long distances during its lifetime.
Inspired by vortex-following and moving-nest numerical models \cite{roberts2020impact}, 3DTCR preserves the global model's large-scale environmental guidance while targeting high-resolution reconstruction around the predicted TC center.
It generates 5-day TC-centered fields in approximately 3 minutes on a single GPU, whereas the full-basin 3-km WRF configuration used here requires about 400 hours on 32 CPU cores, demonstrating the practical value of combining global coverage with targeted regional refinement.
The eastern Pacific and North Atlantic cases suggest that some learned TC representations may transfer across basins, consistent with recent evidence that AI-based weather forecasting models trained on intense TCs in one basin can retain predictive skill in another \cite{sun2025can}.
However, this remains qualitative, because high-resolution reference fields are unavailable for the two out-of-basin cases.

Several limitations point to directions for further improvement.
First, the 3-km WRF targets inherit errors from ERA5 and the selected WRF configuration. 
Their residual negative intensity bias may constrain the extremes learned by 3DTCR and lead to the remaining discrepancy from IBTrACS (\cref{fig:combined_ws10m_fuxi_era5_wrf3km}).
Incorporating aircraft, radar, and high-resolution satellite observations, together with sensitivity experiments using alternative WRF physics, would reduce reliance on a single simulated reference. 
Second, quantitative evaluation is restricted to the WNP and a single independent testing year, while the two out-of-basin cases lack high-resolution reference fields and provide only qualitative evidence of transfer. Future work could further explore the model’s applicability across basins and years.
Third, strong wind gradients remain partially smoothed at longer lead times.
Ensemble reconstruction could help characterize the associated uncertainty and better represent extremes \cite{zhong2025fuxi}.
Finally, 3DTCR inherits errors in the TC track and large-scale guidance provided by the driving forecast, leaves fields outside the moving window unchanged, and does not feed the reconstructed vortex back into the global forecast.
Therefore, this study evaluates diagnostic reconstruction of the predicted vortex but does not show whether the reconstructed inner-core structure would change subsequent TC evolution.
Coupling the regional reconstruction to the global model is an important direction for addressing this question.

Despite these limitations, these results indicate that TC-centered supervision and vortex-following generative reconstruction can enhance global forecasts with localized, three-dimensional vortex structure.
By focusing kilometer-scale reconstruction where it is most consequential, 3DTCR improves TC intensity prediction without requiring basin-wide high-resolution numerical integration.
More broadly, it provides a scalable framework for adding local detail to global predictions of other localized, moving, high-impact weather systems.

\section{Materials and methods}
\label{Materials_and_methods}

\subsection{Data}
\label{sec:method}

We use version 4.6 of the Weather Research and Forecasting model \cite{skamarock2019description} to generate high-resolution simulations of tropical cyclones (TCs) in the western North Pacific (WNP).
ERA5 reanalysis data \cite{hersbach2020era5} at $0.25^\circ$ resolution provide the initial and boundary conditions (Supplementary Information Section~\ref{ECMWF Reanalysis v5 (ERA5)}).
Instead of using a fixed regional domain, we conduct restart-based, TC-centered WRF integrations in a $10^\circ \times 10^\circ$ domain whose reference position is updated to follow the TC positions from the International Best Track Archive for Climate Stewardship (IBTrACS).
We include TC cases that reach at least tropical storm intensity, defined here as an IBTrACS maximum sustained wind speed exceeding 17.2~m/s.
This design focuses the high-resolution simulation on the TC-centered region containing the inner-core wind structures and sharp gradients.
The complete WRF configuration is provided in Supplementary Information Section~\ref{WRF Simulation and Configuration Parameters}. 

The ERA5-driven WRF simulations yield 1903 samples at 3-km resolution, each containing 93 meteorological variables at the surface and on multiple pressure levels (Supplementary Table~\ref{tab:variable_info}).
The outputs are regridded to $0.027^\circ$ ($370 \times 370$ grid points) and are used as the high-resolution targets for model training and evaluation.
IBTrACS provides the best-track TC positions and intensities.
Intensity is defined by \text{USA\_WIND}, the 1-min mean maximum sustained wind speed from the U.S. agencies contributing to IBTrACS.
FuXi forecasts provide the inputs for supervised fine-tuning and inference, and ECMWF-HRES serves as the numerical weather prediction (NWP) baseline.
Samples from 2018--2023 are used for training, and all samples from 2024 are reserved for independent testing.

For model training, we select 21 of the 93 meteorological variables to balance the representation of TC structure against memory and computational constraints.
The 16 upper-air variables are geopotential (Z), temperature (T), and the zonal and meridional wind components (U and V), each at 850, 700, 500, and 200 hPa.
The five surface variables are 2-m temperature ($\mathrm{T2M}$), mean sea level pressure ($\mathrm{MSL}$), 10-m wind speed ($\mathrm{WS10M}$), and 10-m u and v wind components ($\mathrm{U10M}$ and $\mathrm{V10M}$).
These variables characterize the three-dimensional circulation and near-surface wind structure of TCs.
Supplementary Section~\ref{Detailed Description of WRF Post-Processed and Model Training Variables} provides complete variable definitions and post-processing procedures.

\begin{figure}[!htbp]
    \centering
    \includegraphics[width=\linewidth]{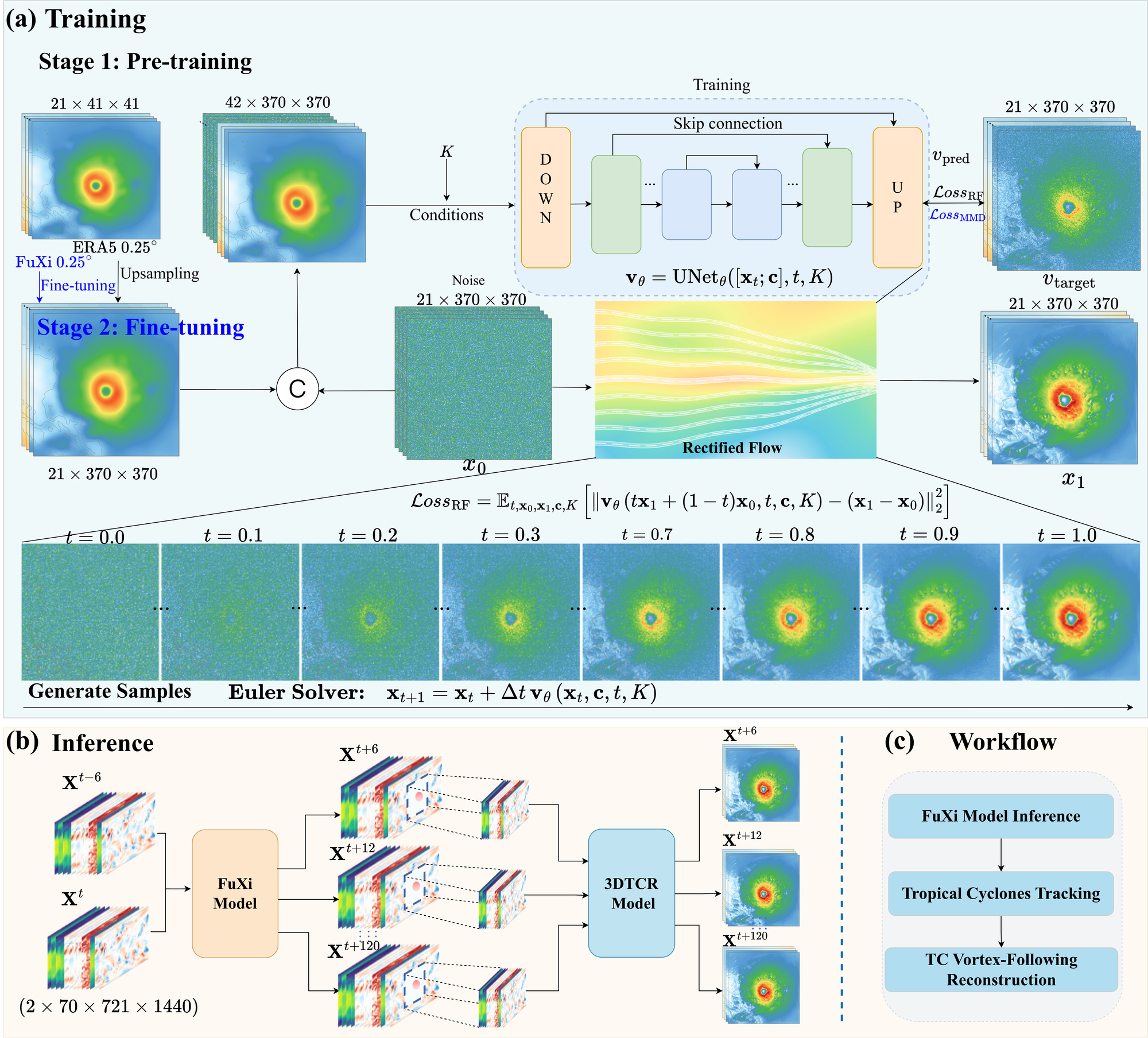}
    \caption{\textbf{Overview of the 3DTCR framework.} (a) During pretraining, the model learns a coarse-to-fine mapping from spatially aligned ERA5 inputs to WRF 3-km targets using Rectified Flow. During supervised fine-tuning (blue arrows), ERA5 and FuXi forecast inputs are introduced, and a maximum mean discrepancy (MMD) loss aligns their latent space feature distributions at the U-Net bottleneck. (b) At inference, the TC tracker locates the TC in each FuXi forecast, a TC-centered window is extracted, and 3DTCR reconstructs the high-resolution vortex at each lead time.}
    \label{arch}    
\end{figure}

\subsection{3DTCR model architecture}

3DTCR is a conditional generative model that uses conditional flow matching (CFM) within a Rectified Flow formulation to reconstruct TC-centered meteorological fields from coarse-resolution global AI weather forecasts (Supplementary Information Section~\ref{Rectified Flow}).
At flow time $t$, the model concatenates the interpolated state $x_t \in \mathbb{R}^{21 \times 370 \times 370}$ with a spatially aligned coarse-resolution condition $c$ of the same dimensions, producing an input of dimensions $42 \times 370 \times 370$.
The continuous flow time $t$ and discrete forecast lead time index $\mathrm{K}$ are provided as global conditioning variables, allowing the model to represent changes in forecast errors with lead time.
As illustrated in \cref{arch}, the velocity estimator $v_\theta$ is parameterized by a U-Net with three downsampling stages and three symmetric upsampling stages.
Attention modules in the deep latent layers capture long-range spatial dependencies between multiscale environments and inner-core structures.
By minimizing the Rectified Flow loss $\mathrm{L}_{\text{RF}}$, the model learns the velocity field $v_\theta$ that transports a Gaussian prior toward the conditional distribution of the high-resolution atmospheric state.
Further details of the model architecture are provided in Supplementary Information Section~\ref{app:training_details}.

\subsection{Model training}
\label{sec:algorithm_details}

3DTCR is trained in two stages to separate learning of the reconstruction mapping from adaptation to FuXi forecast inputs.
During pretraining, the model learns a mapping from spatially aligned ERA5 inputs to WRF 3-km targets, thereby avoiding the displacement and intensity errors present in FuXi forecasts.
The pretrained weights are then used to initialize supervised fine-tuning with ERA5 reanalysis and FuXi forecasts.
During fine-tuning, we continue to optimize the Rectified Flow reconstruction objective and apply a maximum mean discrepancy (MMD) loss to the U-Net bottleneck features to reduce the discrepancy between the FuXi and ERA5 latent distributions.
Forecast lead time is included as a conditioning variable to account for the changing error distribution as the forecast errors accumulate.
To mitigate the double penalty associated with displacement and pixel-wise supervision, the FuXi forecast and WRF target windows are centered on their respective TC centers during fine-tuning.
Detailed formulations of the loss functions and latent alignment, as well as network configuration and optimization settings, are provided in Supplementary Information Section~\ref{app:training_details}.

\subsection{Experimental design}

All ablation variants use the same U-Net backbone and WRF 3-km reconstruction targets.
To isolate the contribution of two-stage training, we compare the full model with 3DTCR-E2E (FuXi, w/ MMD), which is trained from random initialization.
To assess MMD-based latent alignment, we compare the full model with 3DTCR-SFT (FuXi, w/o MMD).
The 3DTCR-Pre variants provide additional references using the pretrained model directly for inference on ERA5 and FuXi inputs, without further fine-tuning.
Table~\ref{tab:method_comparison} summarizes all model configurations, with further details provided in Supplementary Section~\ref{app:ablation_framework}.

\subsection{Vortex-following inference}

Because the training samples consist of moving windows centered on TCs at different locations, the model learns vortex structure relative to the TC center rather than background patterns tied to a fixed geographical domain.
At each forecast lead time, a lightweight TC tracker locates the predicted TC in the FuXi forecast (Supplementary Information Section~\ref{Tropical Cyclone Tracking Method}).
We then extract a $10^\circ \times 10^\circ$ domain, approximately 1000 km on a side, and reconstruct it at 3-km resolution.
This procedure is repeated at each tracked position.
Therefore, the reconstructed fields inherit the TC position from the FuXi forecast while modifying the intensity and structure relative to that center.
This strategy is analogous to the moving-nest approach used in NWP models, focusing computation on the TC inner core while avoiding the substantial cost of full-basin, high-resolution simulation.

\subsection{Evaluation methods}

We evaluate 3DTCR using TC track and intensity time series and the reconstructed WS10M field, defining intensity as the maximum 10-m wind speed in the reconstructed field and evaluating both track and intensity against IBTrACS at each lead time. Track errors are quantified using the mean absolute error (MAE), whereas errors in maximum WS10M are measured using the root mean square error (RMSE) and mean bias error (MBE).
To assess the spatial reconstruction of intense wind regions, we calculate the critical success index (CSI) at six $\mathrm{WS10M}$ thresholds of 13.9, 17.2, 20.8, 24.5, 28.5, and 32.7~m/s.
Definitions of all evaluation metrics are provided in Supplementary Information Section~\ref{Evaluation metrics details}.

\pagestyle{plain} 
\section*{Data availability}
The ERA5 reanalysis dataset is available from the official website of the Copernicus Climate Data Store (CDS) at \url{https://cds.climate.copernicus.eu/}.
The TIGGE dataset is available at \url{https://apps.ecmwf.int/datasets/data/tigge/levtype=sfc/type=cf/}.
ECMWF-HRES TC tracks were retrieved from the TIGGE archive in the form of downloadable XML files, which can be accessed via \url{https://confluence.ecmwf.int/display/TIGGE/Tools}.
Additionally, the International Best Track Archive for Climate Stewardship (IBTrACS) data are publicly available at \url{https://www.ncei.noaa.gov/products/international-best-track-archive}.
Himawari satellite data are obtained from the Japan Meteorological Agency (JMA), with general information available at \url{https://www.jma.go.jp/jma/jma-eng/satellite/}, and the data can be downloaded from the JAXA P-Tree system at \url{https://www.eorc.jaxa.jp/ptree/}.
The FuXi forecast fields and the WRF simulation dataset generated in this study are available from the corresponding authors upon reasonable request.

\pagestyle{plain} 
\section*{Code availability} 
The FuXi forecasting model itself is available at \url{https://zenodo.org/records/10401602}.
The code for 3DTCR can be accessed from \url{https://zenodo.org/records/22842154}.
The WRF version 4.6 source code is available at \url{https://www2.mmm.ucar.edu/wrf/users/download/get_source.html}.

\pagestyle{plain}
\section*{Acknowledgments}
We thank ECMWF for providing the ERA5 reanalysis dataset and HRES forecasts, and the NOAA National Centers for Environmental Information for maintaining the IBTrACS dataset.
We sincerely thank Prof.\ Jie Feng and Dr.\ Zeyi Niu from the Department of Atmospheric and Oceanic Sciences at Fudan University for their valuable guidance and insightful suggestions throughout this research. We thank Shun Dai and Yangyang Zhao from the Shanghai Academy of Artificial Intelligence for Science (SAIS) for their helpful suggestions and assistance. We are grateful to SAIS for providing the internship opportunity during which this work was primarily completed.
The computations were performed using the Computing for the Future at Fudan (CFFF) platform.

\pagestyle{plain}
\section*{Author contributions}
Jun Liu constructed, post-processed, and validated the dataset; designed the experimental and evaluation frameworks; trained and evaluated the model; and wrote the manuscript. Xiaohui Zhong contributed to project planning, provided guidance, and reviewed the manuscript. Kai Zheng conducted a subset of the comparative experiments. Jiarui Li performed data visualization and contributed to manuscript preparation. Yifei Li contributed to a subset of the model evaluation. Tao Zhou conducted the tropical cyclone mechanism analysis. Wenwen Zhang reviewed and edited the manuscript for language, clarity, and presentation. Hao Li supervised the project, led the overall planning and management, reviewed the manuscript, and served as the corresponding author.

\pagestyle{plain}
\section*{Funding}
This research was supported by the National Natural Science Foundation of China (Grant No. 42505143), the AI for Science Program, Shanghai Municipal Commission of Economy
and Informatization (Grant No. 2025-GZL-RGZN-BTBX-02017), and the National Natural Science Foundation of China (U2542210).
\section*{Competing interest statement}
The authors declare no competing interests.

\clearpage
\putbib[main]
\end{bibunit}


\clearpage
\pagestyle{plain}
\setcounter{page}{1}
\renewcommand{\thepage}{S\arabic{page}}
\setcounter{section}{0}
\setcounter{subsection}{0}
\setcounter{subsubsection}{0}
\setcounter{figure}{0}
\setcounter{table}{0}
\setcounter{equation}{0}
\renewcommand{\thesection}{S\arabic{section}}
\renewcommand{\thesubsection}{S\arabic{section}.\arabic{subsection}}
\renewcommand{\thesubsubsection}{S\arabic{section}.\arabic{subsection}.\arabic{subsubsection}}
\renewcommand{\thefigure}{S\arabic{figure}}
\renewcommand{\thetable}{S\arabic{table}}
\renewcommand{\theequation}{S\arabic{equation}}
\renewcommand{\theHsection}{supp.\arabic{section}}
\renewcommand{\theHsubsection}{supp.\arabic{section}.\arabic{subsection}}
\renewcommand{\theHsubsubsection}{supp.\arabic{section}.\arabic{subsection}.\arabic{subsubsection}}
\renewcommand{\theHfigure}{supp.\arabic{figure}}
\renewcommand{\theHtable}{supp.\arabic{table}}
\renewcommand{\theHequation}{supp.\arabic{equation}}
\captionsetup[figure]{name=Supplementary Fig.}
\makeatletter
\def\@extra@binfo{@supp}
\def\@extra@b@citeb{@supp}
\makeatother
\begin{bibunit}[sn-mathphys]

\begin{center}
{\LARGE\bfseries Supplementary Information for:\par}
\vspace{0.75em}
{\Large\bfseries 3DTCR: Vortex-following generative reconstruction improves tropical cyclone intensity prediction from global AI weather forecasts\par}
\vspace{1em}
{\large Jun Liu, Xiaohui Zhong, Kai Zheng, Jiarui Li, Yifei Li, Tao Zhou, Wenwen Zhang, and Hao Li\par}
\vspace{0.75em}
{\normalsize Corresponding authors: Xiaohui Zhong (zhong\_xiaohui@fudan.edu.cn) and Hao Li (lihao\_lh@fudan.edu.cn)\par}
\end{center}

\section*{Contents of this file}
\begin{itemize}
  \item[] Supplementary Notes 1 to \total{section}
  \item[] Supplementary Tables 1 to \total{table}
  \item[] Supplementary Figures 1 to \total{figure}
\end{itemize}
\addtocontents{toc}{\protect\setcounter{tocdepth}{3}}
\clearpage
\setcounter{tocdepth}{3}
\tableofcontents
\clearpage

\section*{Supplementary Notes}
\section{Supplementary background and data}

This section first outlines the motivation for three-dimensional TC reconstruction and describes the roles of ERA5 and FuXi-2.0. It then details the construction and preparation of the WRF 3-km reference dataset. Finally, it presents the analyses used to assess whether the WRF 3-km fields can serve as high-resolution targets for TC-centered reconstruction.

\subsection{Motivation for three-dimensional TC reconstruction}

\begin{figure*}[!htp]
    \centering
    \includegraphics[width=\linewidth]{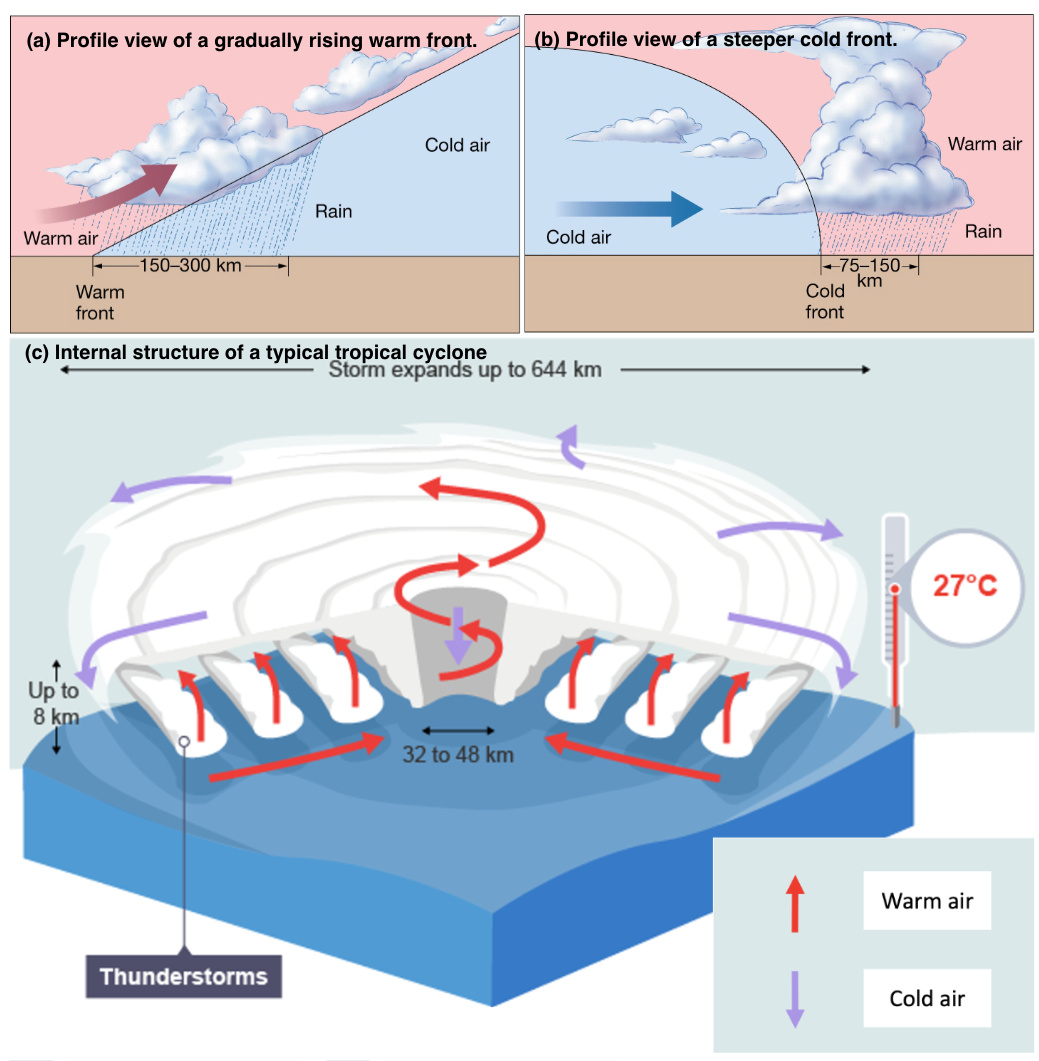}
    \caption{\textbf{Atmospheric frontal systems and typical tropical cyclone structure.}
    (\textbf{a}) Cross section through a gradually ascending warm front generating precipitation through rising warm air masses.
    (\textbf{b}) Cross section through a steep cold front generating precipitation through rising warm air masses.
    (\textbf{c}) Three-dimensional cutaway diagram of a tropical cyclone revealing its internal components, structural organization, and wind circulation patterns.}
    \label{fig:3d-structures}
\end{figure*}

Tropical cyclones (TCs) combine compact horizontal scales with a vertically organized circulation, making their inner-core structure difficult to represent in coarse global fields. A mature vortex includes an eye, an intense eyewall, spiral rainbands, a warm core, inflow at lower levels, and outflow at upper levels (Supplementary Fig.~\ref{fig:3d-structures}). These components vary with height and azimuth and produce sharp gradients in wind, pressure, temperature, and moisture, whereas coarse-resolution fields smooth these gradients and weaken localized extremes. This limitation motivates the TC-centered design of the WRF 3-km reference dataset and 3DTCR: the framework focuses high-resolution learning on the moving inner-core region so that it can recover fine-scale structure while preserving the surrounding circulation and the vertical coherence of the vortex.

Supplementary Fig.~\ref{fig:3d-structures} places the TC schematic alongside frontal cross sections as general examples of vertically organized atmospheric structure; panel \textbf{c} shows the TC-specific structure relevant to 3DTCR.

\FloatBarrier

\subsection{ERA5 reanalysis and FuXi-2.0 forecasts}

ERA5 and FuXi-2.0 serve distinct roles in the framework. ERA5 provides the initial and lateral boundary conditions for the WRF simulations and the input fields used during pretraining, whereas FuXi-2.0 provides the forecast inputs used during fine-tuning and reconstruction.

\subsubsection{ERA5 initial and boundary conditions}
\label{ECMWF Reanalysis v5 (ERA5)}
This study employs the ERA5 (European Centre for Medium-Range Weather Forecasts Reanalysis v5) dataset from the European Centre for Medium-Range Weather Forecasts as the driving data for the WRF model. ERA5 provides global atmospheric reanalysis products at $0.25^\circ$ spatial resolution and 6 h temporal intervals, representing the best estimate of historical atmospheric states. By integrating observations from multiple sources, including satellite retrievals, radiosonde measurements, and surface observations, ERA5 accurately characterizes large-scale environmental fields and serves as a standard data source widely adopted as a training target for AI-based weather prediction models. ERA5 provides reliable initial and lateral boundary conditions for the high-resolution dynamical downscaling simulations in this study.

\subsubsection{FuXi-2.0 inputs for fine-tuning and reconstruction}
\label{sec:fuxi_data}
In this study, FuXi-2.0 forecasts provide the operational forecast inputs used for supervised fine-tuning and vortex-following reconstruction, while the corresponding WRF 3-km fields serve as high-resolution reference targets for reconstruction. FuXi-2.0 delivers global weather forecasts at 1-h temporal resolution,
encompassing a comprehensive suite of meteorological variables: five
upper-atmospheric variables (geopotential, temperature, humidity, and
U/V wind components) across 13 standard pressure levels (50, 100, 150, 200, 250, 300, 400, 500, 600, 700, 850, 925, and 1000~hPa), together with 23 surface variables,
such as 10-m wind speed, 2-m temperature, mean sea level pressure,
100-m wind speed, downward surface solar radiation flux, sea surface
temperature, significant wave height, wave direction, and wave period. From this broader product, 3DTCR uses the 21 atmospheric fields defined in Section~\ref{Detailed Description of WRF Post-Processed and Model Training Variables}. Forecast inputs and their WRF targets are paired at 6 h intervals for the lead times evaluated in this study.

\subsection{Construction of the WRF 3-km reference dataset}
\label{WRF Simulation and Configuration Parameters}
We generated the WRF 3-km reference dataset by conducting high-resolution simulations with version 4.6 of the Weather Research and Forecasting (WRF) model configured with the Advanced Research WRF (ARW) dynamical core, using ERA5 reanalysis as the initial and lateral boundary conditions. The simulations use a configuration with one domain, 3-km horizontal grid spacing on a lat-lon projection (map\_proj = `lat-lon'), 51 vertical $\sigma$-levels, and a model time step of 15~s \cite{skamarock2019description,fierro2009impact}. The lat-lon projection maintains consistency with global latitude-longitude products such as ERA5 and FuXi and simplifies subsequent interpolation and comparison. The physical parameterization schemes include the WRF Single-Moment 6-class microphysics scheme (mp\_physics = 6) for cloud microphysical processes, the Yonsei University (YSU) planetary boundary layer scheme (bl\_pbl\_physics = 1) for turbulent boundary layer characteristics \cite{hong2006new}, and the Rapid Radiative Transfer Model (RRTM) longwave radiation scheme (ra\_lw\_physics = 1) coupled with the RRTMG shortwave radiation scheme (ra\_sw\_physics = 4) for radiative transfer calculations \cite{guzman2021global,hong2006wrf,emanuel2005increasing,osuri2012customization}. Detailed configuration parameters of the WRF simulations are summarized in Table~\ref{tab:wrf_config}. To ensure numerical stability, a 12-h spin-up period was implemented prior to each simulation to reduce spurious oscillations arising from initial field imbalances. During extended integrations, repeated restarts were used to control error accumulation, while spectral nudging was applied to constrain large-scale wave numbers and prevent the simulated circulation from diverging from the ERA5 analysis fields.

\begin{table}[!htp]
\centering
\small
\caption{WRF configuration parameters.}
\label{tab:wrf_config}
\begin{tabularx}{\linewidth}{p{0.28\linewidth}p{0.22\linewidth}X}
\toprule
\textbf{Configuration Parameter} & \textbf{Value/Scheme} & \textbf{Description} \\
\midrule
Model Version & WRF-ARW 4.6 & Weather Research and Forecasting model \\
Domain Configuration & Single domain & - \\
Horizontal Resolution & 3 km & Grid resolution \\
Map Projection & lat-lon & Latitude-longitude projection (map\_proj = `lat-lon') \\
Vertical Levels & 51 $\mathrm{\sigma}$-levels & Sigma coordinate system \\
Time Integration Step & 15 seconds & Temporal resolution \\
Microphysics & WSM6 (mp\_physics = 6) & WRF Single-Moment 6-class scheme \\
Planetary Boundary Layer & YSU (bl\_pbl\_physics = 1) & Yonsei University scheme \\
Longwave Radiation & RRTM (ra\_lw\_physics = 1) & Rapid Radiative Transfer Model \\
Shortwave Radiation & RRTMG (ra\_sw\_physics = 4) & RRTMG shortwave scheme \\
Spin-up Period & 12 hours & Model initialization time \\
Boundary Forcing Update & 6 hours (interval\_seconds = 21600) & ERA5 reanalysis data frequency \\
Driving Data & ERA5 reanalysis & Meteorological forcing data \\
Sea Surface Temperature & Real-time updates enabled & Dynamic SST updating \\
Nudging Coefficient & 0.00003 & Weak large-scale constraint for mesoscale development \\
Nudging & Spectral nudging & Large-scale wavenumber constraint \\
\bottomrule
\end{tabularx}
\end{table}

To capture the fine-scale evolution of TC inner-core structure, the reference latitude and longitude of the computational domain were updated during the restart workflow so that the TC remained within the regional domain. Throughout the simulation period, lateral boundary forcing was updated every 6 hours (interval\_seconds = 21600). Time-varying sea surface temperature provided the lower boundary condition. A comprehensive overview of the simulation framework, including the spatial domain configuration, numerical workflow, and representative three-dimensional atmospheric fields, is presented in Supplementary Fig.~\ref{run_datassts}.

\begin{figure*}[!htp]
    \centering
    \includegraphics[width=\linewidth]{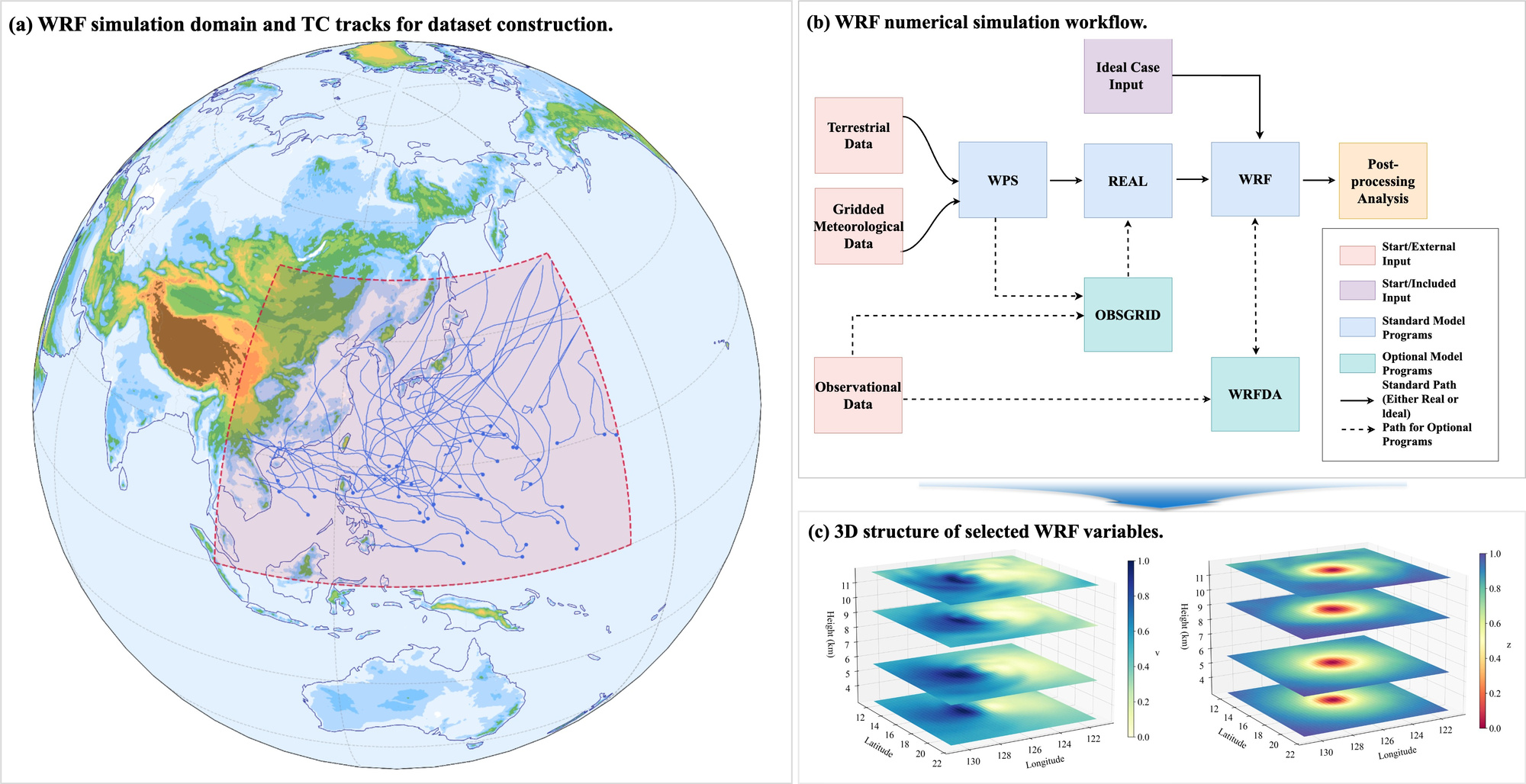}
    \caption{\textbf{WRF simulation framework and dataset construction.} (\textbf{a}) WRF simulation domain (pink shaded area) and tropical cyclone tracks used for dataset construction. Historical TC trajectories are overlaid to show the spatial coverage of extreme weather events within the modeling domain.
(\textbf{b}) WRF numerical simulation workflow. The processing chain includes data ingestion from terrestrial and gridded meteorological sources, preprocessing via WPS, model initialization through REAL, WRF simulation, and post-processing analysis.
(\textbf{c}) Multilevel vertical structure of selected meteorological variables in the WRF dataset. Stacked horizontal slices show how the spatial distributions of these variables vary with height.}
    \label{run_datassts}
\end{figure*}

\FloatBarrier

\subsection{3DTCR training data preparation}
\label{Detailed Description of WRF Post-Processed and Model Training Variables}
Based on the WRF model configuration and post-processing pipeline described above,
the complete dataset encompasses a total of 93 variables organized into
three categories: upper-air variables, surface variables, and diagnostic variables,
as summarized in  Supplementary Table~\ref{tab:variable_info}. Specifically, the upper-air category
includes six key variables: geopotential ($\mathrm{Z}$), temperature ($\mathrm{T}$),
u-wind ($\mathrm{U}$), v-wind ($\mathrm{V}$), relative humidity ($\mathrm{R}$), and specific humidity ($\mathrm{Q}$).
Each of these is provided at 13 standard pressure levels (50, 100, 150, 200, 250, 300, 400, 500, 600, 700, 850, 925, and 1000~hPa),
constituting 78 upper-air variables in total. The remaining variables consist of surface
and diagnostic fields, such as 2-m temperature ($\mathrm{T2M}$), 10-m winds
($\mathrm{U10M}$, $\mathrm{V10M}$), mean sea level pressure ($\mathrm{MSL}$), and total
precipitation ($\mathrm{TP}$). These 93 variables collectively provide comprehensive
multilevel coverage of the three-dimensional atmospheric state.

From the complete set of post-processed variables described above, 21 variables most closely associated with the dynamical and thermodynamic structure of TCs are selected as the training targets for the 3DTCR model. Specifically, these comprise four
upper-air variables, namely geopotential ($\mathrm{Z}$), temperature ($\mathrm{T}$),
and u/v wind components ($\mathrm{U}$/$\mathrm{V}$), which are each resolved
at four pressure levels (200, 500, 700, and 850~hPa). This yields 16 upper-air
fields in total, which are supplemented by five surface variables: mean sea level pressure
($\mathrm{MSL}$), 2-m temperature ($\mathrm{T2M}$), 10-m u/v wind
components ($\mathrm{U10M}$, $\mathrm{V10M}$), and 10-m wind speed
($\mathrm{WS10M}$). Together, these 21 variables enable a systematic and
multilevel characterization of the three-dimensional vertical structure of TCs.

\begin{table}[htp!]
\centering
\small
\caption{Complete variable information for the WRF post-processing dataset.}
\label{tab:variable_info}
\begin{tabularx}{\linewidth}{p{0.28\linewidth} X p{0.18\linewidth}}
\toprule
\textbf{Variable Category} & \textbf{Variable Name} & \textbf{Abbreviation} \\
\midrule
Upper-air Variables & Geopotential & Z \\
 & Temperature & T \\
 & U component of wind & U \\
 & V component of wind & V \\
 & Relative humidity & R \\
 & Specific humidity & Q \\
\midrule
Surface Variables & Mean sea level pressure & MSL \\
 & 2-m temperature & T2M \\
 & Total precipitation & TP \\
 & 10-m U wind component & U10M \\
 & 10-m V wind component & V10M \\
\midrule
Diagnostic Variables & Wind speed at 200 hPa & WS200 \\
 & Wind speed at 500 hPa & WS500 \\
 & Wind speed at 700 hPa & WS700 \\
 & Wind speed at 850 hPa & WS850 \\
 & Vorticity at 200 hPa & VOR200 \\
 & Vorticity at 500 hPa & VOR500 \\
 & Vorticity at 700 hPa & VOR700 \\
 & Vorticity at 850 hPa & VOR850 \\
 & 10-m vorticity & VOR10M \\
 & 10-m wind speed & WS10M \\
\bottomrule
\end{tabularx}
\end{table}

\FloatBarrier

\subsection{Assessment of the WRF reference dataset}

The following analyses complement the assessment of the reference dataset in the main text in two ways: by examining Typhoon DOKSURI as a representative case and by comparing WS10M distributions across sampling domains. Notably, the WRF simulations serve as high-resolution reference targets for reconstruction rather than as observations.

\subsubsection{Reference assessment for Typhoon DOKSURI}
\label{Validation of the Simulated TC Dataset: A Case Study of Typhoon DOKSURI}
To complement the statistical evaluation presented in the main text, we assess the WRF 3-km reference dataset using Typhoon DOKSURI at 1200 UTC 24 July 2023. The corresponding figure in the main text compares the distributions of 10-m wind speed (WS10M) for DOKSURI across FuXi, ERA5, and WRF 3-km (panels \textbf{a}, \textbf{c}, and \textbf{e}, respectively), together with maximum WS10M evaluated against IBTrACS best-track intensity estimates over the full 2024 western North Pacific testing dataset (panels \textbf{b}, \textbf{d}, and \textbf{f}).

The DOKSURI spatial fields highlight a fundamental limitation of coarse-resolution global products. Both FuXi (panel \textbf{a}) and ERA5 (panel \textbf{c}) exhibit excessively smoothed inner-core wind structures, with peak winds confined to a broad and weak ring around the eye. The sharp radial gradients and localized wind maxima required to sustain extreme typhoon intensity are largely absent in both products. In contrast, the WRF 3-km field (panel \textbf{e}) successfully resolves a tight, asymmetric eyewall with pronounced localized extreme wind maxima, together with finer-scale features such as spiral rainbands and asymmetric inflow structures.

These structural differences are quantitatively reflected in the statistical scatter plots. Both FuXi (panel \textbf{b}) and ERA5 (panel \textbf{d}) exhibit pronounced negative biases and elevated RMSE at high observed wind speeds, with predicted maxima plateauing well below the 1:1 reference line. WRF 3-km (panel \textbf{f}) substantially alleviates this underestimation, with both the bias and RMSE markedly reduced and predicted maxima distributed closer to the 1:1 reference across the full intensity spectrum.

\begin{figure*}[!htp]
    \centering
    \includegraphics[width=0.90\linewidth]{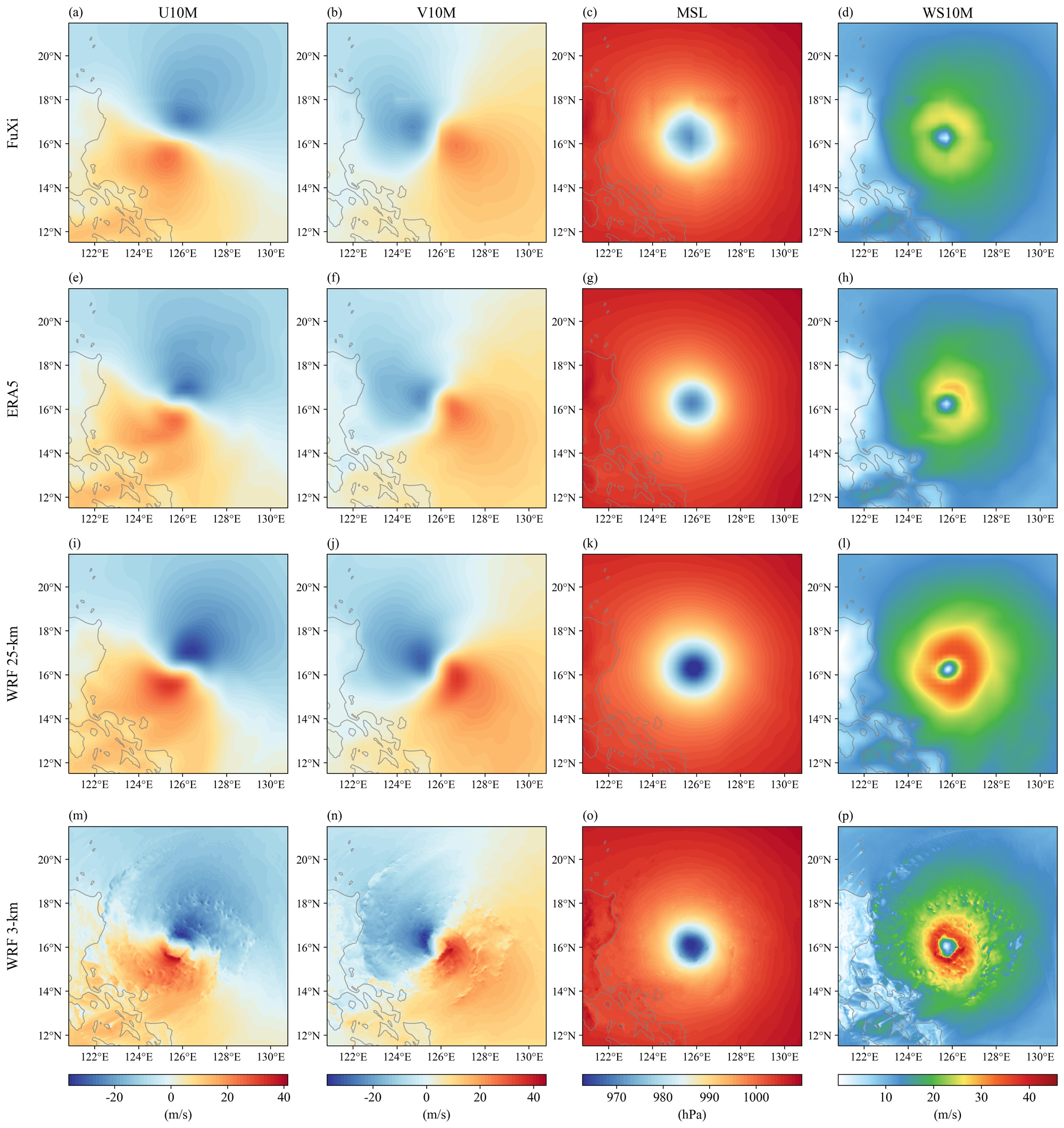}
    \caption{\textbf{Comparison of tropical cyclone fields across datasets and resolutions.} Spatial distribution of key meteorological variables for Typhoon DOKSURI at 1200 UTC on 24 July 2023. The rows (from top to bottom) compare fields from FuXi, ERA5, the WRF 25-km simulation, and the WRF 3-km simulation. The columns show: (a, e, i, m) 10-m U wind component (U10M, m/s), (b, f, j, n) 10-m V wind component (V10M, m/s), (c, g, k, o) mean sea level pressure (MSL, hPa), and (d, h, l, p) 10-m wind speed (WS10M, m/s). The black contour lines represent coastlines. The WRF 3-km simulation displays finer inner-core wind features and sharper pressure gradients than FuXi, ERA5, and the WRF 25-km configuration. Color scales are indicated at the bottom of each column.}
    \label{fig:DOKSURI}
\end{figure*}

Supplementary Fig.~\ref{fig:DOKSURI} extends this assessment through a field comparison at the same valid time. It presents U10M, V10M, MSL, and WS10M from FuXi, ERA5, WRF 25-km, and WRF 3-km side by side, allowing differences in the representation of near-surface TC structure across datasets and resolutions to be examined directly.

Together, the evaluation in the main text and the supplementary field comparison support the use of the WRF 3-km fields as high-resolution reference targets for reconstruction with 3DTCR. The remaining intensity underestimation identified in the main text nevertheless remains a limitation of this reference dataset.

\FloatBarrier

\subsubsection{Wind speed distributions across sampling domains}
\label{sec:ws10m_frequency_multiscale}

\begin{figure*}[!htp]
    \centering
    \includegraphics[width=1.0\linewidth]{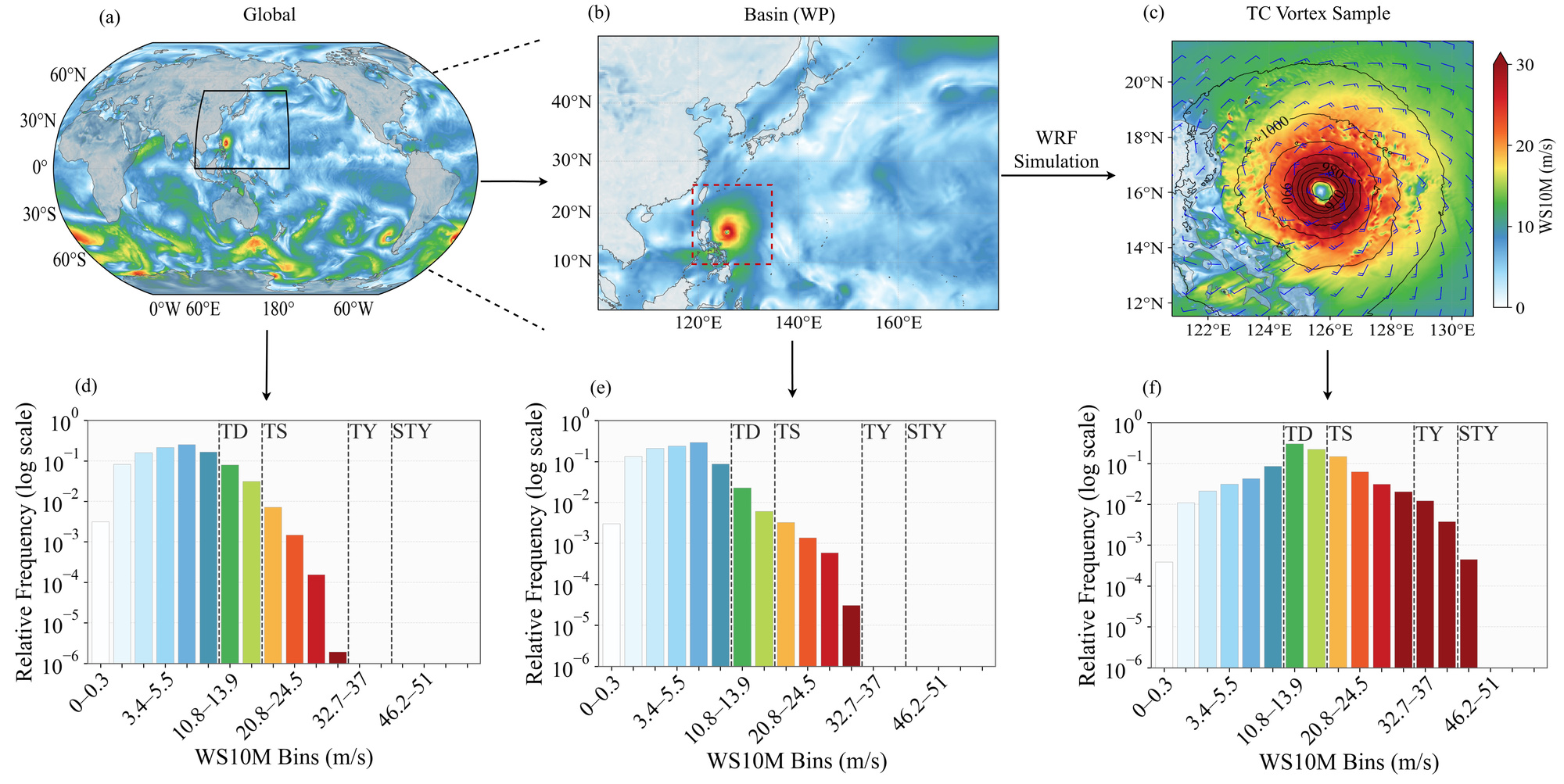}
    \caption{\textbf{Comparison of 10-m wind speed (WS10M) relative frequencies across wind speed bins at different spatial scales.} Panels (\textbf{a}--\textbf{c}) show the field at the global scale, the field over the western North Pacific basin, and the 10$^\circ$ $\times$ 10$^\circ$ TC vortex sample at 3 km resolution, respectively. Panels (\textbf{d}--\textbf{f}) show the corresponding WS10M relative frequencies across different wind speed bins. For each scale, relative frequency is the fraction of grid points within the corresponding sampling domain that fall in each WS10M bin. Vertical dashed lines denote different TC intensity categories.}
    \label{fig:ws10m_frequency_multiscale}
\end{figure*}

The frequency distribution of 10-m wind speed exhibits a long tail that varies strongly with spatial scale (Supplementary Fig.~\ref{fig:ws10m_frequency_multiscale}). At the global scale (panels a, d), the spatial field is overwhelmingly dominated by weak to moderate background winds, while TC inner-core extremes occupy only a vanishingly small fraction of grid points; the resulting histogram decays by several orders of magnitude before reaching hurricane intensities, so extreme samples are statistically submerged in the bulk of the distribution. Restricting the domain to the western North Pacific (panels b, e) geographically concentrates TC activity, yet the field over the basin still contains extensive regions of quiescent monsoon trough and trade wind flow, and the upper tail remains sparse. It is only at the object level, within the $10^{\circ}$ vortex-following window at 3 km resolution (panels c, f), that the eyewall is resolved as a coherent population of high wind speeds and the upper tail becomes a substantive fraction of the empirical distribution.

This scale dependence has direct implications for AI-based TC forecasting. Standard global or basin-scale models trained on coarse reanalysis such as ERA5 at roughly 25 km inherit the same severe imbalance: the loss landscape is dominated by abundant weak wind samples, and gradient updates during training rarely encounter genuine extreme events. The resulting parameter estimates are biased toward climatology, which likely contributes to the systematic underestimation of TC intensity reported across recent AI weather models.

These sampling and computational constraints motivate two complementary components of 3DTCR. At the data level, the training samples are constructed in TC-centered coordinates, with each input and reference window centered on its own vortex within a $10^{\circ} \times 10^{\circ}$ domain at 3-km resolution. This sampling strategy increases the representation of intense inner-core fields while retaining the surrounding environment. At the reconstruction level, the vortex-following generative framework enables flexible high-resolution reconstruction over a region of interest that follows the moving TC. As summarized in Supplementary Table~\ref{tab:computational_cost}, 3DTCR completes a 5-day reconstruction in approximately 3 min on one NVIDIA A100 GPU, compared with approximately 12 h for a TC-centered WRF simulation and approximately 400 h for a full-basin WRF 3-km integration on 32 CPU cores. The data construction therefore mitigates the scarcity of extreme wind samples, while the vortex-following reconstruction provides an efficient alternative to basin-wide kilometer-scale simulation.

\FloatBarrier

\section{Supplementary methods}

This section supplements the framework description in the main text by detailing how TCs are located, how Rectified Flow generates the reconstruction, and how the network, center alignment during fine-tuning, lead time conditioning, latent space alignment, training objectives, and evaluation are implemented.

\subsection{Tropical cyclone tracking method}
\label{Tropical Cyclone Tracking Method}
\begin{figure}[!htbp]
    \centering
    \includegraphics[width=0.75\linewidth]{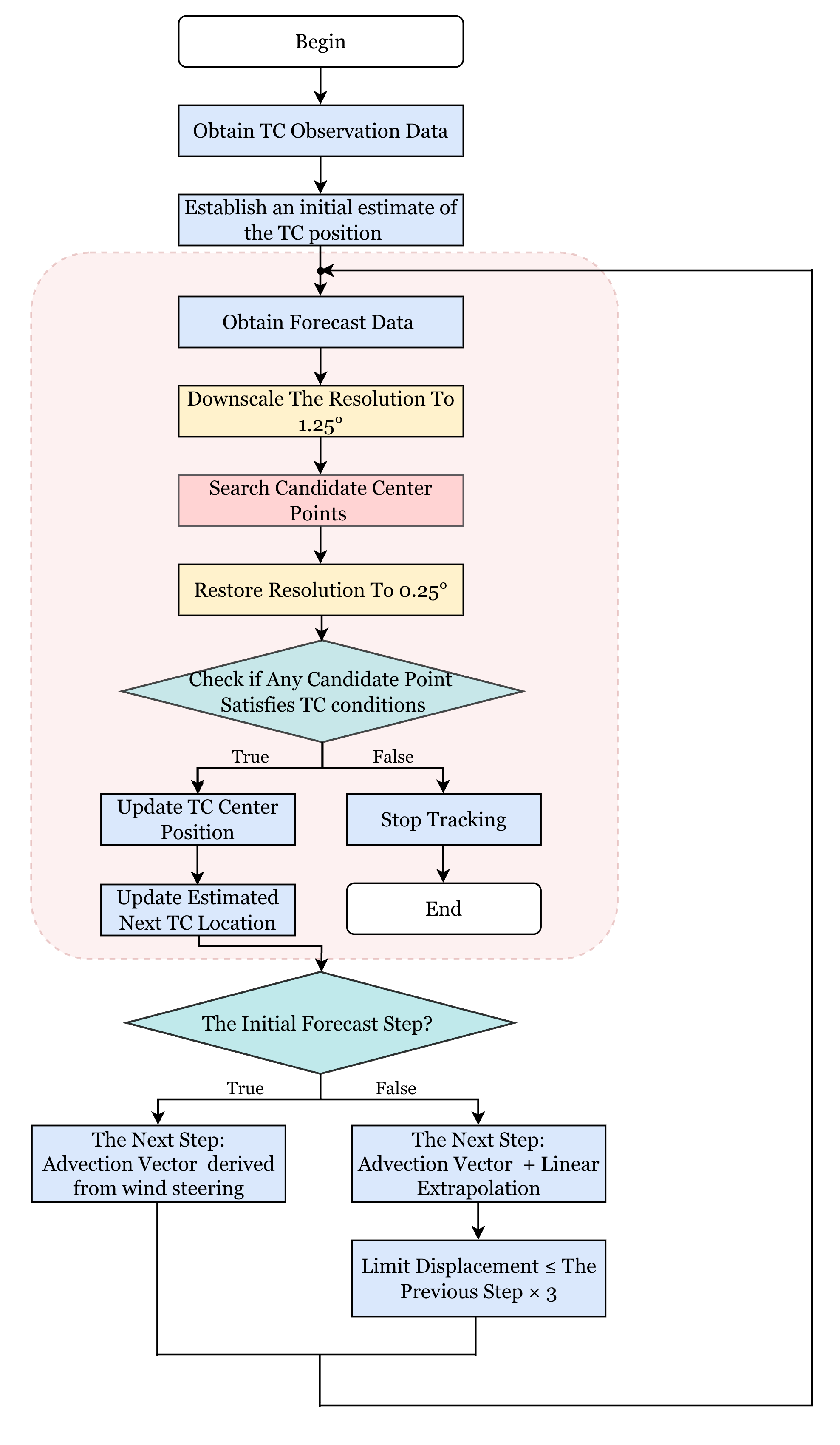}
    \caption{\textbf{Schematic of the TC tracking algorithm.}
    The algorithm first uses TC observations to determine an initial estimate of the TC position. It then coarsens the forecast data and searches for candidate TC centers. If a candidate satisfies the predefined TC criteria, the TC center is updated; otherwise, tracking stops. For subsequent tracking steps, an advection vector derived from the steering flow is used, either alone or in combination with linear extrapolation, depending on whether the current step is the first forecast step.}
    \label{tracker}
\end{figure}

The TC tracking method utilized in 3DTCR is detailed as follows. This scheme builds upon the ECMWF algorithm introduced by Zhong et al. \citep{zhong2024fuxi,zhang2015verification}. The process begins by establishing a baseline TC location derived from observational data. Within a search radius of 445 km from this initial point, the algorithm identifies potential TC centers by locating local minima in mean sea level pressure (MSL) alongside local maxima in 10-m wind vorticity. The algorithm then evaluates these candidates against a series of predefined thresholds (as shown in Supplementary Table~\ref{tab:tracking_config_parameters}) and designates the closest valid point as the true TC center. Should no candidate satisfy these requirements, the system assumes the TC has dissipated and terminates the tracking procedure (Supplementary Fig.~\ref{tracker}).

To predict the TC's position for the subsequent forecast step, the algorithm employs a hybrid estimation technique. This involves blending linear extrapolation calculated from the TC's coordinates in the previous two time steps with an advection vector derived from wind fields at a designated pressure level. This localization cycle is iteratively applied across all forecast steps.

For the initial forecast step, since only one historical location is available,
the anticipated position is calculated using the advection vector method alone.
In addition, AI-based forecasts often exhibit inherent spatial smoothness that
can generate spurious local minima. To address this, the forecast fields are
coarsened from $0.25^\circ$ to $1.25^\circ$ during the candidate search phase, and subsequently restored to their original
resolution. Finally, to prevent unrealistic positional jumps, the algorithm
restricts any positional shift to a maximum of three times the displacement
observed in the preceding forecast time step.


\begin{table}[!h]
\centering
\caption{Parameters and thresholds related to the TC tracking.}
\label{tab:tracking_config_parameters}
\begin{tabularx}{\textwidth}{cXXXX}
\hline
\textbf{Parameter} & \textbf{Level} & \textbf{Radius} & \textbf{Threshold} & \textbf{Comments} \\
\hline
$\mathrm{WS10M}$ & Surface & 278 km & $>8$~m/s & Only required over land. \\
Vorticity & 850 hPa & 278 km & absolute value $\geq 5 \times 10^{-5} s^{-1}$ \\
\makecell{Geopotential \\ thickness} & 850 hPa and 200 hPa & 278 km & & A maximum in thickness is necessary only after the TC has transitioned into an extratropical cyclone. \\
\hline
\end{tabularx}
\end{table}

\subsection{Rectified Flow formulation}
\label{Rectified Flow}
Given a high-resolution atmospheric state distribution $\mathrm{\pi}_1$ and coarse-resolution fields as conditional input $\mathrm{c}$, our objective is to learn a transport map $\mathrm{T}: \mathbb{R}^\mathrm{d} \rightarrow \mathbb{R}^\mathrm{d}$ that generates high-resolution forecasts $\mathrm{X}_1 \sim \mathrm{\pi}_1$ from standard Gaussian noise $\mathrm{X}_0 \sim \mathrm{N}(\mathrm{0}, \mathrm{I})$. We adopt Rectified Flow \cite{liu2022flow,esser2024scaling,fotiadis2025adaptive}, an ordinary differential equation (ODE) framework that learns to transport distributions along straight-line paths.

Rectified Flow defines the transport process by constructing a conditional time-dependent velocity field
  $\mathrm{v}_{\theta}(\mathrm{X}_t,\mathrm{c},t,K)$. Formally, we define a time-dependent flow $\{\mathrm{X}_\mathrm{t} : \mathrm{t} \in [0,1]\}$ that evolves according to the ODE:
\begin{equation}
\frac{\mathrm{d} \mathrm{X}_\mathrm{t}}{\mathrm{d} \mathrm{t}} = \mathrm{v}_{\mathrm{\theta}}(\mathrm{X}_\mathrm{t}, \mathrm{c}, \mathrm{t},\mathrm{K}), \quad \mathrm{X}_0 \sim \mathrm{N}(\mathrm{0}, \mathrm{I}).
\end{equation}
During training, $\mathrm{v}_{\mathrm{\theta}}$ is trained to match the direction from $\mathrm{X}_0$ to $\mathrm{X}_1$. Specifically, for each training sample, we construct the interpolation:
\begin{equation}
\mathrm{X}_\mathrm{t} = (1 - \mathrm{t}) \mathrm{X}_0 + \mathrm{t} \mathrm{X}_1, \quad \mathrm{t} \sim \mathrm{Uniform}([0, 1]).
\end{equation}
The training objective minimizes the discrepancy between the predicted and true velocity:
\begin{equation}
    \mathrm{L}_{\mathrm{RF}}(\mathrm{\theta}) =
    \mathbb{E}_{\mathrm{t}, \mathrm{X}_0, \mathrm{X}_1, \mathrm{c}, \mathrm{K}}
    \left[
    \left\|
    \mathrm{v}_{\mathrm{\theta}}
    (\mathrm{X}_\mathrm{t}, \mathrm{c}, \mathrm{t}, \mathrm{K})
    - (\mathrm{X}_1 - \mathrm{X}_0)
    \right\|_2^2
    \right].
\end{equation}

This approach ensures marginal distribution preservation at all time steps, achieves provably lower convex transport costs than arbitrary couplings, and produces increasingly straight trajectories that enhance computational efficiency.

In our implementation, the velocity field takes the noisy state $\mathrm{X}_\mathrm{t}$ and coarse-resolution fields $\mathrm{c}$ as inputs, with forecast lead time $\mathrm{K}$ provided as an additional conditioning variable, modeled through a U-Net architecture:
\begin{equation}
    \mathrm{v}_{\mathrm{\theta}}(\mathrm{X}_\mathrm{t}, \mathrm{c}, \mathrm{t},\mathrm{K}) = \text{U-Net}_{\mathrm{\theta}}([\mathrm{c}, \mathrm{X}_\mathrm{t}], \mathrm{t},\mathrm{K}).
\end{equation}
 Here, $t\in[0,1]$ denotes the continuous flow time, whereas
  $K\in\{0,\ldots,20\}$ denotes the discrete forecast lead time index.

During sampling, we solve the learned ODE forward in time using the Euler method with $\mathrm{N}$ discretization steps:
\begin{equation}
    \mathrm{X}_{\mathrm{t}+\Delta \mathrm{t}} = \mathrm{X}_\mathrm{t} + \Delta \mathrm{t} \cdot \mathrm{v}_{\mathrm{\theta}}(\mathrm{X}_\mathrm{t}, \mathrm{c}, \mathrm{t},\mathrm{K}), \quad \Delta \mathrm{t} = \frac{1}{\mathrm{N}}.
\end{equation}
Starting from \ensuremath{\mathbf{x}_0}, the state is iteratively updated at uniformly spaced time steps over \ensuremath{t \in [0,1]}. Unlike DDPM \cite{ho2020denoising} and DDIM \cite{song2020denoising}, which rely on a predefined noise schedule, Rectified Flow directly learns the transport velocity field, thereby simplifying the training objective and enabling efficient deterministic sampling.

\begin{algorithm}[H]
\caption{Conditional Rectified Flow training}
\label{alg:rectified_flow}
\begin{algorithmic}[1]
\State \textbf{Input:} Dataset \ensuremath{\mathrm{D} = \{(\mathbf{c}, \mathbf{x}_1, K)\}}, velocity model \ensuremath{v_{\boldsymbol{\theta}}}
\State \hspace{1.3cm} where \ensuremath{\mathbf{c}} is the coarse field, \ensuremath{\mathbf{x}_1} is the WRF 3-km fine-scale field, and \ensuremath{K} is the forecast lead time index
\State Initialize \ensuremath{\boldsymbol{\theta}} and the optimizer
\State \textbf{repeat}
\State \hspace{0.5cm} Sample \ensuremath{(\mathbf{c}, \mathbf{x}_1, K)} from \ensuremath{\mathrm{D}}
\State \hspace{0.5cm} Sample \ensuremath{t \sim \mathrm{Uniform}(0, 1)} and \ensuremath{\mathbf{x}_0 \sim \mathrm{N}(\mathbf{0}, \mathbf{I})}
\State \hspace{0.5cm} Resize \ensuremath{\mathbf{c}} to the target resolution
\State \hspace{0.5cm} Compute the flow interpolation:
\State \hspace{1.5cm} \ensuremath{\mathbf{x}_t = t\mathbf{x}_1 + (1-t)\mathbf{x}_0}
\State \hspace{0.5cm} Predict the velocity:
\State \hspace{1.5cm} \ensuremath{\mathbf{v}_{\mathrm{pred}}
= v_{\boldsymbol{\theta}}
\bigl(\mathrm{Concat}(\mathbf{c}, \mathbf{x}_t), t, K\bigr)}
\State \hspace{0.5cm} Update \ensuremath{\boldsymbol{\theta}} via gradient descent:
\State \hspace{1.5cm} \ensuremath{\boldsymbol{\theta}
\leftarrow \arg\min_{\boldsymbol{\theta}}
\mathbb{E}\!\left[
\left\|(\mathbf{x}_1-\mathbf{x}_0)-\mathbf{v}_{\mathrm{pred}}\right\|_2^2
\right]}
\State \textbf{until} convergence
\end{algorithmic}
\end{algorithm}

\begin{algorithm}[H]
\caption{Sample generation}
\label{alg:sampling}
\begin{algorithmic}[1]
\State \textbf{Input:} Coarse field $\mathbf{c}$, forecast lead time index $K$, trained velocity field $v_{{\boldsymbol{\theta}}}$, number of steps $N$
\State \textbf{Output:} WRF 3-km fine field $\mathbf{x}_1$
\State \textbf{Initialize:} $\mathbf{x}_0 \sim \mathrm{N}(\mathbf{0}, \mathbf{I})$, $\Delta t \leftarrow 1/N$
\State Resize $\mathbf{c}$ to target resolution
\For{$t = 0$ to $1$ step $\Delta t$}
    \State Predict velocity:
    \State \hspace{1cm} $\mathbf{v}_t = v_{{\boldsymbol{\theta}}}(\text{Concat}(\mathbf{c}, \mathbf{x}_t), t, K)$
    \State Euler solver:
    \State \hspace{1cm} $\mathbf{x}_{t+\Delta t} = \mathbf{x}_t + \mathbf{v}_t \cdot \Delta t$
\EndFor
\State Set $\mathbf{x}_1 \leftarrow \mathbf{x}_t$
\State Resize $\mathbf{x}_1$ to original resolution
\State \textbf{return} $\mathbf{x}_1$
\end{algorithmic}
\end{algorithm}

\subsection{Model architecture and training}
\label{app:training_details}

This section provides additional details on the 3DTCR network architecture, center alignment during fine-tuning, forecast lead time conditioning, MMD-based latent space alignment, two-stage training objectives, and implementation settings. These descriptions complement the framework overview in the main text and provide the information needed to reproduce the model.

\subsubsection{Network architecture}
\label{app:network_details}

3DTCR is a vortex-following generative framework built on Rectified Flow. As described in the main text, the velocity field estimator \( \mathrm{v}_\theta \) is parameterized by a symmetric U-Net backbone. At each flow time, the U-Net predicts the conditional velocity that advances the current state from Gaussian noise toward the target atmospheric state.

The network contains six encoder-decoder levels with channel dimensions \((128, 128, 256, 256, 512, 512)\). The encoder progressively aggregates large-scale contextual information through convolutional downsampling, whereas the decoder restores spatial resolution through symmetric upsampling and predicts the conditional velocity field at the target resolution. The final high-resolution atmospheric state is obtained by integrating the learned velocity field with the Euler solver described in Section~\ref{Rectified Flow}. This hierarchical design enables the velocity estimator to combine broad environmental context with localized spatial detail.

Attention modules are inserted into the deeper layers at low resolution, where dependencies over long spatial distances can be modeled more efficiently. The attention head dimension is fixed at \( \mathrm{d}_k = 8 \). The Sigmoid Linear Unit (SiLU) activation is used throughout the network to provide smooth nonlinear transformations and stable gradient propagation during training. Together, these design choices support the representation of both large-scale circulation and fine-scale vortex features.

\subsubsection{Spatial displacement and center alignment during fine-tuning}
\label{sec:ws10m_4src_displacement}

\begin{figure*}[!htbp]
    \centering
    \includegraphics[width=1.0\linewidth]{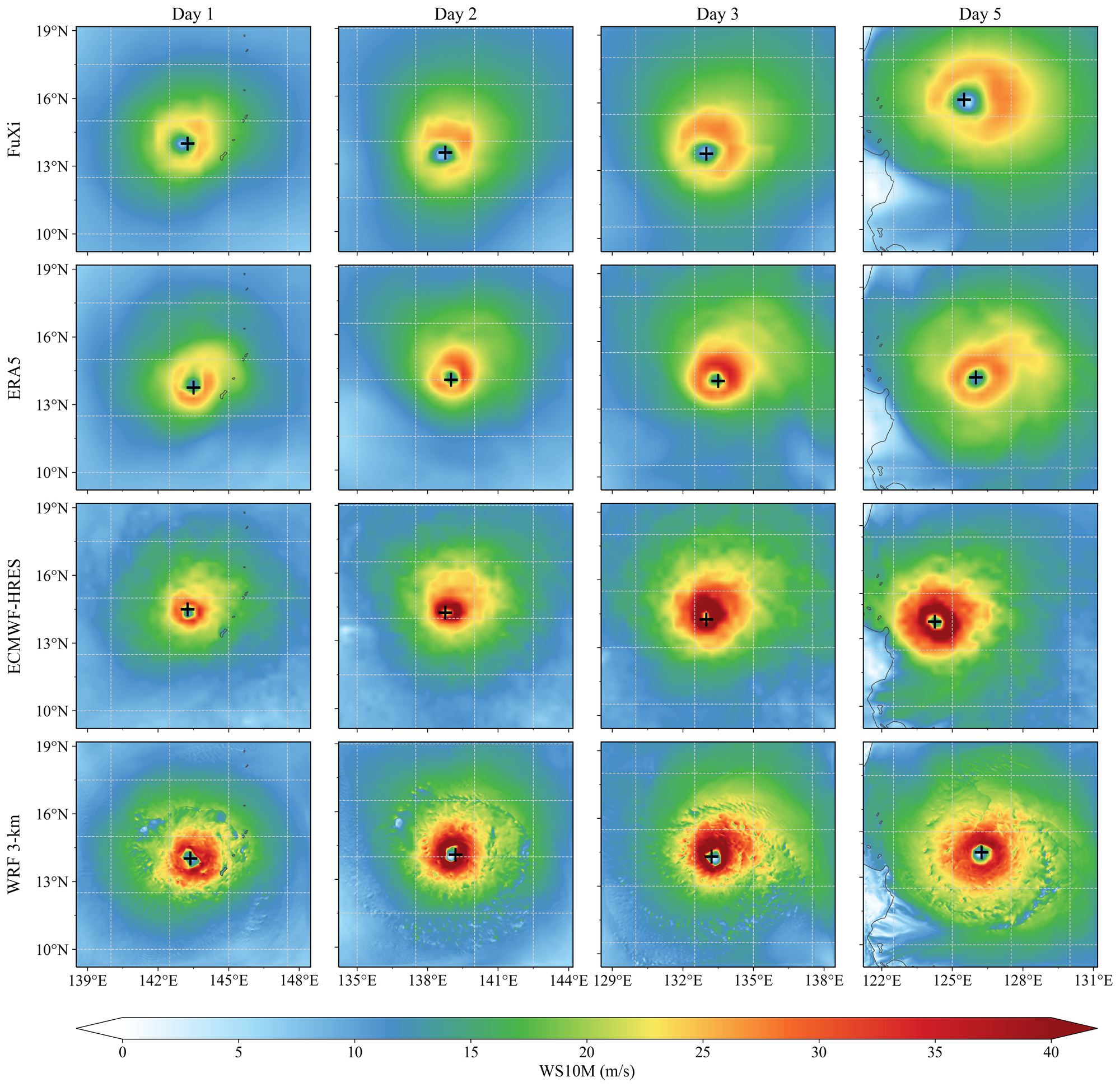}
    \caption{\textbf{Spatial displacement of 10-m wind speed (WS10M) for Typhoon MAWAR.}
    FuXi, ERA5, ECMWF-HRES, and WRF 3-km fields are shown for Days 1--3 and 5 from the forecast initialized at 0000 UTC 24 May 2023. All fields are displayed in the common WRF 3-km domain. Shading denotes WS10M, and black crosses indicate the TC center in each field, defined by the minimum mean sea level pressure.}
    \label{fig:ws10m_4src_4day_MAWAR_20230524}
\end{figure*}

\begin{figure*}[!htbp]
    \centering
    \includegraphics[width=1.0\linewidth]{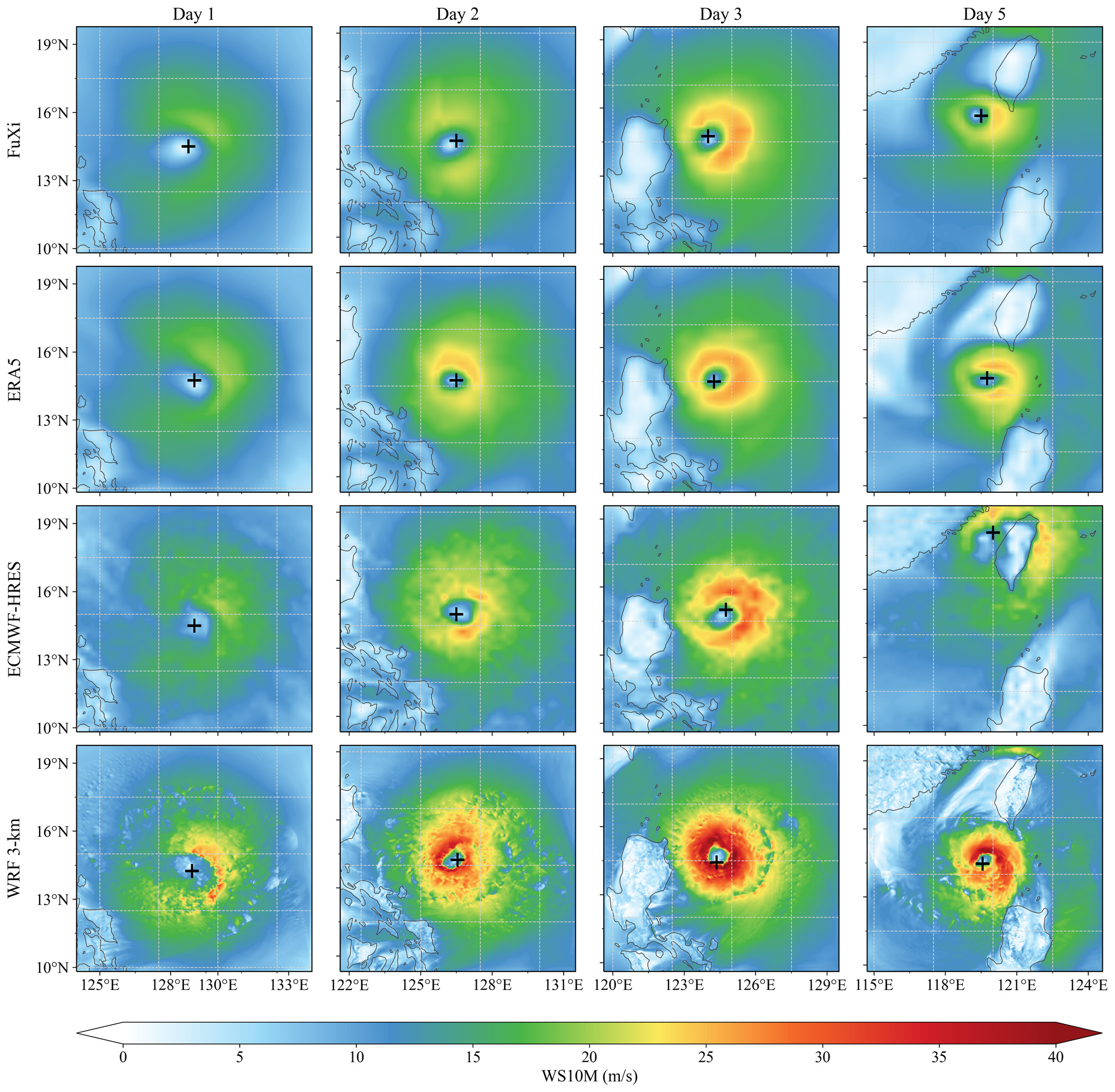}
    \caption{\textbf{Spatial displacement of 10-m wind speed (WS10M) for Typhoon DOKSURI.}
    FuXi, ERA5, ECMWF-HRES, and WRF 3-km fields are shown for Days 1--3 and 5 from the forecast initialized at 0000 UTC 22 July 2023. All fields are displayed in the common WRF 3-km domain. Shading denotes WS10M, and black crosses indicate the TC center in each field, defined by the minimum mean sea level pressure.}
    \label{fig:ws10m_4src_4day_DOKSURI_20230722}
\end{figure*}

As forecast lead time increases, accumulated errors in AI-based global atmospheric models such as FuXi progressively shift the predicted TC vortex away from its true position, a behavior we refer to as spatial pattern displacement. Supplementary Figs.~\ref{fig:ws10m_4src_4day_MAWAR_20230524} and \ref{fig:ws10m_4src_4day_DOKSURI_20230722} illustrate this effect by comparing 10-m wind fields from multiple data sources across successive lead times. Although the inner-core morphology remains broadly consistent at short ranges, the predicted TC center increasingly departs from the analyzed TC center with lead time. The same physical structure therefore appears at different grid locations across data sources.

Such displacement creates a fundamental challenge for reconstruction models trained with pixel-wise supervision on a fixed geographical grid. When the input and target vortices are offset, even a physically realistic reconstruction can be penalized twice: once because the vortex is not reconstructed at the target location, and again because it is reconstructed at a location where the target field is weak. This \enquote{double penalty} becomes more severe at longer lead times and encourages conservative, smoothed results rather than sharp vortex structure. The problem is most pronounced within the inner-core region, where strong wind and pressure gradients make positional errors particularly costly.

3DTCR handles this problem during supervised fine-tuning. In pretraining, ERA5 inputs and WRF 3-km targets are already close in position because the WRF simulations are driven by ERA5 and performed in regional domains centered on historical best-track positions. The mismatch becomes important only when FuXi forecasts are introduced. During fine-tuning, the input and target windows are extracted around their respective centers: the FuXi window uses the TC center identified in the FuXi forecast, whereas the WRF window uses the corresponding reference center. The two windows are therefore not sampled around one shared geographical center. Because the simulated WRF vortex usually remains close to the center of its reference domain, the model can learn a mapping from the FuXi forecast structure around its own center to the WRF structure around the reference center without being dominated by track displacement. During inference, the reconstructed window continues to follow the center supplied by the input forecast, and 3DTCR does not independently correct the track.

Supplementary Figs.~\ref{fig:ws10m_4src_4day_MAWAR_20230524} and \ref{fig:ws10m_4src_4day_DOKSURI_20230722} display all data sources in a common WRF domain to show the positional differences that motivate center alignment during fine-tuning. This common domain is used only for illustration and does not represent the separately centered input and target windows used for training.


\FloatBarrier

\subsubsection{Forecast lead time conditioning}
\label{app:leadtime_details}

Forecast uncertainty evolves with lead time, so 3DTCR incorporates forecast lead time through a learnable embedding vector. The embedding module contains $\mathrm{N} = 21$ categories in total. Index 0 denotes the initialization time, corresponding to the analysis field with forecast lead time $\mathrm{K} = 0$, while indices 1 to 20 represent discrete forecast lead times.

This conditioning strategy allows the model to distinguish different forecast regimes within a unified framework. Inputs at short lead times are generally closer to the reanalysis distribution, whereas forecasts at longer lead times may contain larger accumulated errors and spatial displacement. Explicit encoding therefore provides temporal context and may help the model learn correction behavior that varies with lead time.

\subsubsection{MMD-based latent space alignment}
\label{app:mmd_application}

During supervised fine-tuning, ERA5 reanalysis serves as the reference domain, and FuXi forecasts form the forecast domain to be aligned with it. Because the two domains differ in their latent feature distributions, 3DTCR applies maximum mean discrepancy (MMD) at the U-Net bottleneck rather than directly to the output fields.

\textbf{MMD formulation.}
\label{app:mmd_formulation}
The MMD objective used for latent space alignment is formulated as follows.

\textbf{Latent space distributions.}
\label{Latent Space Domain Adaptation}
Let $\mathrm{D}_f = \{ \mathbf{x}_f^{(i)} \}$ denote samples from the FuXi forecast domain and $\mathrm{D}_r = \{ \mathbf{x}_r^{(i)} \}$ samples from the ERA5 reference domain. Both inputs are processed through a shared encoder $\mathrm{E}_\phi$ to obtain $\mathbf{z}_f = \mathrm{E}_\phi(\mathbf{x}_f)$ and $\mathbf{z}_r = \mathrm{E}_\phi(\mathbf{x}_r)$.

The primary objective is to minimize the distributional divergence between the forecast latent space $P(\mathbf{z}_f)$ and the reanalysis latent space $Q(\mathbf{z}_r)$. To quantify this divergence, we employ the MMD, an integral probability metric that uses kernels to measure the distance between two distributions within a reproducing kernel Hilbert space (RKHS) $\mathrm{H}_k$. Given a characteristic kernel $k(\cdot, \cdot)$, the squared MMD is formally defined as:

\begin{equation}
\mathrm{MMD}^2(P, Q) = \left\| \mathbb{E}_{\mathbf{z}_f \sim P} [\phi(\mathbf{z}_f)] - \mathbb{E}_{\mathbf{z}_r \sim Q} [\phi(\mathbf{z}_r)] \right\|_{\mathrm{H}_k}^2.
\end{equation}

In practice, by applying the kernel trick, we compute the biased empirical estimator of the squared MMD, which serves as our regularization loss during fine-tuning:

\begin{equation}
\mathrm{L}_{\mathrm{MMD}} = \frac{1}{N^2} \sum_{i,j} k(\mathbf{z}_f^{(i)}, \mathbf{z}_f^{(j)}) + \frac{1}{N^2} \sum_{i,j} k(\mathbf{z}_r^{(i)}, \mathbf{z}_r^{(j)}) - \frac{2}{N^2} \sum_{i,j} k(\mathbf{z}_f^{(i)}, \mathbf{z}_r^{(j)}).
\end{equation}

Here $N$ represents the batch size. To capture feature discrepancies across various abstraction levels, we utilize a multiscale Gaussian kernel, defined as $k(\mathbf{u}, \mathbf{v}) = \sum_l \exp(-\|\mathbf{u}-\mathbf{v}\|_2^2 / 2\sigma_l^2)$.

\textbf{Role in optimization.} Minimizing $\mathrm{L}_{\mathrm{MMD}}$ aligns the FuXi latent distribution with the ERA5 reference distribution. This supports transfer of the mapping learned during pretraining, while ERA5 is not required at inference.

The following subsection combines this alignment term with the Rectified Flow reconstruction objective.

\subsubsection{Two-stage training and objective functions}
\label{app:training_objectives}

Training proceeds in two stages. Pretraining pairs ERA5 inputs with high-resolution WRF targets to learn the coarse-to-fine mapping using the Rectified Flow objective:
\begin{equation}
\mathrm{L}_{\text{RF}} = \frac{1}{N} \sum_{i=1}^{N} \left\| \hat{v}_i - v_i \right\|_2^2,
\label{eq:mse_loss_app}
\end{equation}
where \(\hat{v}_i\) and \(v_i\) denote the predicted and target velocities, respectively, and \(N\) is the number of elements included in the loss.

Supervised fine-tuning then initializes from the pretrained weights and jointly uses ERA5 and FuXi inputs. Its objective is
\begin{equation}
\mathrm{L}_{\text{total}} = \mathrm{L}_{\text{RF}} + \lambda \mathrm{L}_{\text{MMD}},
\label{eq:total_loss_app}
\end{equation}
where \(\mathrm{L}_{\text{RF}}\) retains the reconstruction objective and \(\mathrm{L}_{\text{MMD}}\) aligns the FuXi bottleneck feature distribution with the ERA5 reference distribution.

\subsubsection{Implementation details}
\label{app:optimization_details}

The model was implemented in PyTorch \cite{paszke2017automatic} and trained on a single NVIDIA A100 GPU. During pretraining, the Rectified Flow model was optimized for 175,000 iterations with a batch size of 8. AdamW \cite{kingma2014adam,loshchilov2017decoupled} was used with $\mathrm{\beta}_1 = 0.9$, $\mathrm{\beta}_2 = 0.95$, an initial learning rate of $\mathrm{1\times10^{-5}}$, and a weight decay coefficient of $\mathrm{1\times10^{-5}}$. A cosine annealing schedule \cite{loshchilov2016sgdr} with 500 warmup steps was adopted to improve training stability.

During supervised fine-tuning, the learning rate was set to \ensuremath{1 \times 10^{-6}}, and the MMD weight was fixed at \ensuremath{\mathrm{\lambda} = 0.01}. For Rectified Flow training, the interpolation path was discretized into 1000 time levels, with one level uniformly sampled for each training example. During inference, the learned flow ODE was solved using 10 Euler steps for efficient reconstruction.

\subsection{Experimental design and ablation configurations}
  \label{app:ablation_framework}

  The main text provides a concise overview of the model variants and
  their configurations. Here, we provide a more detailed description of
  the ablation design, including the initialization strategy, training
  data, optimization objectives, inference inputs, and experimental role
  of each variant.

  The ablation analysis is organized along two complementary dimensions.
  The first examines the contribution of the two-stage training strategy
  by comparing the full two-stage model, 3DTCR-SFT (FuXi, w/ MMD), with
  the one-stage 3DTCR-E2E (FuXi, w/ MMD) variant. The E2E variant uses
  the same U-Net backbone and combined reconstruction and MMD objectives
  but is trained from random initialization without the ERA5-WRF
  pretraining stage. This comparison therefore assesses whether learning
  a physically consistent coarse-to-fine mapping before adaptation to
  FuXi forecasts reduces the optimization difficulty.

  The second dimension examines the contribution of MMD-based latent
  space alignment. For this purpose, the full model is compared
  with 3DTCR-SFT (FuXi, w/o MMD), which is initialized from the
  pretrained weights and subsequently fine-tuned on FuXi inputs using
  only the $\mathrm{L}_{\text{RF}}$ reconstruction objective. Unlike
  the full model, this variant does not explicitly align the latent
  representations of FuXi forecasts and ERA5 reanalysis. The comparison
  therefore assesses the additional benefit of latent space alignment
  beyond supervised fine-tuning.

  All 3DTCR variants use the same U-Net backbone and WRF 3-km
  reconstruction target. Their differences arise from the initialization
  strategy, training data, and optimization objectives rather than from
  changes in model architecture or the high-resolution reference. FuXi,
  ERA5, and ECMWF-HRES are included as baseline datasets or forecasting
  systems, whereas the WRF simulations provide the common
  high-resolution reconstruction reference.

  For all 3DTCR variants, the data source in parentheses denotes the
  input used during inference rather than necessarily the data used
  during training. Both 3DTCR-Pre variants are trained exclusively on
  spatially aligned ERA5-WRF pairs. The 3DTCR-Pre (ERA5) variant
  evaluates the pretrained reconstruction mapping within the ERA5 input
  domain, whereas 3DTCR-Pre (FuXi) evaluates its direct zero-shot
  application to FuXi forecasts without forecast-domain adaptation.

  All SFT variants initialize their parameters from the ERA5-WRF
  pretrained model. The 3DTCR-SFT (FuXi, w/o MMD) variant is
  fine-tuned on FuXi inputs using only the reconstruction loss. In
  contrast, the full 3DTCR-SFT (FuXi, w/ MMD) model is fine-tuned using
  both ERA5 and FuXi inputs and optimized with

  \begin{equation}
  \mathrm{L}_{\mathrm{total}}
  =
  \mathrm{L}_{\mathrm{RF}}
  +
  \lambda \mathrm{L}_{\mathrm{MMD}}.
  \end{equation}

  The MMD term is applied to the bottleneck features of the U-Net to
  encourage alignment between the ERA5 and FuXi latent distributions.
  Finally, 3DTCR-E2E (FuXi, w/ MMD) uses the same combined objective
  but is trained from random initialization, providing the one-stage
  reference for evaluating the contribution of the two-stage strategy.
  In the variant names, \enquote{w/ MMD} and \enquote{w/o MMD} indicate
  whether MMD-based latent alignment is included during optimization.

\subsection{Evaluation metrics}
\label{Evaluation metrics details}

The main text evaluates the time series of scalar TC intensity at each forecast lead time using maximum WS10M and reports the corresponding TC track errors. Track error at each lead time is calculated as the great-circle distance between the predicted TC center coordinates and the corresponding IBTrACS best-track coordinates. To complement these scalar intensity and track metrics, we further assess the reconstruction of fine-scale extreme wind features using the critical success index (CSI), which is a standard metric for verifying rare events in meteorology. Let $x$ and $\hat{x}$ denote the reference and reconstructed 10-m wind speed (WS10M), respectively. For a given intensity threshold $\theta$, CSI formulates the reconstruction task as a binary event detection problem and is defined as
\begin{equation}
\mathrm{CSI} = \frac{\mathrm{TP}}{\mathrm{TP} + \mathrm{FP} + \mathrm{FN}}
\label{eq:csi_app},
\end{equation}
where $\mathrm{TP}$ denotes true positives, corresponding to grid points where both $x \geq \theta$ and $\hat{x} \geq \theta$; $\mathrm{FP}$ denotes false positives, corresponding to grid points where $x < \theta$ but $\hat{x} \geq \theta$; and $\mathrm{FN}$ denotes false negatives, corresponding to grid points where $x \geq \theta$ but $\hat{x} < \theta$.

$\mathrm{CSI}$ ranges from 0 to 1, with higher values indicating better agreement in the spatial identification of wind events exceeding the threshold. Unlike bulk error metrics such as RMSE, $\mathrm{CSI}$ is insensitive to the number of true negatives and is therefore particularly suitable for evaluating rare and spatially localized extremes, where grid points without events dominate the field.

To systematically evaluate model performance across different wind intensities, we compute $\mathrm{CSI}$ at six WS10M thresholds: 13.9, 17.2, 20.8, 24.5, 28.5, and 32.7~m/s. These thresholds are based on standard meteorological intensity classifications and span representative wind speed regimes from tropical depression to typhoon conditions. By computing $\mathrm{CSI}$ independently at each threshold, we obtain a stratified assessment of the model's ability to reconstruct extreme inner-core wind features across progressively stronger TC intensity levels.

\section{Supplementary results}

The supplementary results follow the sequence of the Results section in the main text. We first provide additional evaluation of TC intensity and track forecasts. We then extend the structural analysis through near-surface fields, vertical wind structure, multilevel meteorological fields, comparisons with satellite cloud patterns, temporal evolution, transfer to other basins, and spectral and harmonic diagnostics. The final two subsections complement the ablation of two-stage training and latent space alignment and the comparison of computational cost presented in the main text. Typhoon KONG-REY remains the primary structural case, Typhoon YAGI provides a second case evaluated with the same diagnostics, and Typhoon YINXING provides an additional near-surface example. Throughout these comparisons, 3DTCR denotes the full 3DTCR-SFT (FuXi, w/ MMD) model, consistent with the terminology in the main text.

\subsection{TC intensity and track evaluation}

This subsection supplements the forecast evaluation in the main text by reporting maximum WS10M performance and TC track errors separately. The intensity results evaluate the reconstructed vortex, whereas the track results reflect the TC positions supplied by the corresponding input forecasts.

\subsubsection{Maximum 10-m wind speed evaluation}

\begin{figure*}[!htbp]
    \centering
    \includegraphics[width=1.0\linewidth]{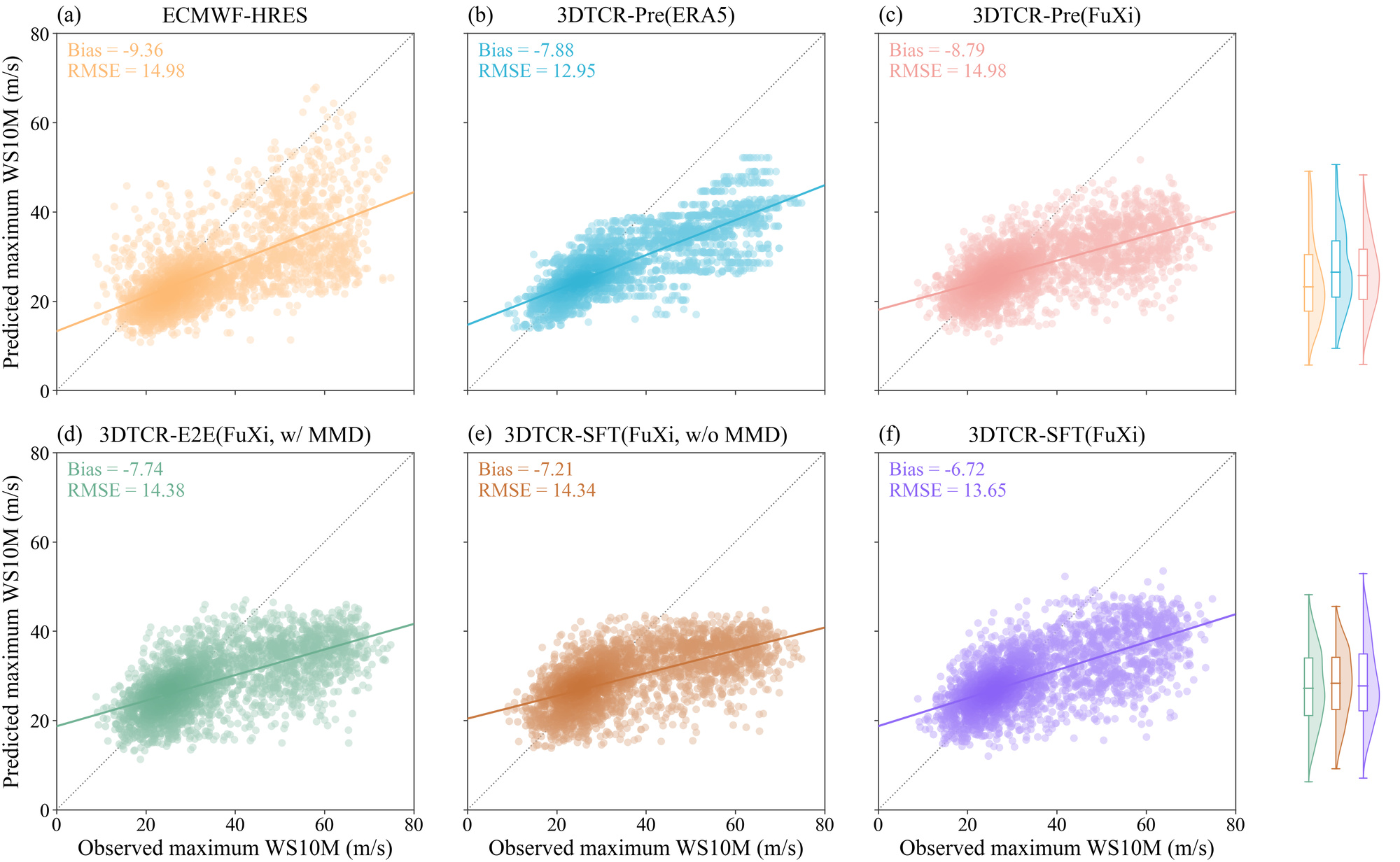}
    \caption{\textbf{Comparison of predicted and observed maximum 10-m wind speed (WS10M).}
    Panels (\textbf{a}--\textbf{f}) compare predictions with IBTrACS observations for 2242 samples pooled across all forecast lead times in the 2024 WNP test dataset: (\textbf{a}) ECMWF-HRES, (\textbf{b}) 3DTCR-Pre (ERA5), (\textbf{c}) 3DTCR-Pre (FuXi), (\textbf{d}) 3DTCR-E2E (FuXi, w/ MMD), (\textbf{e}) 3DTCR-SFT (FuXi, w/o MMD), and (\textbf{f}) 3DTCR-SFT (FuXi, w/ MMD; full model). Solid lines show linear fits with 95\% confidence intervals, and dashed lines indicate the 1:1 reference. The rightmost column shows marginal distributions using box plots and kernel density estimates (KDEs); inset values report the overall mean bias and RMSE. Because best-track intensities are typically reported in 5 kt increments, and forecasts from different initialization times can correspond to the same verifying observation, repeated observed values may appear in the scatter plots. To reduce overplotting, a random jitter of $\pm 2$~m/s was applied only along the horizontal axis to the plotted observational values for visualization purposes. All verification metrics were computed using the original, unjittered observational values.}
    \label{figures/scatter_distribution_adjusted.png}
\end{figure*}

The main text reports the lead-time evolution of maximum WS10M errors for 1--5-day forecasts. Supplementary Table~\ref{tab:forecast_comparison} provides the corresponding RMSE values at 24-h intervals for the baselines, 3DTCR variants, and WRF 3-km reference. Supplementary Fig.~\ref{figures/scatter_distribution_adjusted.png} complements these lead-time statistics by comparing predicted and observed maximum WS10M across all test samples and model variants.

\FloatBarrier
\begin{table}[!htp]
\centering
\small
\caption{Root mean square error (RMSE) of maximum 10-m wind speed (WS10M) predictions (m/s; lower is better $\downarrow$).}
\label{tab:forecast_comparison}
\begin{tabularx}{\textwidth}{@{}X*{5}{c}@{}}
\toprule
\textbf{Method} & \textbf{24~h} & \textbf{48~h} & \textbf{72~h} & \textbf{96~h} & \textbf{120~h} \\
\midrule
\multicolumn{6}{l}{\hspace{-4pt}\textbf{Baselines}} \\
\midrule
FuXi & 20.19 & 21.64 & 22.80 & 22.48 & 23.91 \\
ERA5 & 18.92 & 19.77 & 20.04 & 19.73 & 21.20 \\
ECMWF-HRES & 13.96 & 14.90 & 15.63 & 16.83 & 19.12 \\
\midrule
\multicolumn{6}{l}{\hspace{-4pt}\textbf{Ours}} \\
\midrule
\text{3DTCR-E2E (FuXi, w/ MMD)} & \text{13.55} & \text{15.06} & \text{16.67} & \text{16.56} & \text{17.49} \\
3DTCR-Pre (ERA5) & \textbf{12.42} & \textbf{12.82} & \textbf{13.01} & \textbf{12.30} & \textbf{13.71} \\
3DTCR-Pre (FuXi) & 13.43 & 14.99 & 16.64 & 16.44 & 17.33 \\
\text{3DTCR-SFT (FuXi, w/o MMD)} & \text{12.99} & \text{14.19} & \text{16.32} & \text{17.05} & \text{17.56} \\
\textbf{3DTCR-SFT (FuXi, w/ MMD)} & \textbf{12.35} & \textbf{13.99} & \textbf{15.84} & \textbf{15.60} & \textbf{16.31} \\

\midrule
\multicolumn{6}{l}{\hspace{-4pt}\textbf{Target}} \\
\midrule
WRF 3-km & 10.95 & 11.47 & 11.89 & 10.89 & 12.13 \\
\bottomrule
\end{tabularx}
\end{table}

\FloatBarrier
\subsubsection{TC track forecast errors}
\label{sec:track_error}

\begin{figure*}[!htbp]
    \centering
    \includegraphics[width=\linewidth]{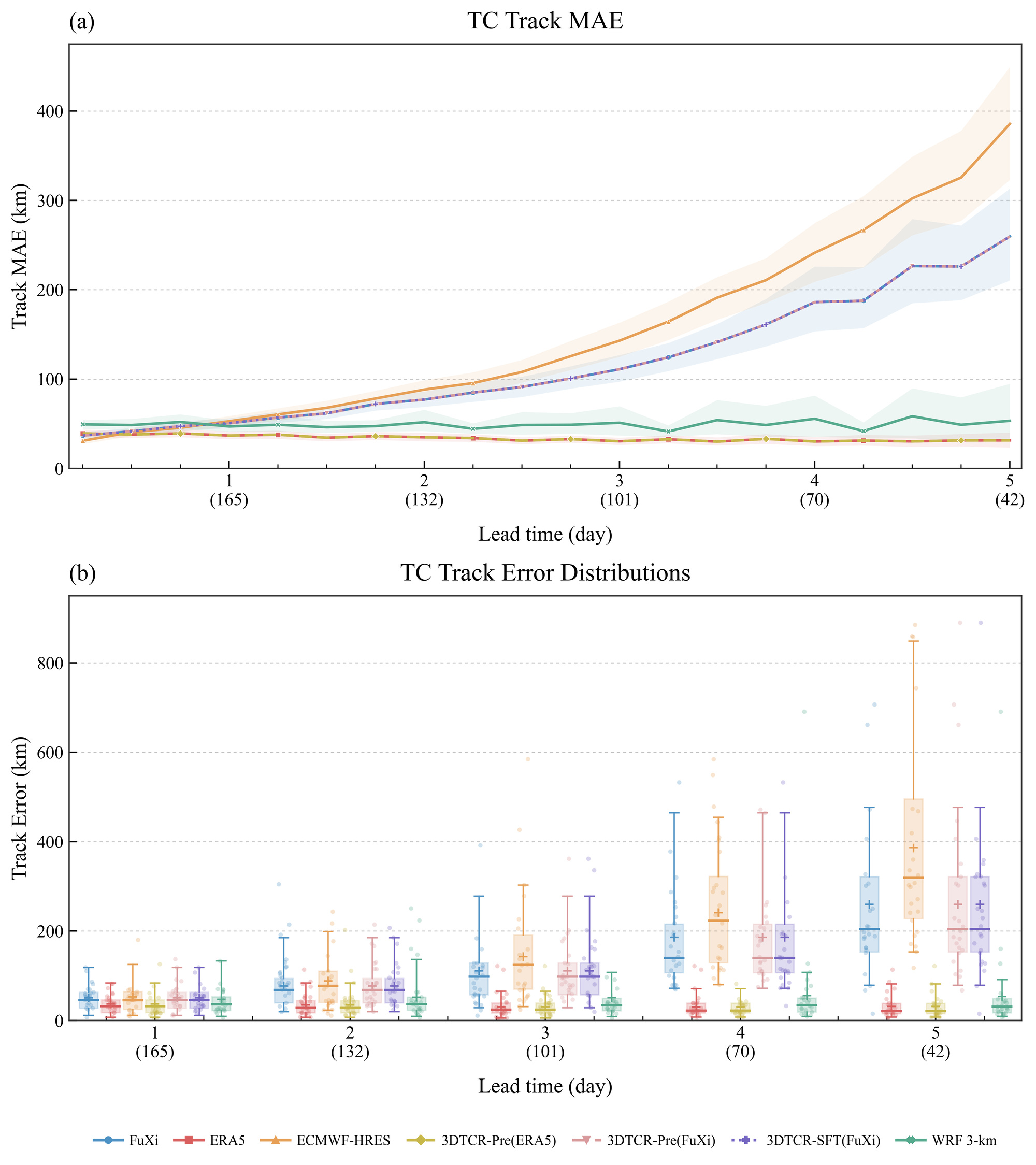}
    \caption{\textbf{TC track errors across forecast lead times.} Track error is calculated as the great-circle distance between the predicted and IBTrACS TC center coordinates. (a) Track MAE for 1--5-day forecasts, with shading denoting the 95\% bootstrap confidence intervals estimated from 1000 resamples. (b) Distributions of TC track errors for 1--5-day forecasts. Boxes indicate the interquartile range, center lines the median, plus signs the mean, and whiskers the 5th--95th percentiles. Points represent randomly selected samples, and numbers in parentheses indicate the sample sizes. Because 3DTCR uses the TC centers provided by FuXi or ERA5, its track error curves and distributions overlap with those of the corresponding inputs.}
    \label{fig:track_error}
\end{figure*}

Recent AI-based global weather models~\cite{shi2025comparison,pathak2022fourcastnet,bi2023accurate,chen2023fuxi,lam2023learning} have substantially improved TC track prediction. At each lead time, track error is calculated as the great-circle distance between the TC center supplied by the input forecast and the corresponding IBTrACS best-track coordinates. Because 3DTCR reconstructs a window centered on the TC position supplied by the input field, it does not independently predict or correct the track. Supplementary Fig.~\ref{fig:track_error} therefore reports the track errors associated with the corresponding FuXi or ERA5 inputs. The text in parentheses identifies the input source; for example, 3DTCR-Pre (FuXi) uses FuXi TC positions, whereas 3DTCR-Pre (ERA5) uses ERA5 positions. The overlapping curves confirm that 3DTCR preserves the position supplied by the input while reconstructing the inner-core structure.

\FloatBarrier
\subsection{Reconstruction of TC structure}

This section provides additional evidence for TC structure reconstruction to complement the corresponding results in the main text. We first present the spatial distributions of fields near the surface for additional TCs and forecast lead times. We then provide additional results for KONG-REY and YAGI that complement the structural analysis in the main text, covering vertical wind structure, multilevel meteorological fields, comparisons between near-surface winds and satellite cloud imagery, and temporal evolution. The remaining results assess transfer to other basins and further examine vortex structure through spectral and azimuthal harmonic analysis.

\subsubsection{Structure of near-surface fields}

\begin{figure}[!htbp]
    \centering
    \includegraphics[width=1.0\linewidth]{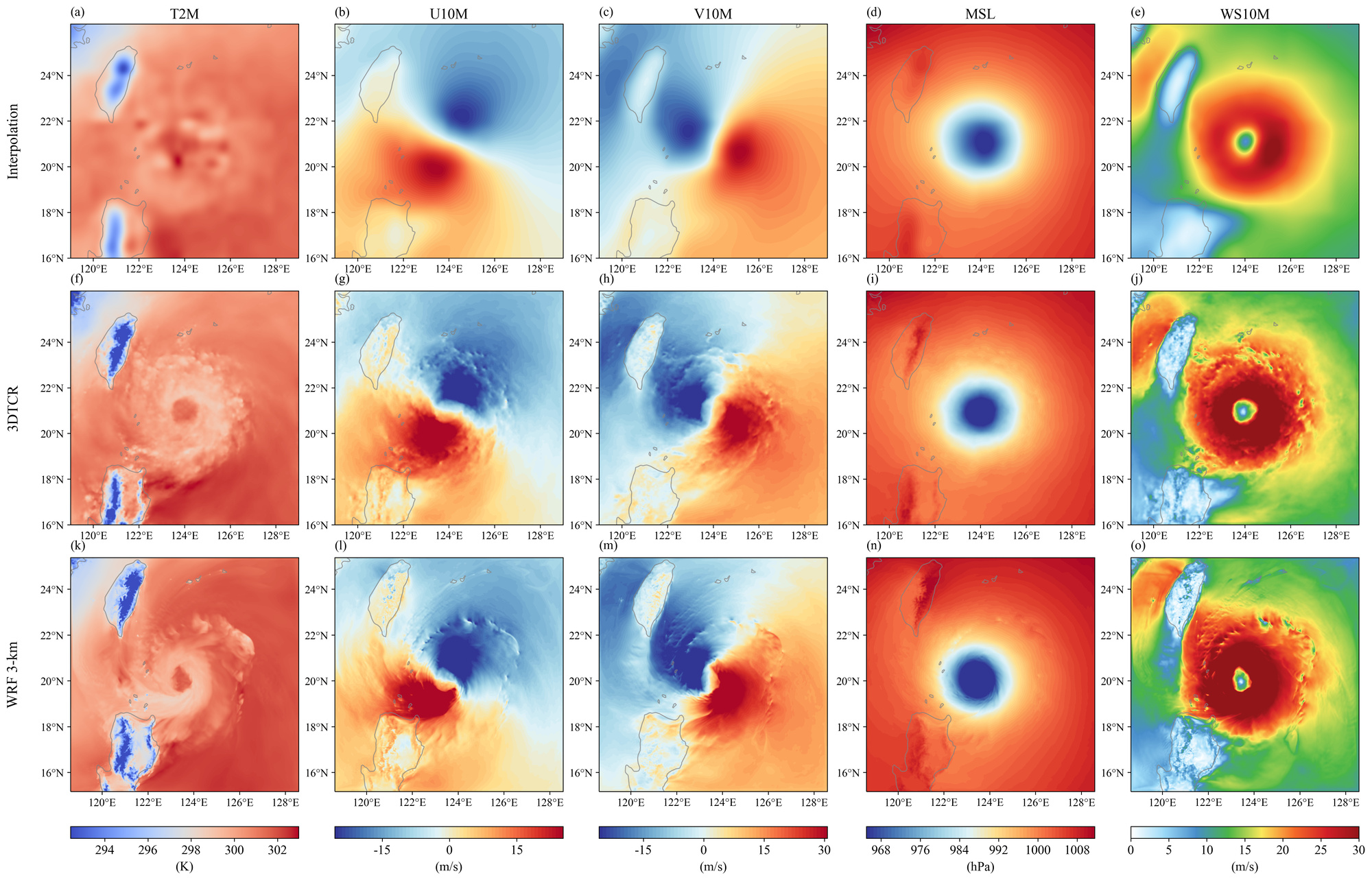}
    \caption{\textbf{Typhoon KONG-REY multivariable field reconstruction at 96 h forecast lead time.} The forecast was initialized at 1200 UTC 26 October 2024. Columns show 2-m temperature (T2M), 10-m u and v wind components (U10M and V10M), mean sea level pressure (MSL), and 10-m wind speed (WS10M). Rows show bilinearly interpolated FuXi fields, the 3DTCR reconstruction, and the WRF 3-km reference.}
    \label{figures/KONG-REY_Init2024102612_LeadStep16.jpg}
\end{figure}

\begin{figure*}[!htbp]
    \centering
    \includegraphics[width=1.0\linewidth]{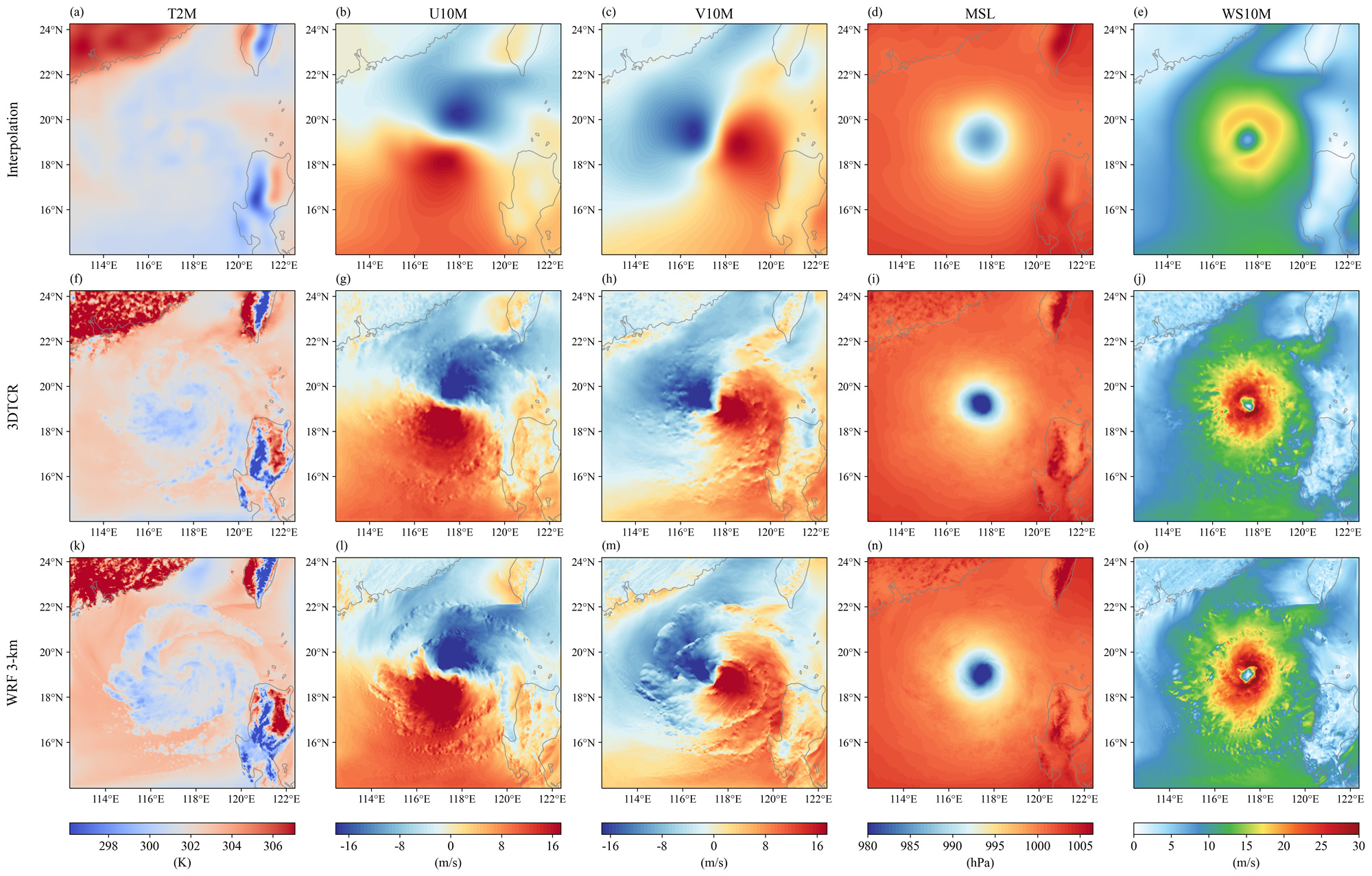}
    \caption{\textbf{Typhoon YAGI multivariable field reconstruction at 18 h forecast lead time.} The forecast was initialized at 1200 UTC 3 September 2024. Columns show 2-m temperature (T2M), 10-m u and v wind components (U10M and V10M), mean sea level pressure (MSL), and 10-m wind speed (WS10M). Rows show bilinearly interpolated FuXi fields, the 3DTCR reconstruction, and the WRF 3-km reference.}
    \label{figures/YAGI_Init2024090312_LeadStep3.jpg}
\end{figure*}

\begin{figure*}[!htbp]
    \centering
    \includegraphics[width=1.0\linewidth]{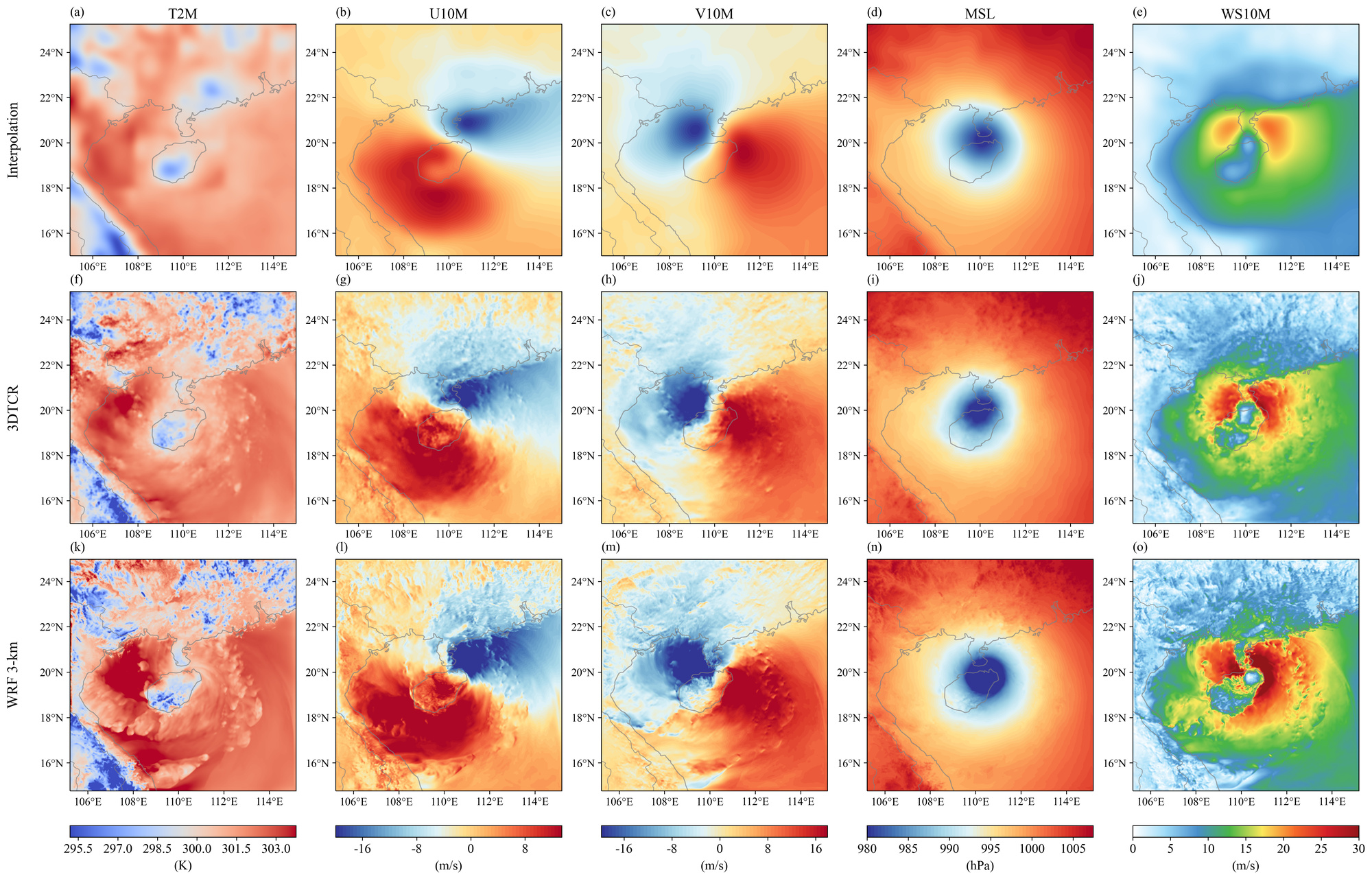}
    \caption{\textbf{Typhoon YAGI multivariable field reconstruction at 108 h forecast lead time.} The forecast was initialized at 0000 UTC 2 September 2024. Columns show 2-m temperature (T2M), 10-m u and v wind components (U10M and V10M), mean sea level pressure (MSL), and 10-m wind speed (WS10M). Rows show bilinearly interpolated FuXi fields, the 3DTCR reconstruction, and the WRF 3-km reference.}
    \label{figures/YAGI_Init2024090200_LeadStep18.jpg}
\end{figure*}

\begin{figure*}[!htbp]
    \centering
    \includegraphics[width=1.0\linewidth]{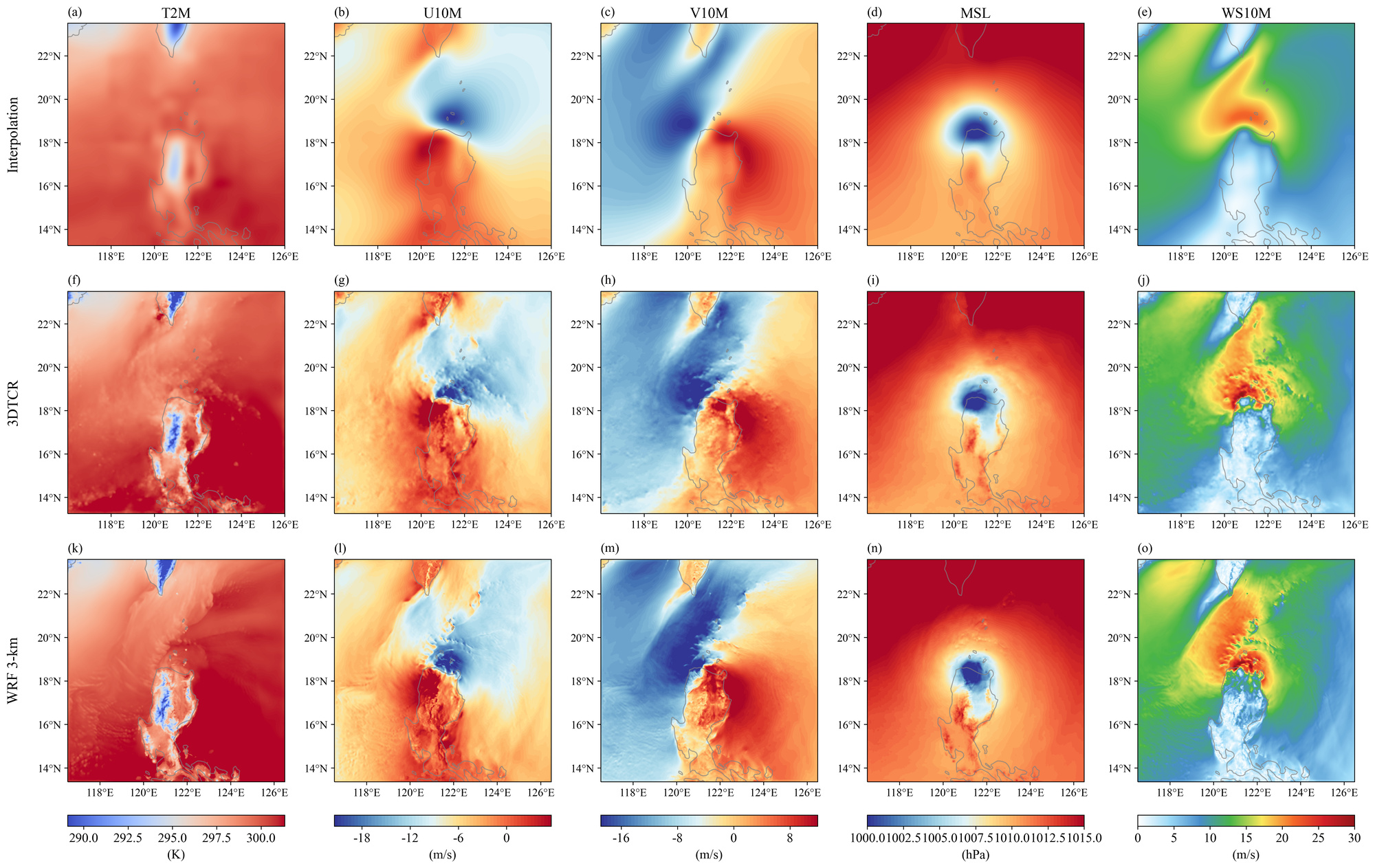}
    \caption{\textbf{Typhoon YINXING multivariable field reconstruction at 72 h forecast lead time.} The forecast was initialized at 1200 UTC 4 November 2024. Columns show 2-m temperature (T2M), 10-m u and v wind components (U10M and V10M), mean sea level pressure (MSL), and 10-m wind speed (WS10M). Rows show bilinearly interpolated FuXi fields, the 3DTCR reconstruction, and the WRF 3-km reference.}
    \label{figures/YINXING_Init2024110412_LeadStep12.jpg}
\end{figure*}

Supplementary Figs.~\ref{figures/KONG-REY_Init2024102612_LeadStep16.jpg}, \ref{figures/YAGI_Init2024090312_LeadStep3.jpg}, \ref{figures/YAGI_Init2024090200_LeadStep18.jpg}, and \ref{figures/YINXING_Init2024110412_LeadStep12.jpg} extend the near-surface structure comparison in the main text to KONG-REY at 96 h, YAGI at 18 and 108 h, and YINXING at 72 h. Consistent with the KONG-REY example in the main text, the reconstructed fields retain the large-scale environment while producing sharper wind speed gradients from the eye to the eyewall and greater azimuthal variability than the interpolated FuXi fields.

\FloatBarrier

\FloatBarrier
\subsubsection{Structure and evolution of Typhoon KONG-REY}

The main text presents KONG-REY as the primary structural example. Here, the same case is examined in greater detail using vertical wind diagnostics, multilevel meteorological fields, a qualitative comparison with satellite cloud patterns, and the temporal evolution of mean sea level pressure and near-surface winds.

\begin{figure*}[!htbp]
    \centering
    \includegraphics[width=\linewidth]{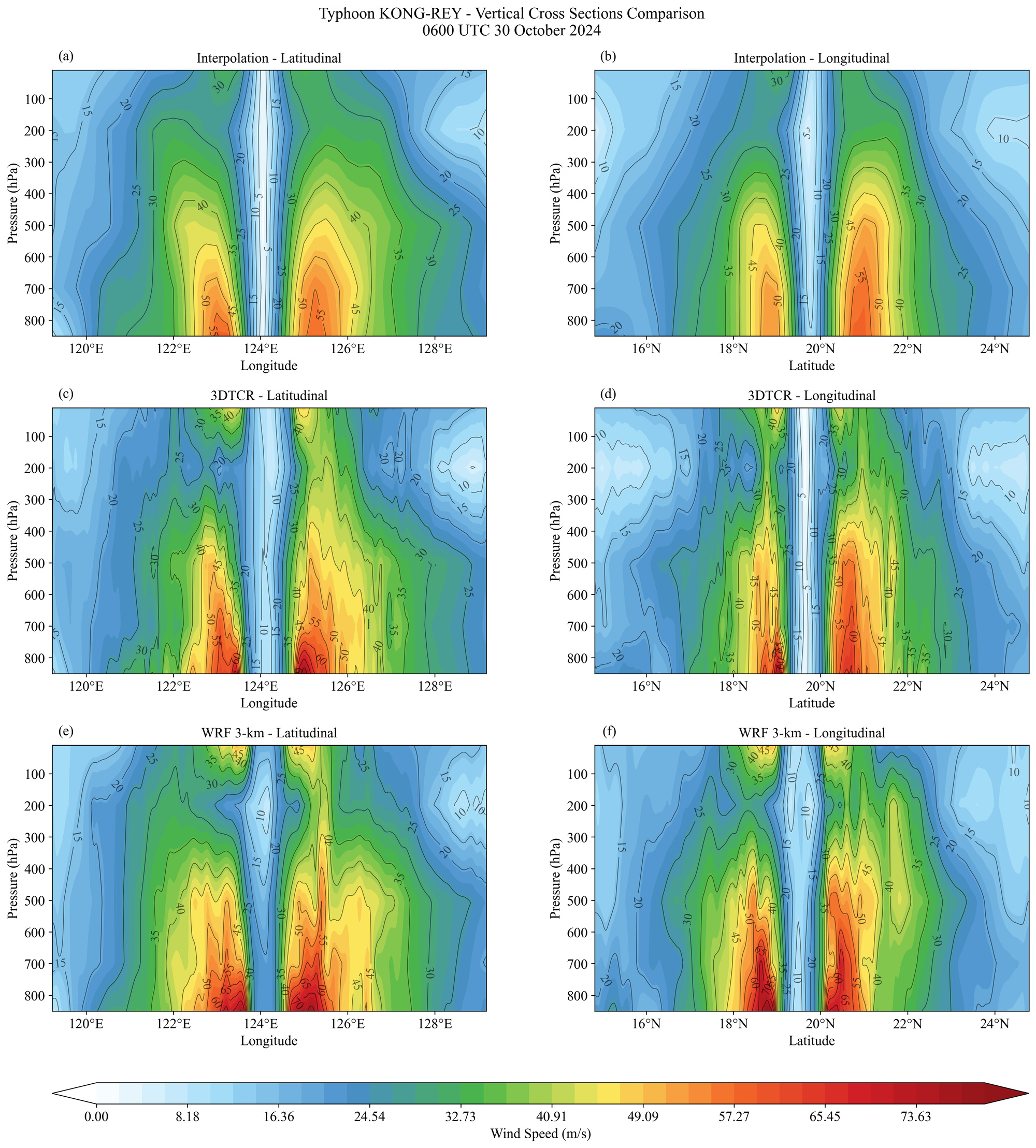}
    \caption{\textbf{Vertical wind structure of Typhoon KONG-REY at 0600 UTC on 30 October 2024.}
    (a, b) Interpolation, (c, d) 3DTCR, and (e, f) WRF 3-km results.
    The left and right columns show latitudinal and longitudinal cross sections, respectively.
    Shading and contours indicate wind speed (m/s), with pressure levels (hPa) on the vertical axis.
    The central white region and adjacent regions of high wind speed correspond to the eye and eyewall, respectively.}
    \label{fig:KONG-REY_vertical_cross_sections_all_2024-10-30T06-00-00_black_contour_k.png}
\end{figure*}

\begin{figure*}[!htbp]
    \centering
    \includegraphics[width=\linewidth]{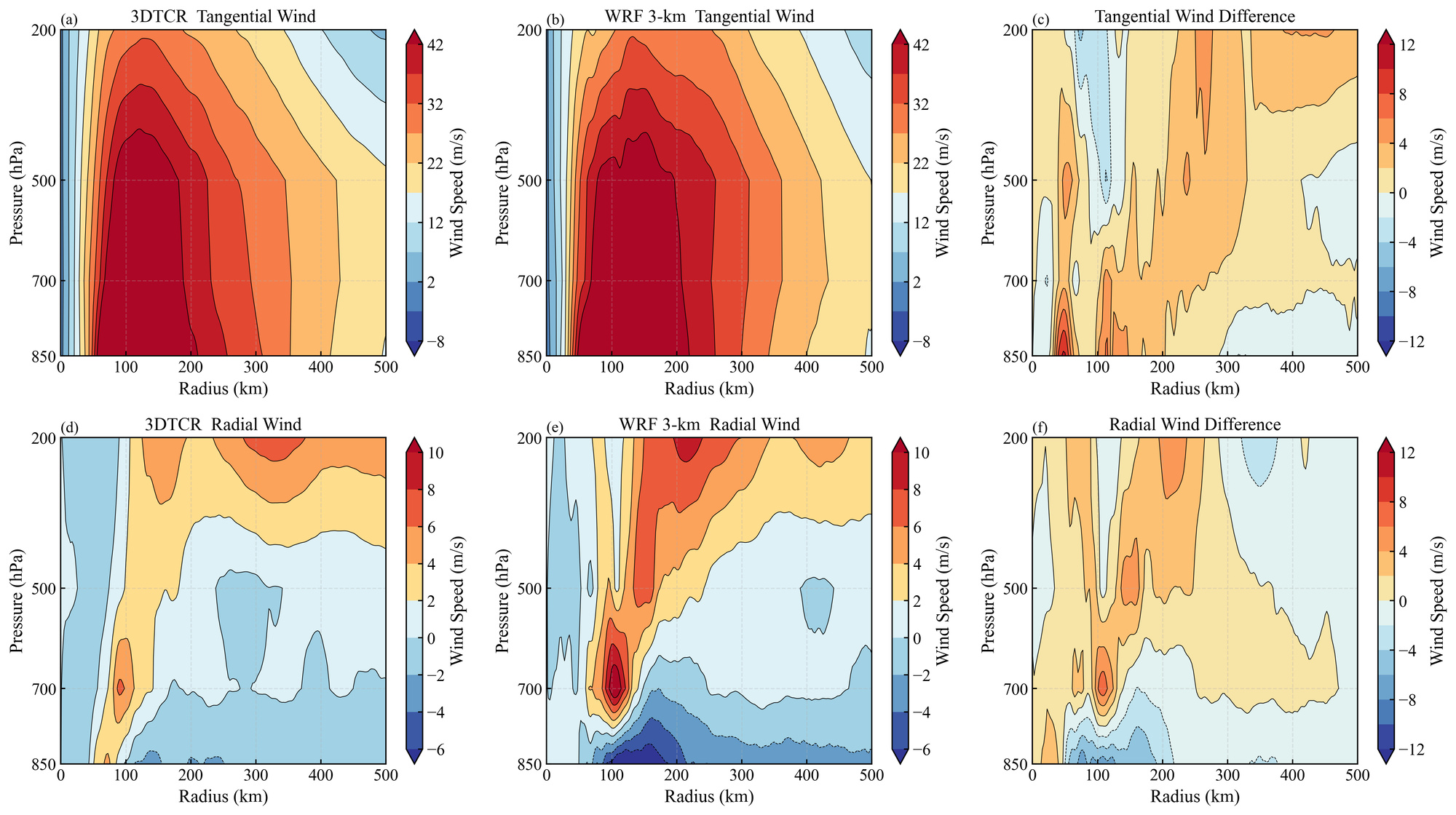}
    \caption{\textbf{Vertical cross sections of tangential and radial wind for Typhoon KONG-REY (0600 UTC on 30 October 2024).} (a) 3DTCR tangential wind; (b) WRF 3-km tangential wind; (c) Tangential wind difference (3DTCR minus WRF 3-km); (d) 3DTCR radial wind; (e) WRF 3-km radial wind; (f) Radial wind difference (3DTCR minus WRF 3-km). The horizontal axis represents the radius from the TC center (km), and the vertical axis represents pressure levels (hPa).}
    \label{fig:KONG-REY_fuxi_wrf_tangential_radial_winds_diff.png}
\end{figure*}

\paragraph*{Vertical wind structure.}
Supplementary Figs.~\ref{fig:KONG-REY_vertical_cross_sections_all_2024-10-30T06-00-00_black_contour_k.png} and \ref{fig:KONG-REY_fuxi_wrf_tangential_radial_winds_diff.png} extend the vertical analysis in the main text. The latitudinal and longitudinal wind speed cross sections show an eye with weak winds flanked by deep regions of high wind speed, while the azimuthal means show a pronounced maximum in tangential wind near the eyewall together with inflow at low levels and outflow at upper levels. These diagnostics provide additional evidence that 3DTCR reconstructs an organized three-dimensional structure rather than only increasing near-surface wind speed.

\begin{figure*}[!htbp]
    \centering
    \includegraphics[width=\linewidth]{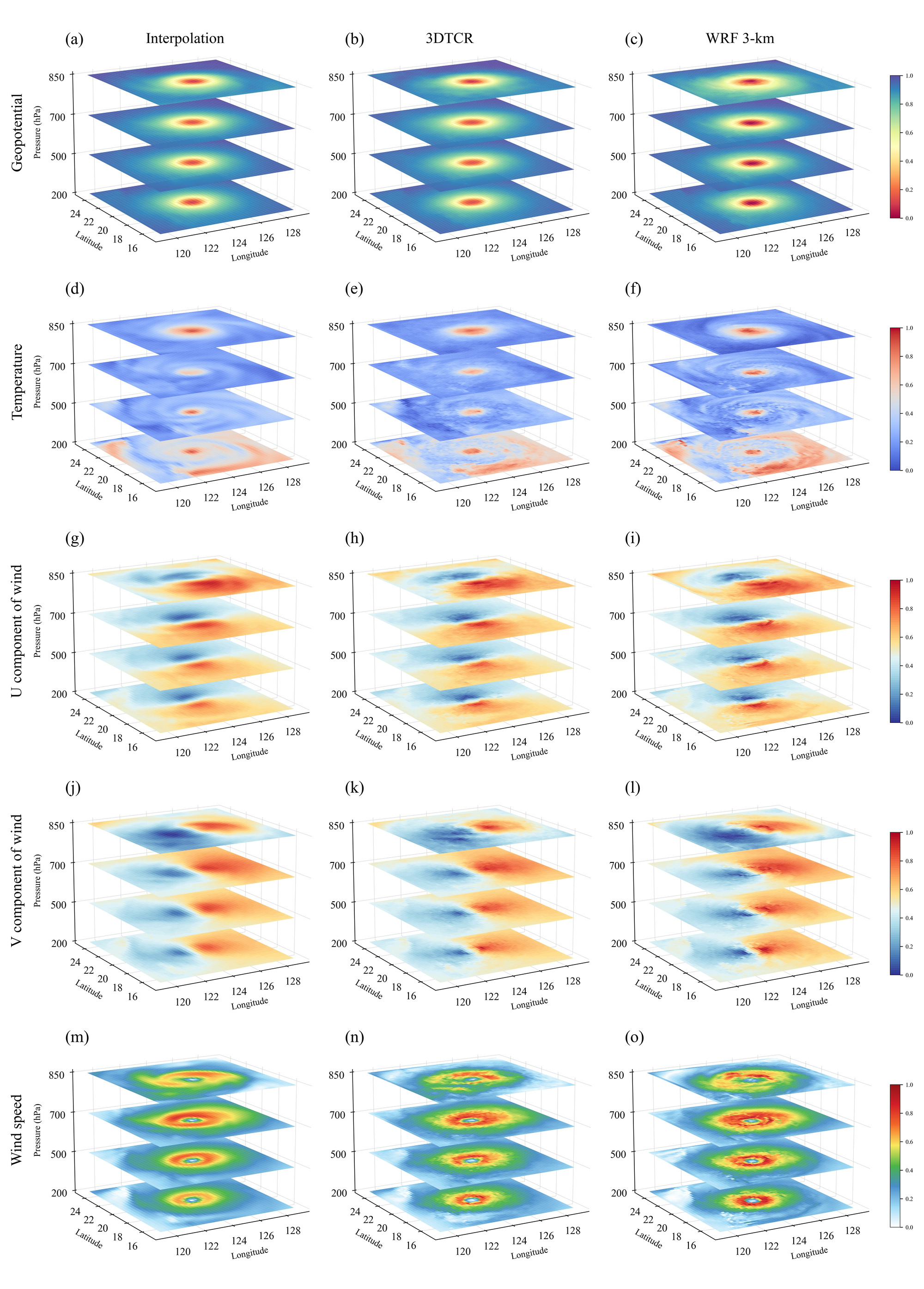}
    \caption{\textbf{Typhoon KONG-REY three-dimensional multivariable field reconstruction at 18 h lead time.} Comparison across vertical pressure levels ($200$, $500$, $700$, and $850\text{ hPa}$) for geopotential, temperature, u/v wind components, and 10-m surface wind speed, initialized at 1200 UTC 29 October 2024. Columns from left to right: Interpolation (FuXi), 3DTCR reconstruction, and WRF 3-km simulation (reference target).}
    \label{fig:3d_reconstruction}
\end{figure*}

\begin{figure*}[!htbp]
    \centering
    \includegraphics[width=\linewidth]{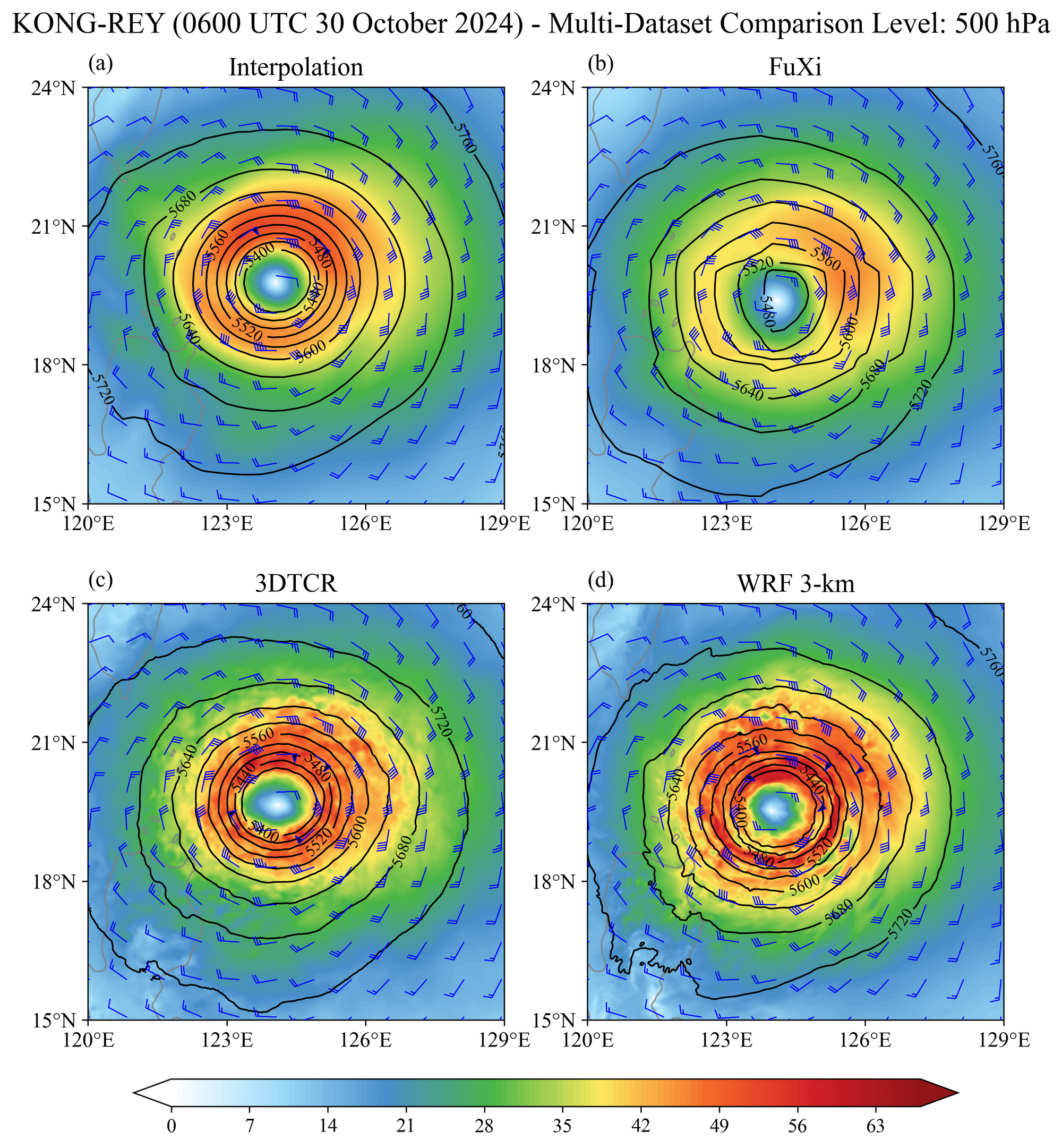}
    \caption{\textbf{Comparison of 500 hPa wind and geopotential height fields for Typhoon KONG-REY (0600 UTC on 30 October 2024).} (a) ERA5 interpolation data, (b) FuXi forecast interpolation data, (c) 3DTCR, (d) WRF 3-km. Color shading represents wind speed (m/s), black contours show geopotential height (gpm), and wind barbs indicate the 500 hPa wind field.}
    \label{fig:KONG-REY_circulation_500.png}
\end{figure*}

\begin{figure*}[!htbp]
    \centering
    \includegraphics[width=\linewidth]{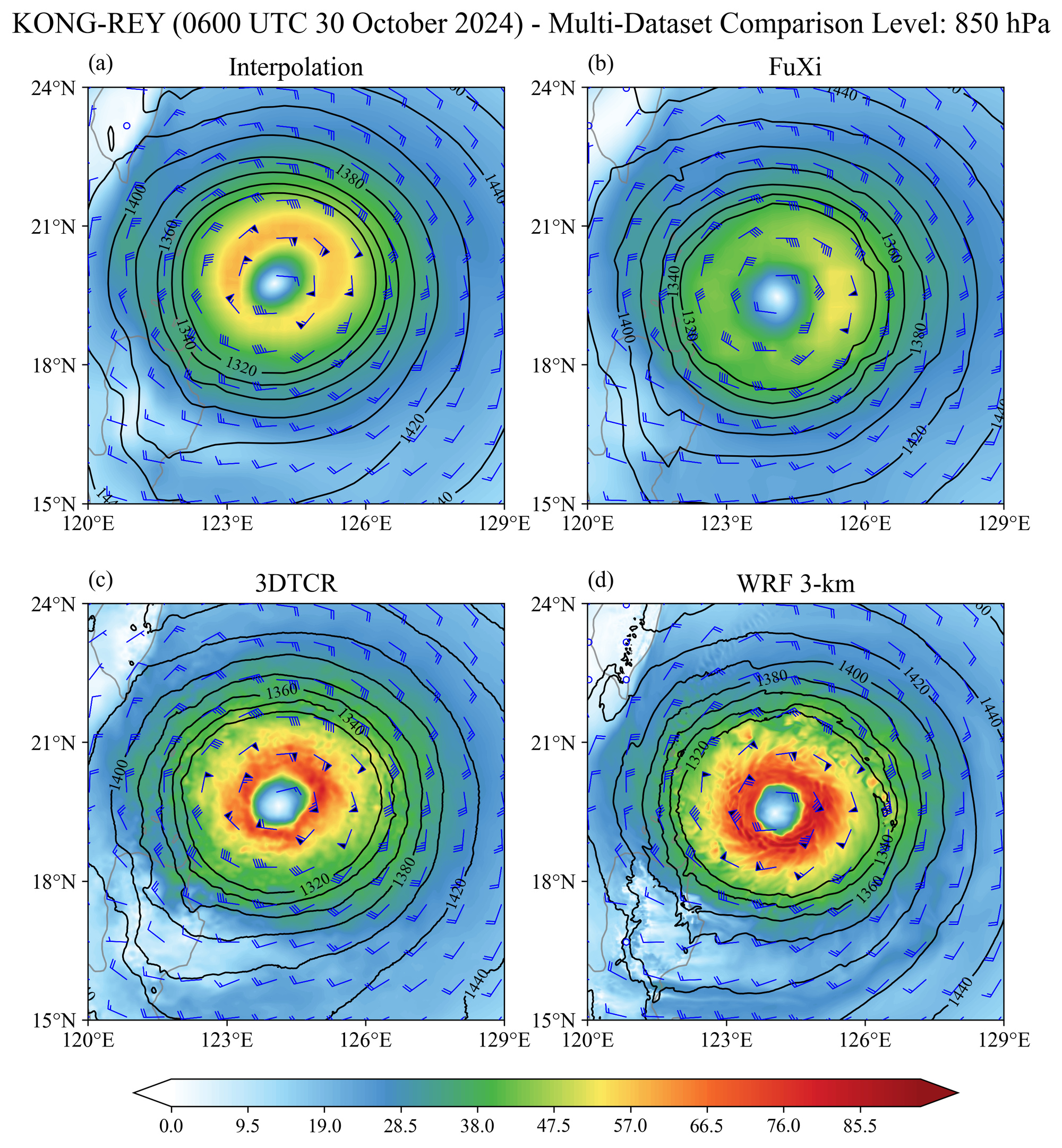}
    \caption{\textbf{Comparison of 850 hPa wind and geopotential height fields for Typhoon KONG-REY (0600 UTC on 30 October 2024).} (a) ERA5 interpolation data, (b) FuXi forecast interpolation data, (c) 3DTCR, (d) WRF 3-km. Color shading represents wind speed (m/s), black contours show geopotential height (gpm), and wind barbs indicate the 850 hPa wind field.}
    \label{fig:KONG-REY_circulation_850.png}
\end{figure*}

\begin{figure*}[!htbp]
    \centering
    \includegraphics[width=\linewidth]{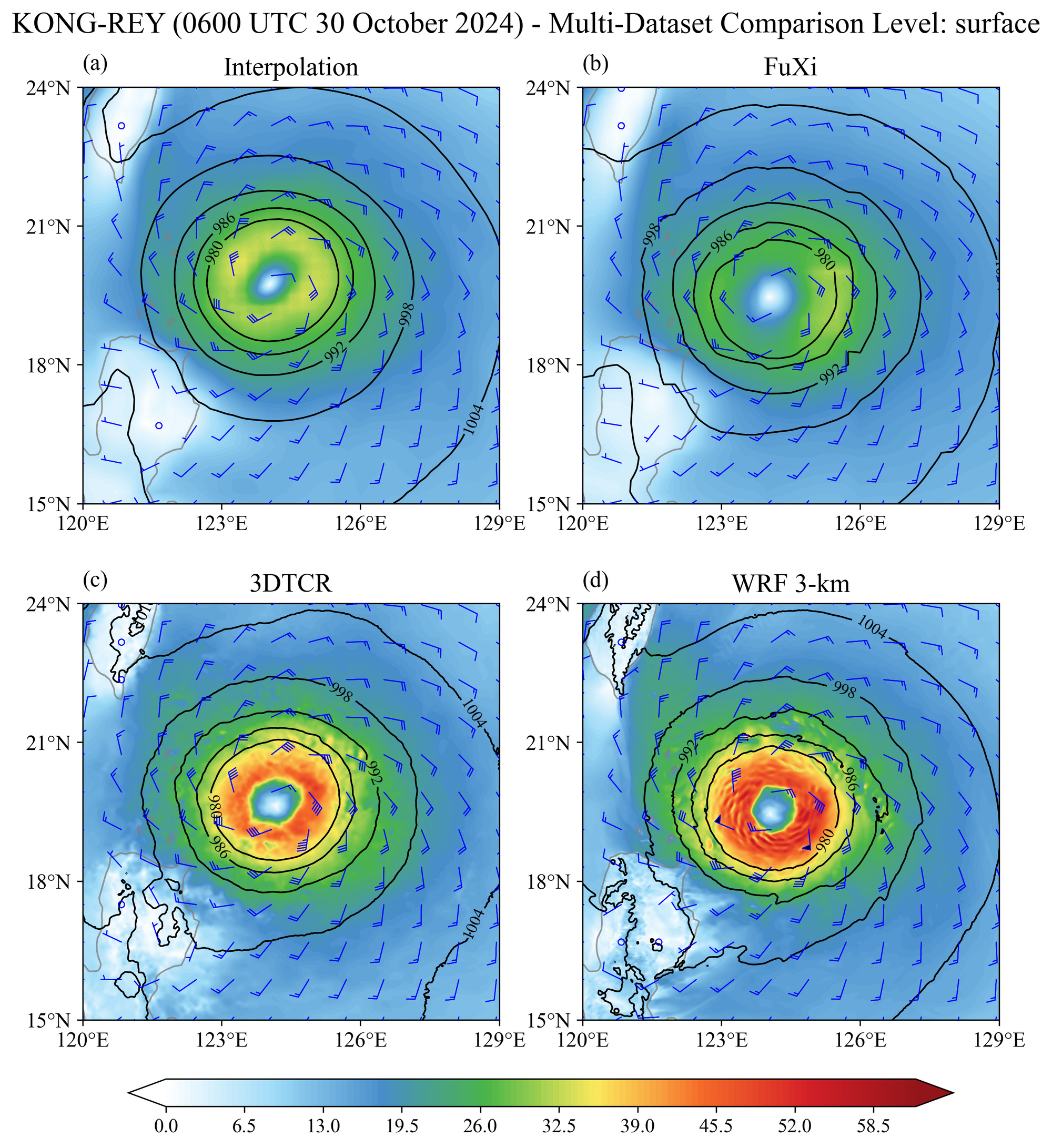}
    \caption{\textbf{Comparison of near-surface wind and mean sea level pressure fields for Typhoon KONG-REY (0600 UTC on 30 October 2024).} (a) ERA5 interpolation data, (b) FuXi forecast interpolation data, (c) 3DTCR, (d) WRF 3-km. Color shading represents wind speed (m/s), black contours show mean sea level pressure (hPa), and wind barbs indicate the 10-m wind field.}
    \label{fig:KONG-REY_circulation_surface.png}
\end{figure*}

\FloatBarrier
\paragraph*{Multilevel structure of meteorological fields.}
Supplementary Figs.~\ref{fig:3d_reconstruction}, \ref{fig:KONG-REY_circulation_500.png}, \ref{fig:KONG-REY_circulation_850.png}, and \ref{fig:KONG-REY_circulation_surface.png} complement the vertical diagnostics by comparing temperature, geopotential, wind, and mean sea level pressure fields across pressure levels and near the surface.


\begin{figure*}[!htbp]
\centering
\includegraphics[width=\linewidth]{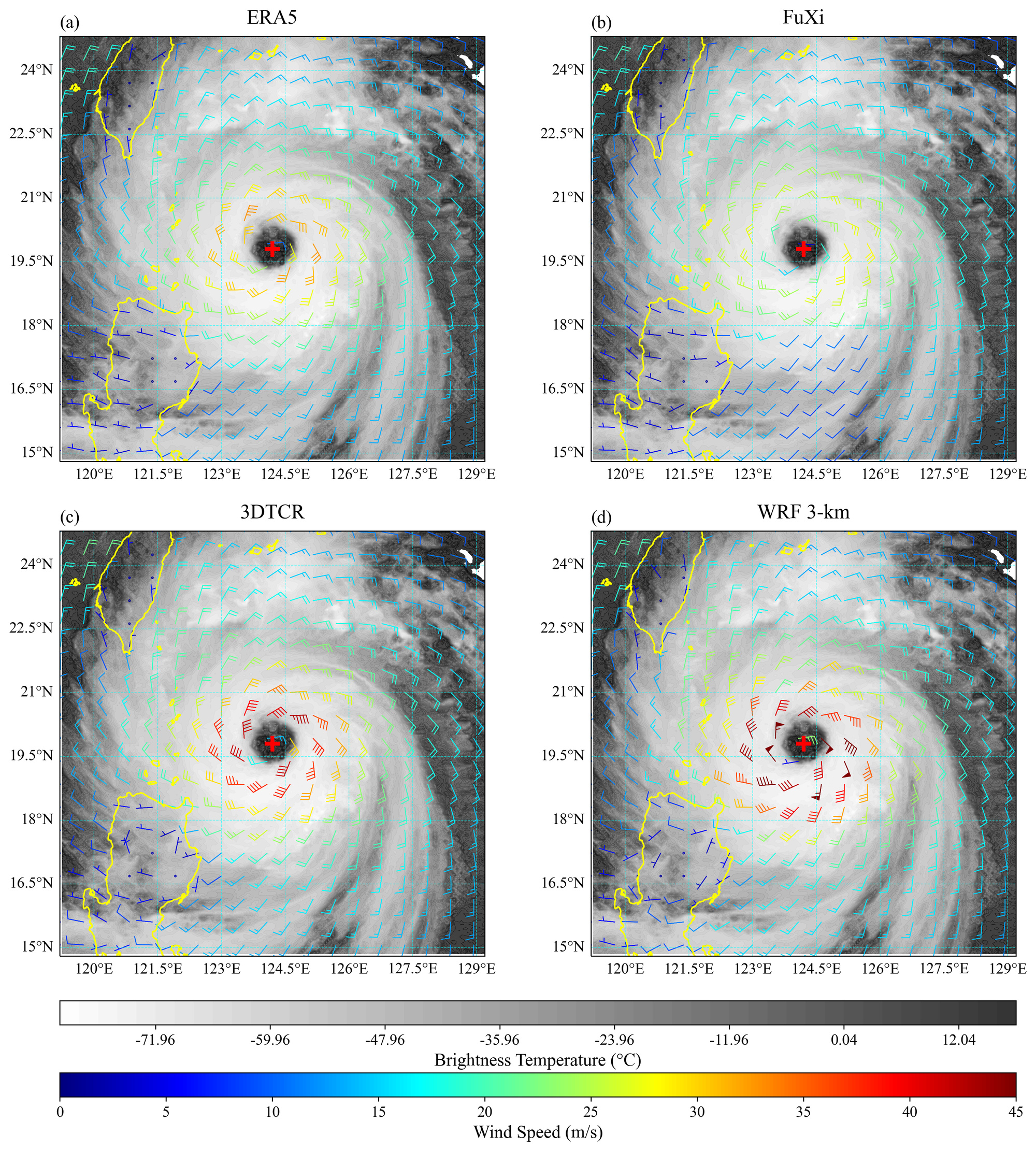}
\caption{\textbf{Comparison of cloud-top brightness temperature and 10-m wind fields for Typhoon KONG-REY (0600 UTC 30 October 2024).}
(a) ERA5; (b) FuXi; (c) 3DTCR; (d) WRF 3-km simulation.
Grayscale shading shows cloud-top brightness temperature (°C), colored wind barbs indicate 10-m wind speed (m/s) and direction, yellow lines denote coastlines, and red plus signs mark the IBTrACS best-track position.}
    \label{fig:KONG-REY_Cloud_Wind_barbs.png}
\end{figure*}

\FloatBarrier
\paragraph*{Comparison of surface winds with satellite cloud patterns.}
Supplementary Fig.~\ref{fig:KONG-REY_Cloud_Wind_barbs.png} compares 10-m wind fields with the observed cloud-top brightness temperature. The comparison is qualitative: the satellite image provides the cloud pattern rather than a direct observation of the 10-m wind field.

\begin{figure*}[!htbp]
    \centering
    \includegraphics[width=\linewidth]{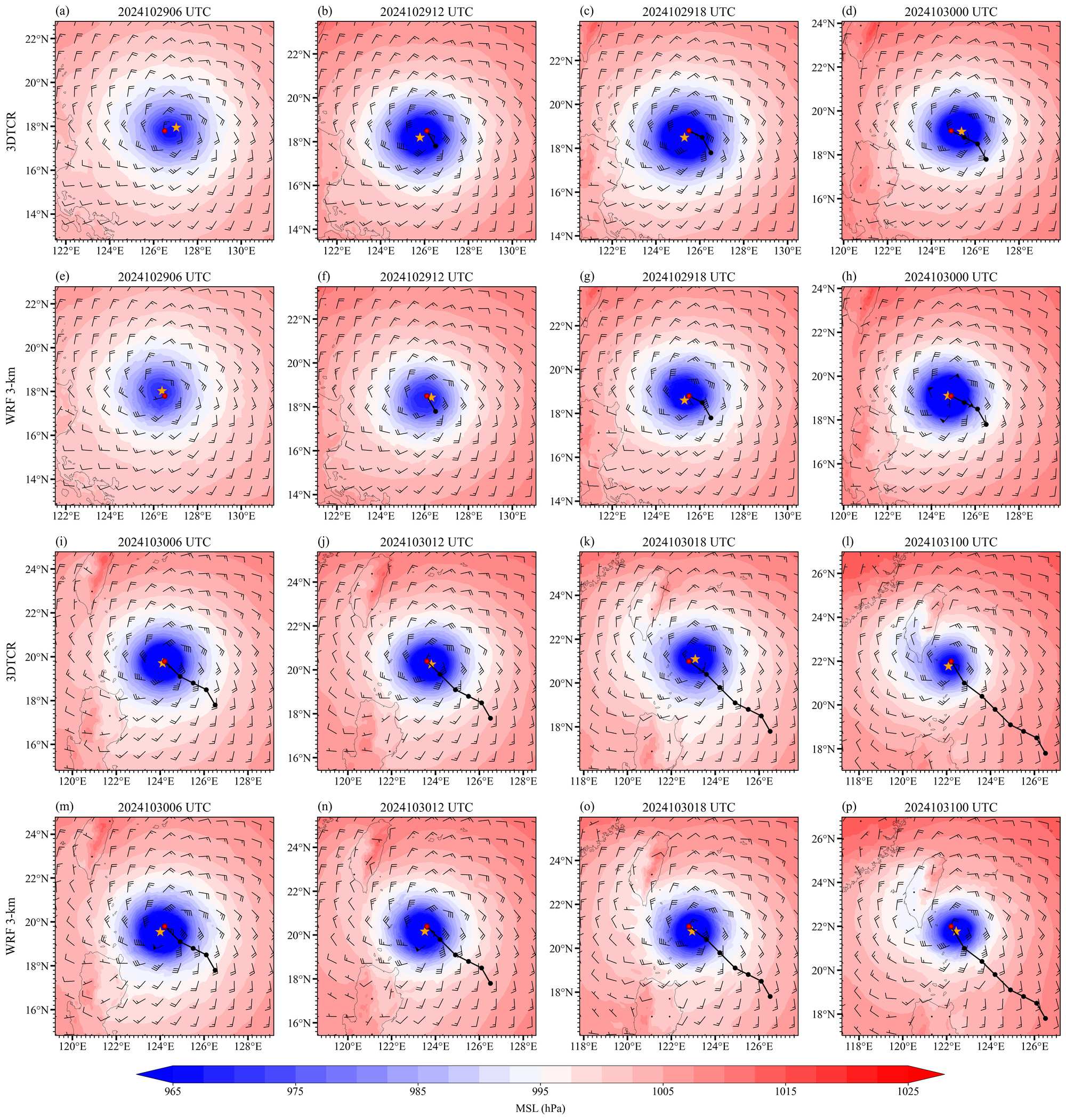}
    \caption{\textbf{Temporal evolution of mean sea level pressure and 10-m wind field for Typhoon KONG-REY (0600 UTC on 29 October to 0000 UTC on 31 October 2024).} Rows 1 and 3 show the 3DTCR results, while rows 2 and 4 show the corresponding WRF 3-km simulation results. Each group of four columns represents consecutive valid times. Color shading represents mean sea level pressure, wind barbs indicate 10-m wind speed and direction, the yellow star marks the predicted TC center, and the black solid line shows the IBTrACS best track. The time interval is 6 hours.}
    \label{fig:KONG-REY_2024-10-29_06-00-00_tc_track_add_msl_barbs.png}
\end{figure*}

\FloatBarrier
\paragraph*{Temporal evolution of mean sea level pressure and near-surface winds.}
Supplementary Fig.~\ref{fig:KONG-REY_2024-10-29_06-00-00_tc_track_add_msl_barbs.png} follows the mean sea level pressure and 10-m wind fields at 6 h intervals. This sequence complements the diagnostics at an individual time above by showing how the reconstructed and WRF reference structures evolve over consecutive valid times.

\FloatBarrier
\subsubsection{Structure and evolution of Typhoon YAGI}

Typhoon YAGI provides a second case for assessing whether the structural behavior identified for KONG-REY is reproduced for a different storm. It follows the same sequence of vertical wind diagnostics, multilevel meteorological fields, a qualitative comparison with satellite cloud patterns, and temporal evolution.

\begin{figure*}[!htbp]
    \centering
    \includegraphics[width=\linewidth]{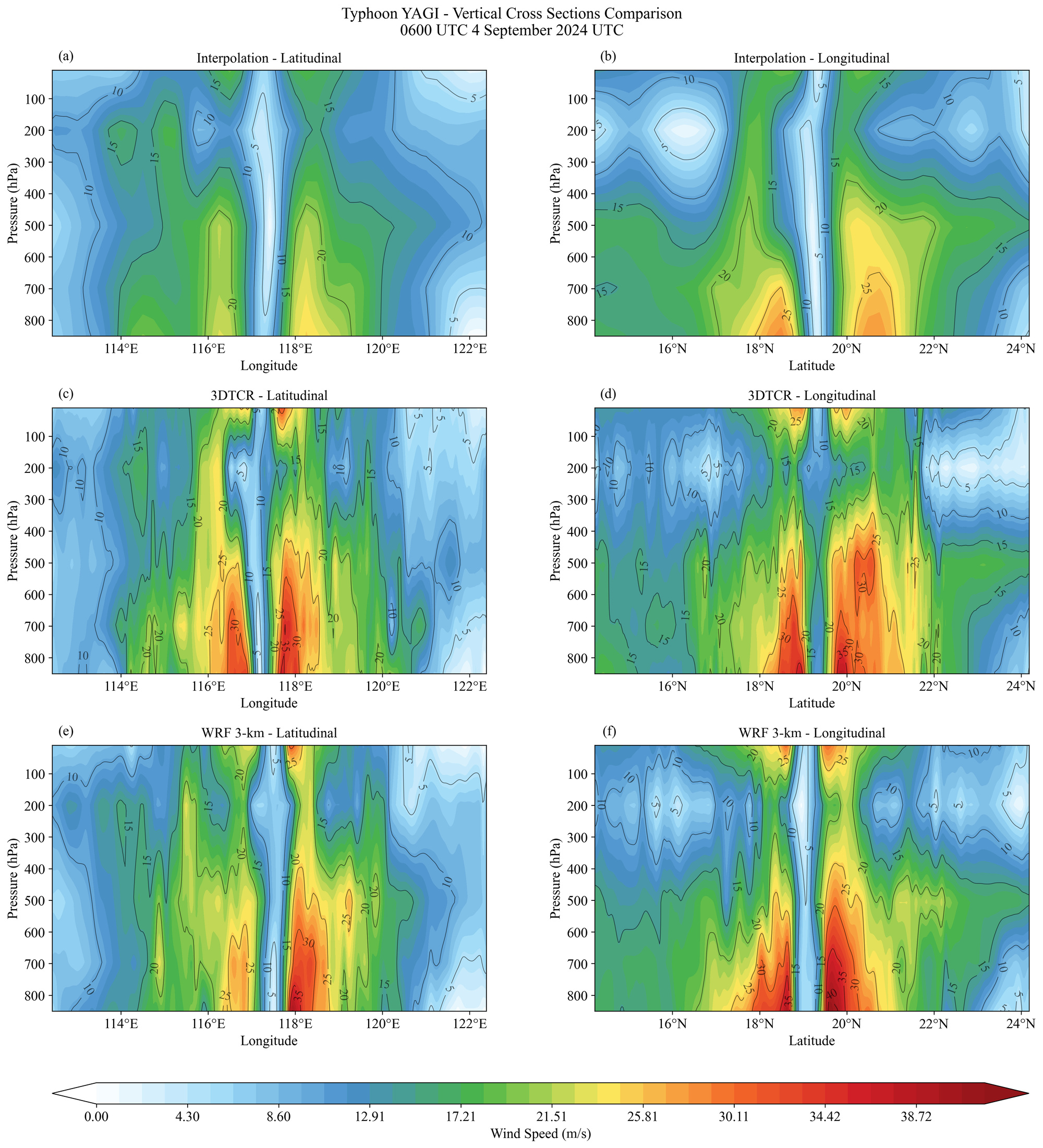}
    \caption{\textbf{Vertical wind structure of Typhoon YAGI at 0600 UTC on 4 September 2024.}
    (a, b) Interpolation, (c, d) 3DTCR, and (e, f) WRF 3-km results.
    The left and right columns show latitudinal and longitudinal cross sections, respectively.
    Shading and contours indicate wind speed (m/s), with pressure levels (hPa) on the vertical axis.
    The central white region and adjacent regions of high wind speed correspond to the eye and eyewall, respectively.}
    \label{fig:YAGI_vertical_cross_sections_all_2024-09-04T06-00-00_black_contour_k.png}
\end{figure*}

\begin{figure*}[!htbp]
    \centering
    \includegraphics[width=\linewidth]{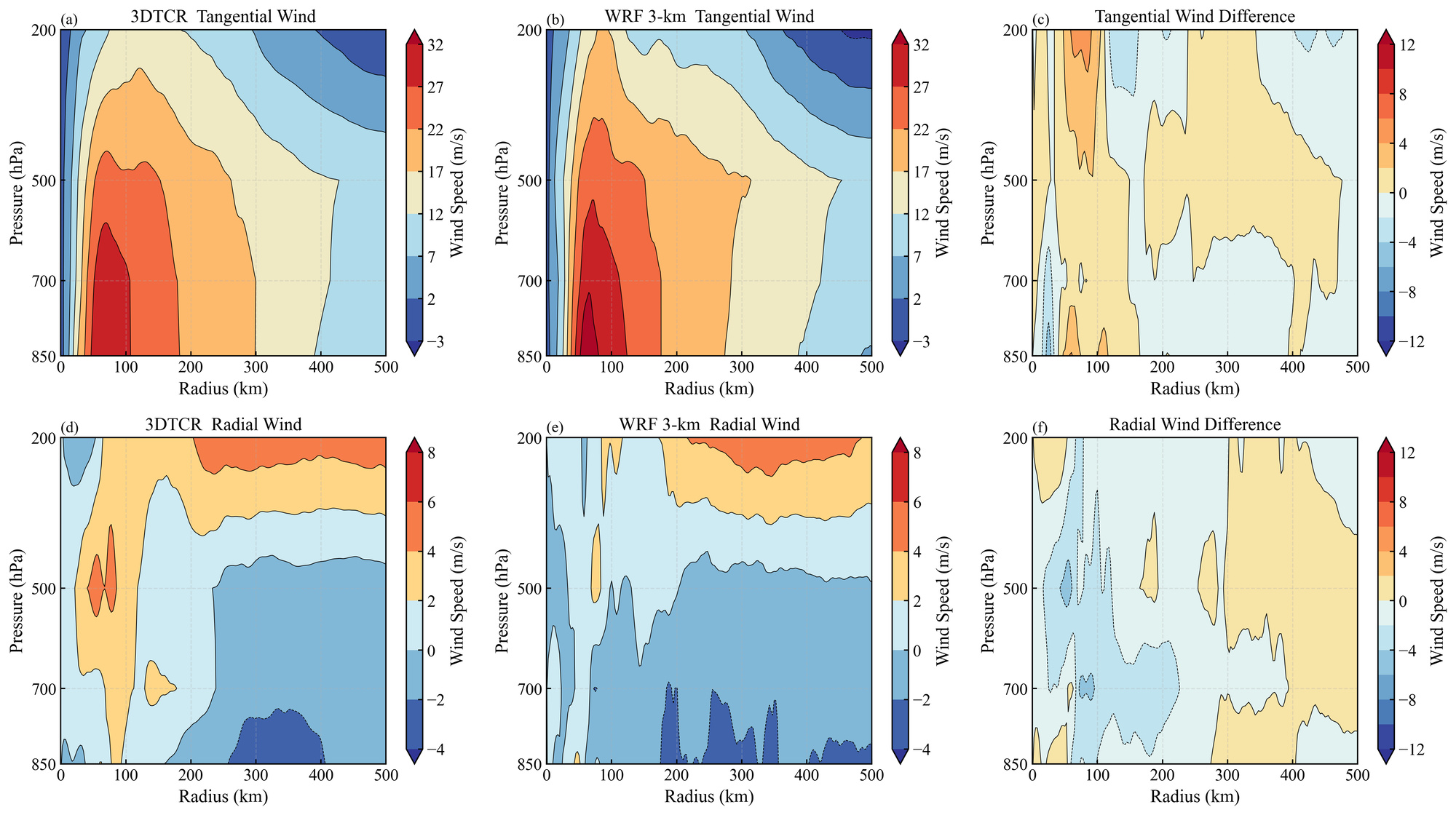}
    \caption{\textbf{Vertical cross sections of tangential and radial wind for Typhoon YAGI (0600 UTC 4 September 2024).} (a) 3DTCR tangential wind; (b) WRF 3-km tangential wind; (c) Tangential wind difference (3DTCR minus WRF 3-km); (d) 3DTCR radial wind; (e) WRF 3-km radial wind; (f) Radial wind difference (3DTCR minus WRF 3-km). The horizontal axis represents the radius from the TC center (km), and the vertical axis represents pressure levels (hPa).}
    \label{fig:YAGI_fuxi_wrf_tangential_radial_winds_diff.png}
\end{figure*}

\paragraph*{Vertical wind structure.}Supplementary Figs.~\ref{fig:YAGI_vertical_cross_sections_all_2024-09-04T06-00-00_black_contour_k.png} and \ref{fig:YAGI_fuxi_wrf_tangential_radial_winds_diff.png} show the latitudinal and longitudinal wind speed cross sections and the azimuthally averaged tangential and radial winds for YAGI. As in KONG-REY, the reconstructed fields contain a vertically organized inner-core structure and secondary circulation that more closely resemble the WRF 3-km reference than the interpolated input.

\begin{figure*}[!htbp]
    \centering
    \includegraphics[width=\linewidth]{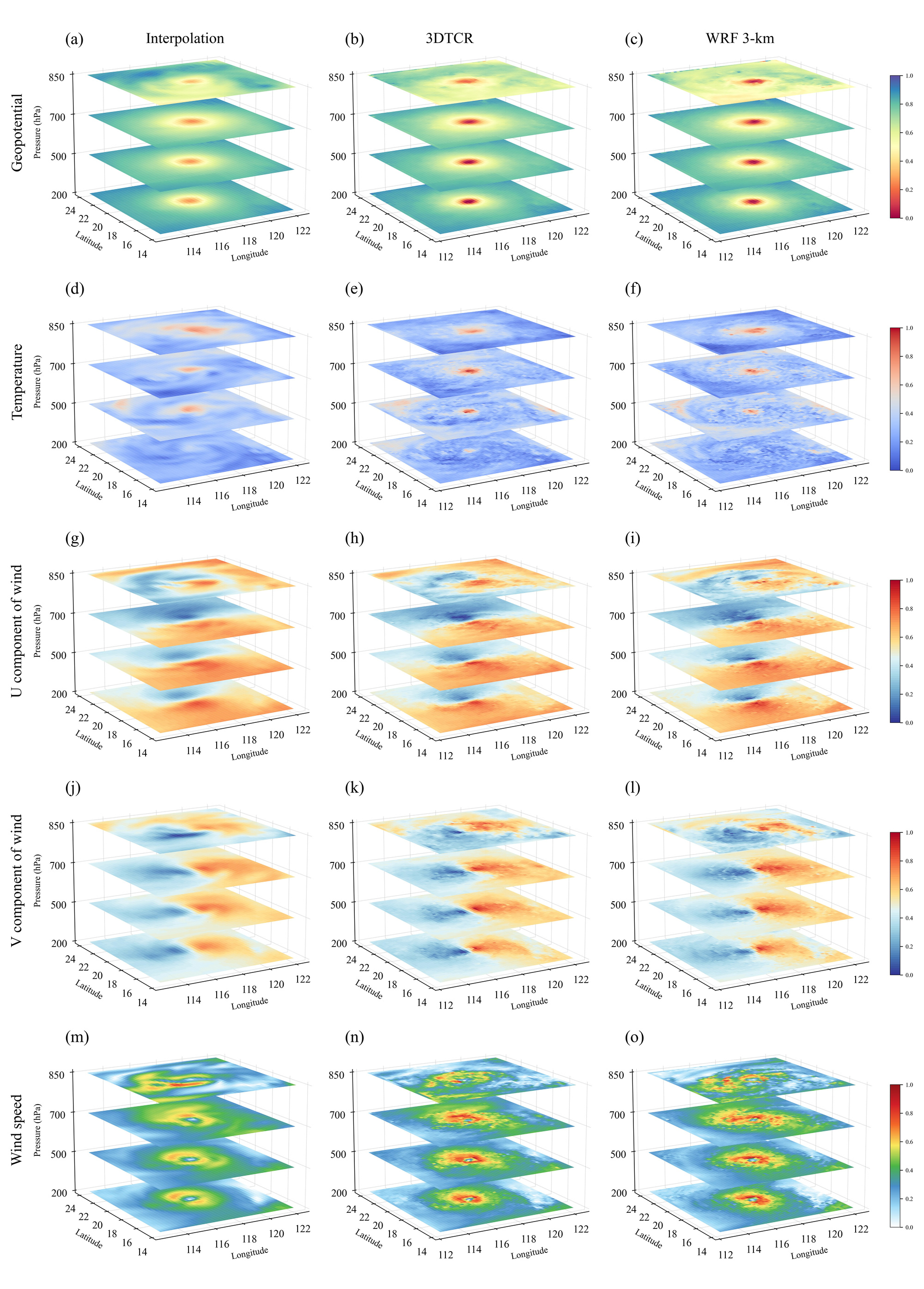}
    \caption{\textbf{Typhoon YAGI three-dimensional multivariable field reconstruction at 18 h lead time.} Comparison across vertical pressure levels ($200$, $500$, $700$, and $850\text{ hPa}$) for geopotential, temperature, u/v wind components, and 10-m surface wind speed, initialized at 1200 UTC 3 September 2024. Columns from left to right: Interpolation (FuXi), 3DTCR reconstruction, and WRF 3-km simulation (reference target).}
    \label{fig:3d_reconstruction_yagi.png}
\end{figure*}

\begin{figure*}[!htbp]
    \centering
    \includegraphics[width=\linewidth]{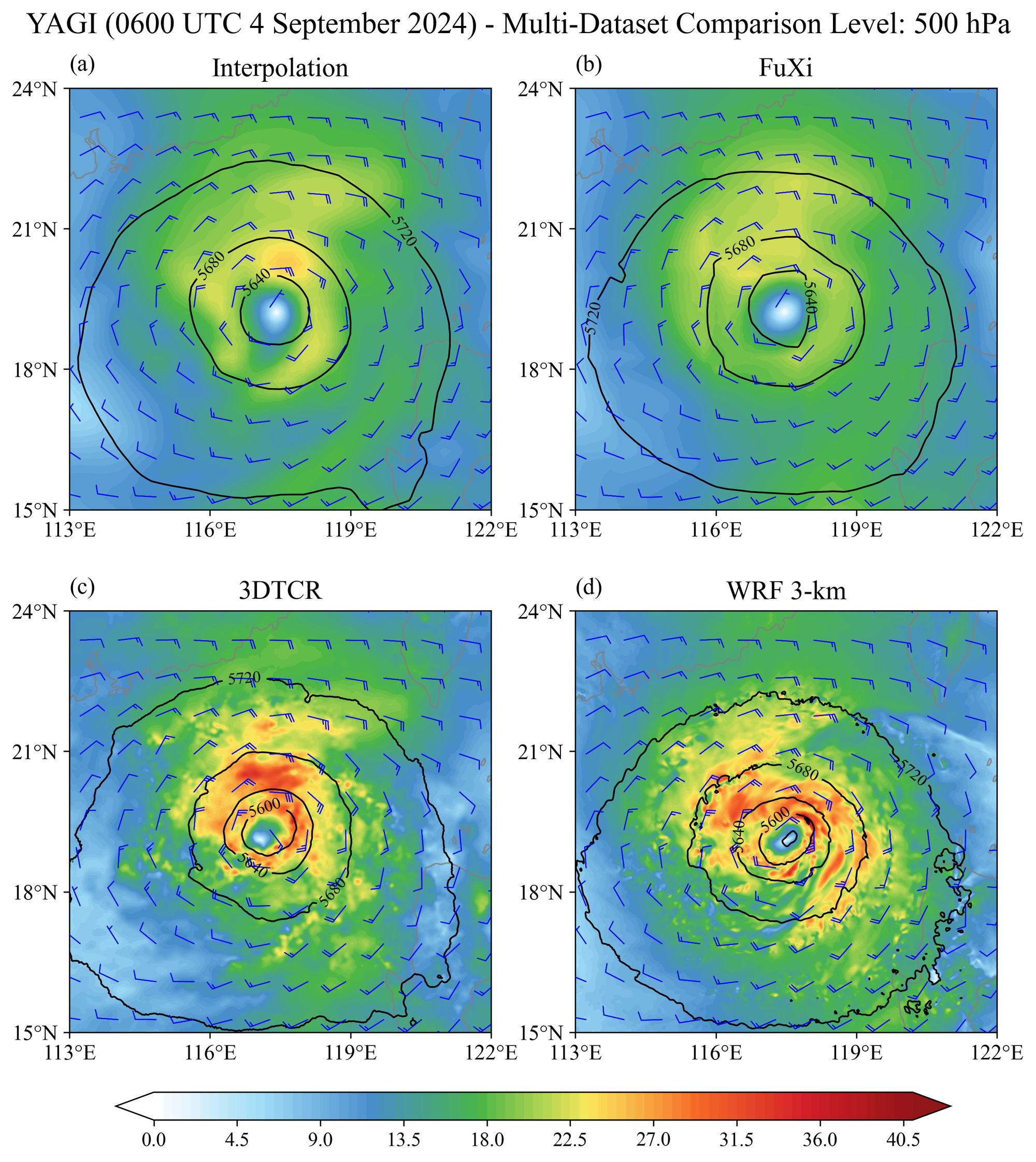}
    \caption{\textbf{Comparison of 500 hPa wind and geopotential height fields for Typhoon YAGI (0600 UTC 4 September 2024).} (a) ERA5 interpolation data, (b) FuXi forecast interpolation data, (c) 3DTCR, (d) WRF 3-km. Color shading represents wind speed (m/s), black contours show geopotential height (gpm), and wind barbs indicate the 500 hPa wind field.}
    \label{fig:YAGI_circulation_500.png}
\end{figure*}

\begin{figure*}[!htbp]
    \centering
    \includegraphics[width=\linewidth]{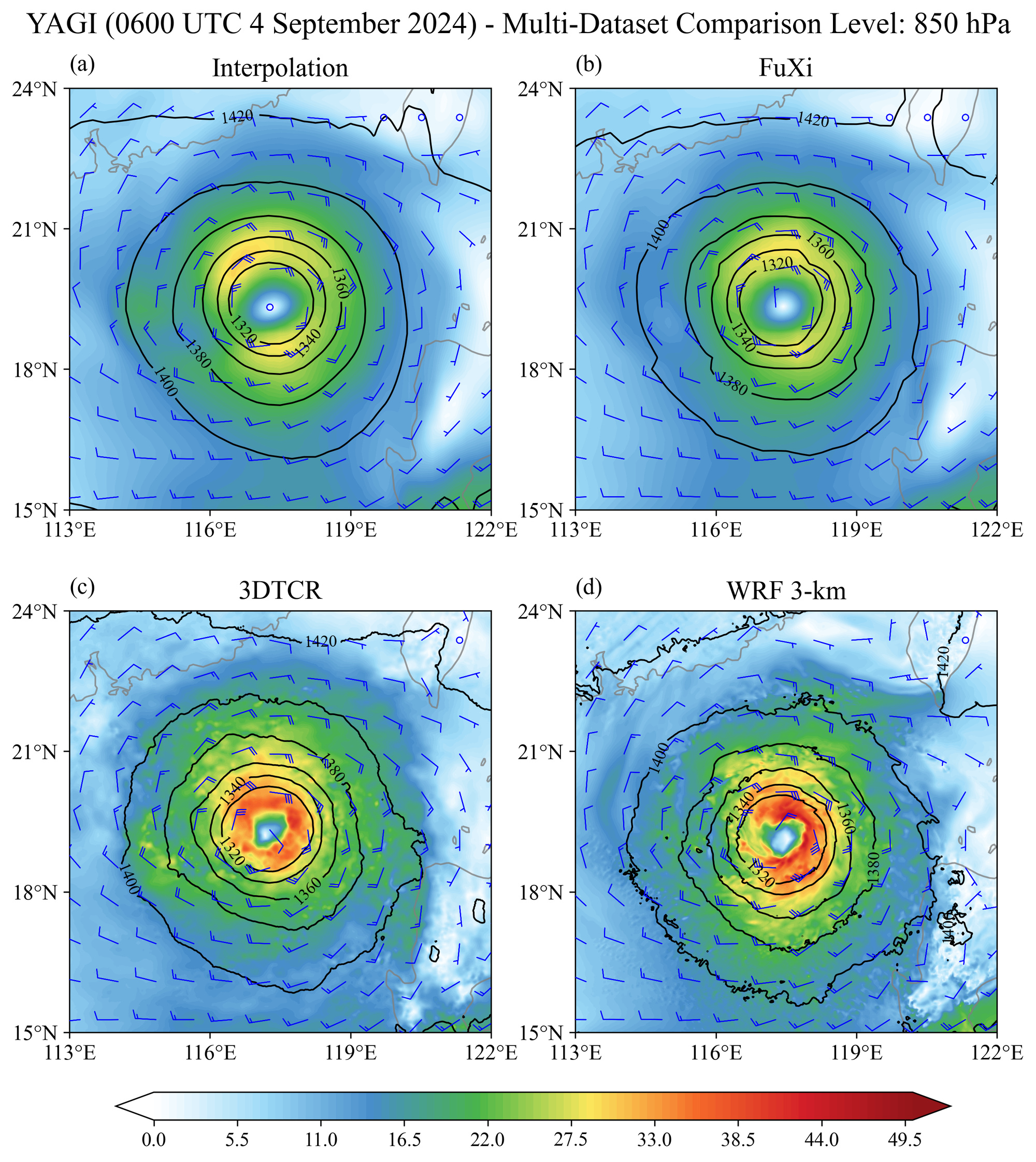}
    \caption{\textbf{Comparison of 850 hPa wind and geopotential height fields for Typhoon YAGI (0600 UTC 4 September 2024).} (a) ERA5 interpolation data, (b) FuXi forecast interpolation data, (c) 3DTCR, (d) WRF 3-km. Color shading represents wind speed (m/s), black contours show geopotential height (gpm), and wind barbs indicate the 850 hPa wind field.}
    \label{fig:YAGI_circulation_850.png}
\end{figure*}

\begin{figure*}[!htbp]
    \centering
    \includegraphics[width=\linewidth]{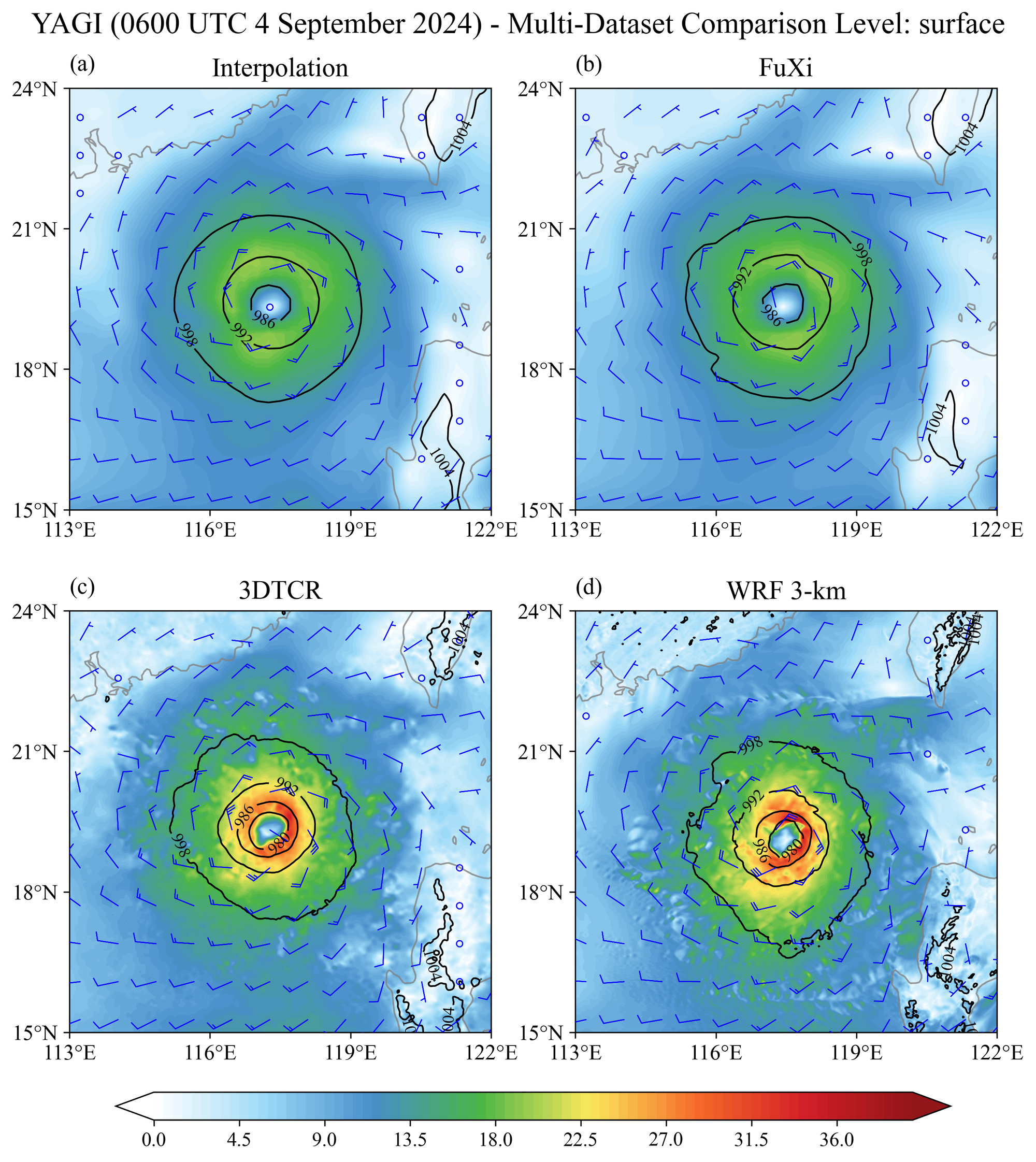}
    \caption{\textbf{Comparison of near-surface wind and mean sea level pressure fields for Typhoon YAGI (0600 UTC 4 September 2024).} (a) ERA5 interpolation data, (b) FuXi forecast interpolation data, (c) 3DTCR, (d) WRF 3-km. Color shading represents wind speed (m/s), black
    contours show mean sea level pressure (hPa), and wind barbs indicate the 10-m wind field.}
    \label{fig:YAGI_circulation_surface.png}
\end{figure*}

\FloatBarrier
\paragraph*{Multilevel structure of meteorological fields.}
Supplementary Figs.~\ref{fig:3d_reconstruction_yagi.png}, \ref{fig:YAGI_circulation_500.png}, \ref{fig:YAGI_circulation_850.png}, and \ref{fig:YAGI_circulation_surface.png} provide the corresponding multilevel comparison for YAGI, including temperature, geopotential, wind, and mean sea level pressure fields.

\begin{figure}[H]
\centering
\includegraphics[width=\linewidth]{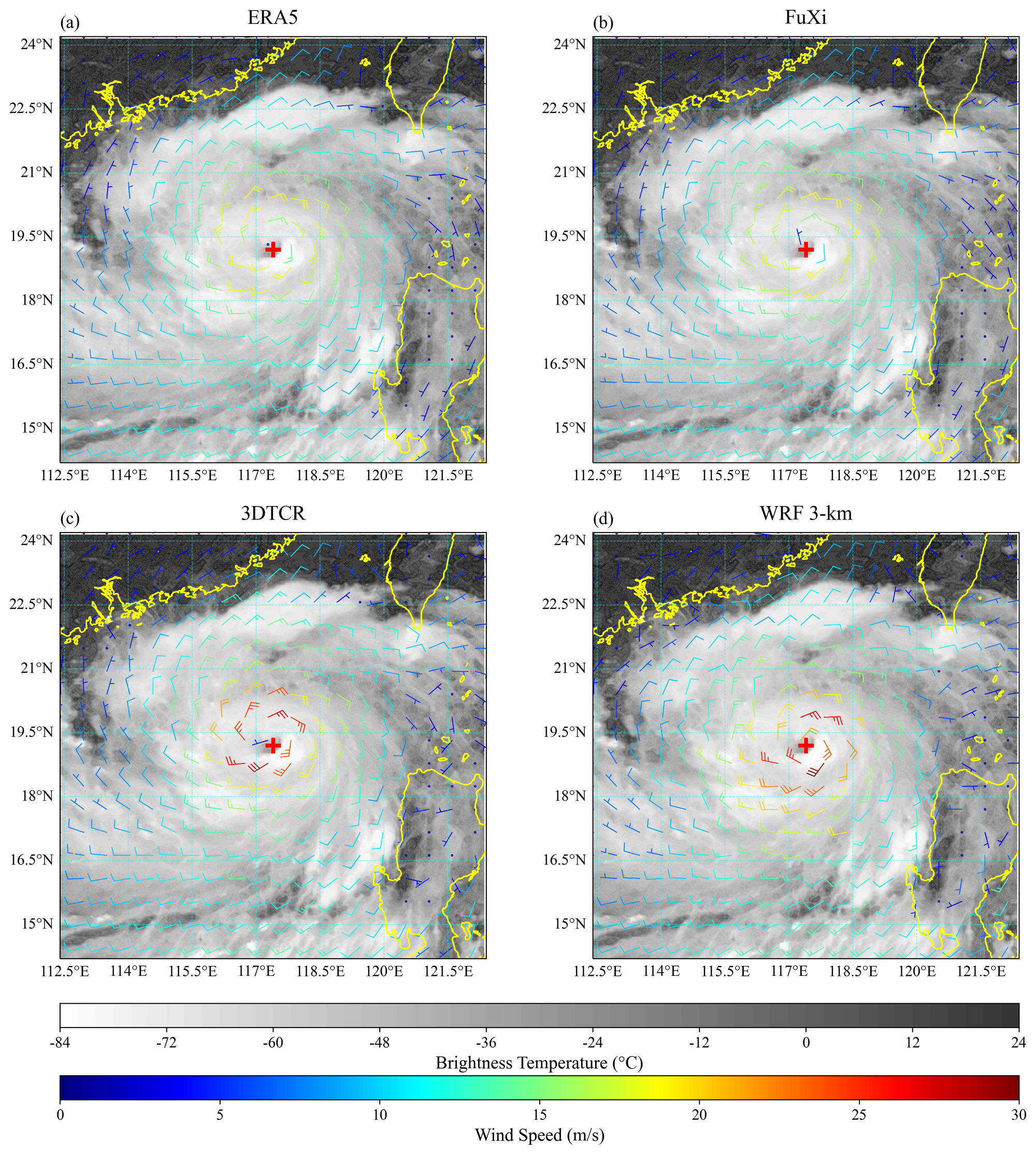}
\caption{\textbf{Comparison of cloud-top brightness temperature and 10-m wind fields for Typhoon YAGI (0600 UTC 4 September 2024).}
(a) ERA5; (b) FuXi; (c) 3DTCR; (d) WRF 3-km simulation.
Grayscale shading shows cloud-top brightness temperature (°C), colored wind barbs indicate 10-m wind speed (m/s) and direction, yellow lines denote coastlines, and red plus signs mark the IBTrACS best-track position.}
\label{fig:YAGI_Cloud_Wind_barbs.png}
\end{figure}

\FloatBarrier
\paragraph*{Comparison of surface winds with satellite cloud patterns.}
Supplementary Fig.~\ref{fig:YAGI_Cloud_Wind_barbs.png} provides the same qualitative comparison for YAGI. As for KONG-REY, the cloud-top brightness temperature describes the observed cloud pattern and is not treated as a direct measurement of the 10-m wind field.

\begin{figure}[H]
    \centering
    \includegraphics[width=\linewidth]{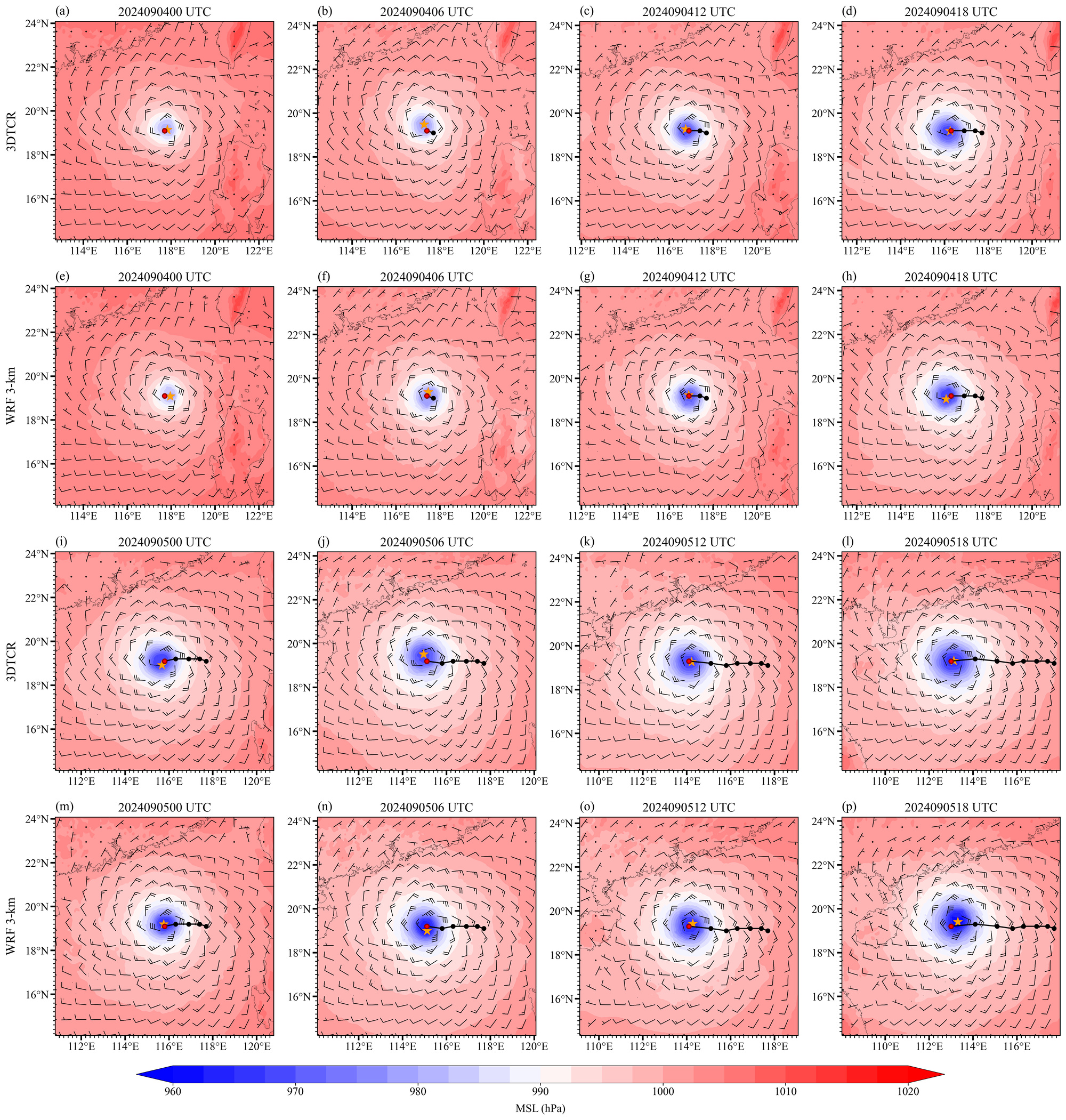}
    \caption{\textbf{Temporal evolution of mean sea level pressure and 10-m wind field for Typhoon YAGI (from 0000 UTC 4 September to 1800 UTC 5 September 2024).} Rows 1 and 3 show the 3DTCR results, while rows 2 and 4 show the corresponding WRF 3-km simulation results. Panels (a--p) show the temporal evolution at consecutive valid times. Color shading represents mean sea level pressure, wind barbs indicate 10-m wind speed and direction, the yellow star marks the predicted TC center, and the black solid line shows the IBTrACS best track. The time interval is 6 hours.}
    \label{fig:YAGI_2024-09-04_00-00-00_tc_track_add_msl_barbs.png}
\end{figure}

\FloatBarrier
\paragraph*{Temporal evolution of mean sea level pressure and near-surface winds.}
Supplementary Fig.~\ref{fig:YAGI_2024-09-04_00-00-00_tc_track_add_msl_barbs.png} follows the corresponding YAGI fields at 6 h intervals, allowing the temporal behavior of the reconstruction to be compared with the WRF reference using the same diagnostic as for KONG-REY.

\FloatBarrier
\FloatBarrier
\subsubsection{Transfer to other basins}
\label{sec:cross_basin_generalization}

\begin{figure}[!htbp]
    \centering
    \includegraphics[width=1.0\linewidth]{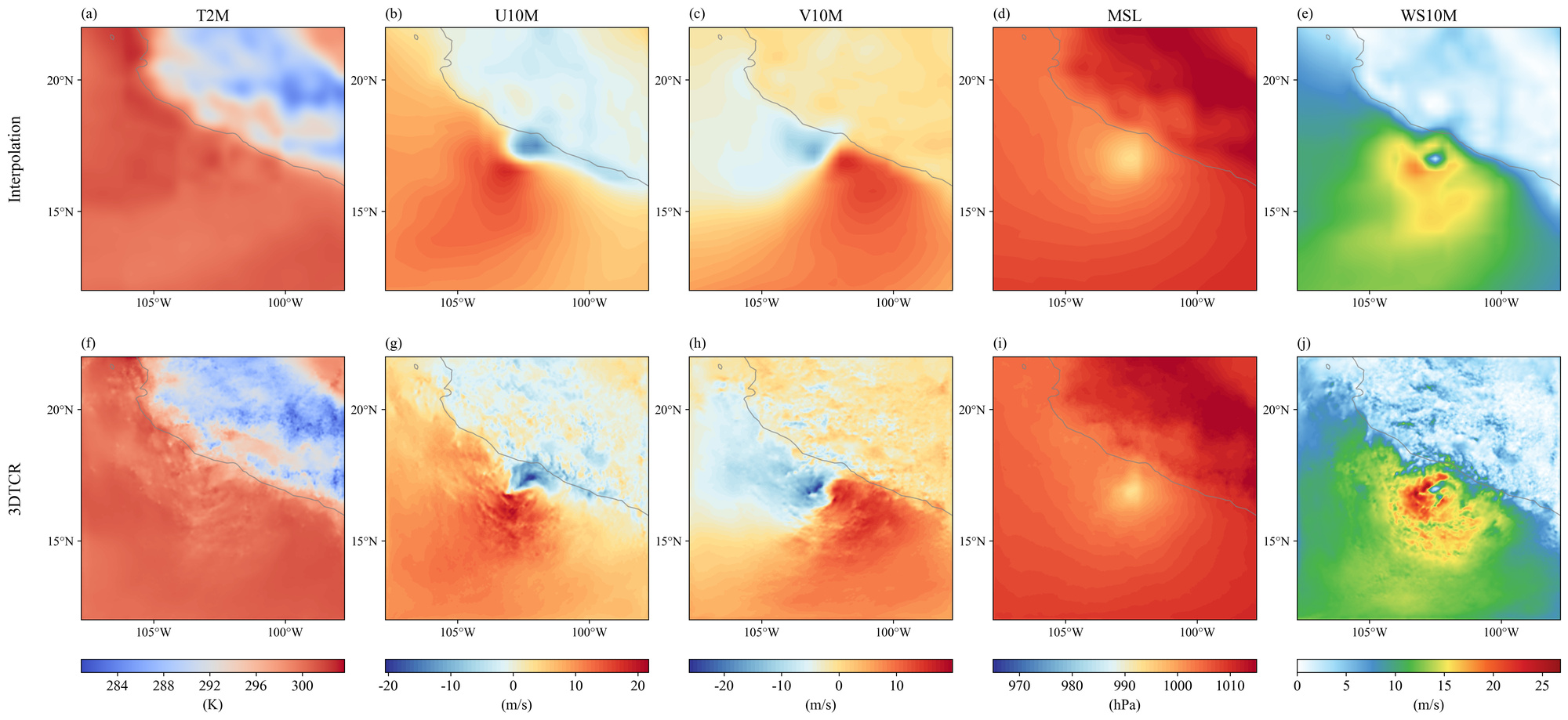}
    \caption{\textbf{Multivariable TC vortex reconstruction for Tropical Storm JOHN over the eastern Pacific at a 12 h forecast lead time.} The forecast was initialized at 0000 UTC on 26 September 2024. Columns show 2-m temperature (T2M), 10-m u and v wind components (U10M and V10M), mean sea level pressure (MSL), and 10-m wind speed (WS10M). Rows show bilinearly interpolated FuXi fields and the 3DTCR reconstruction.}
    \label{fig:JOHN_EP_init2024092600_lead2_valid2024092612}
\end{figure}

\begin{figure*}[!htbp]
    \centering
    \includegraphics[width=1.0\linewidth]{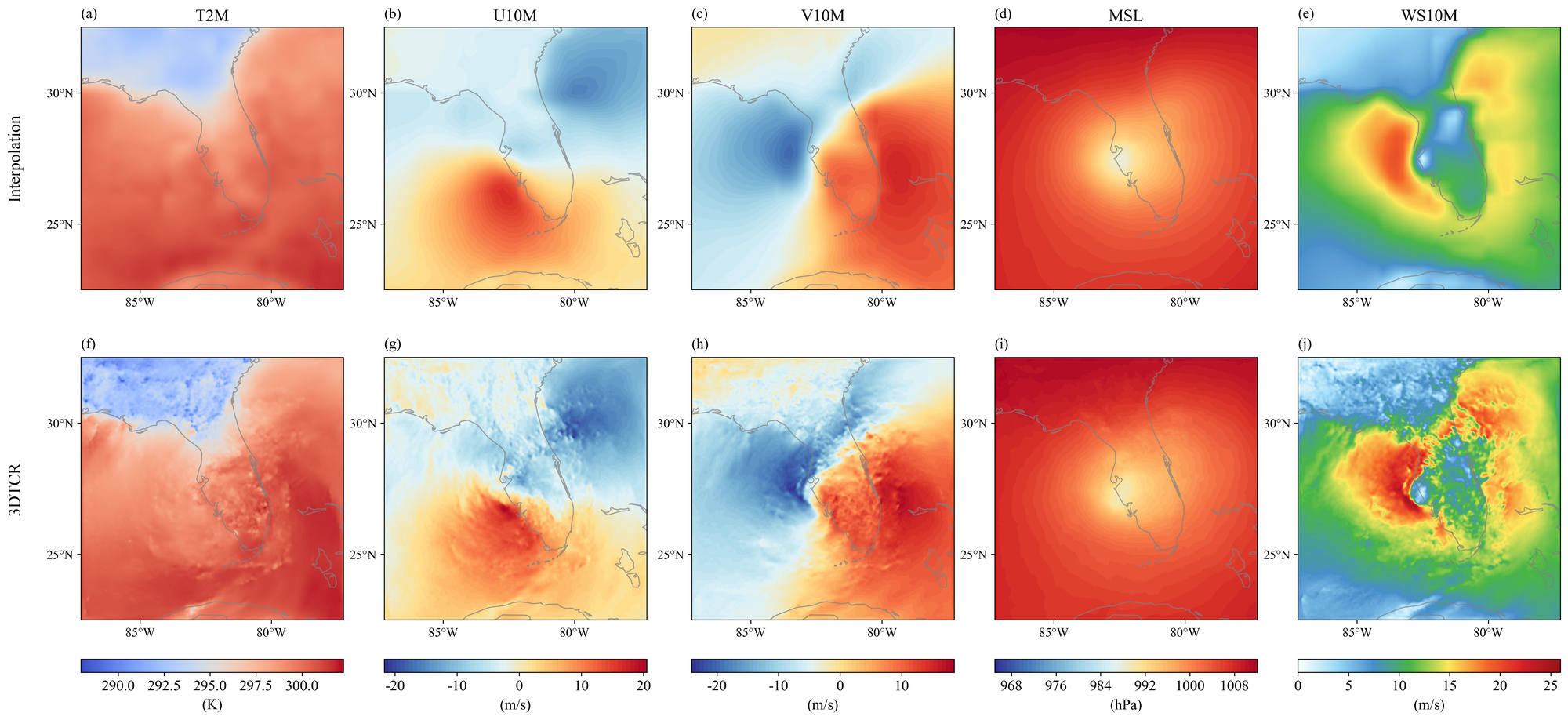}
    \caption{\textbf{Multivariable TC vortex reconstruction for Hurricane MILTON over the North Atlantic at a 114 h forecast lead time.} The forecast was initialized at 1200 UTC 5 October 2024. Columns show 2-m temperature (T2M), 10-m u and v wind components (U10M and V10M), mean sea level pressure (MSL), and 10-m wind speed (WS10M). Rows show bilinearly interpolated FuXi fields and the 3DTCR reconstruction.}
    \label{fig:MILTON_NA_init2024100512_lead19_valid2024101006}
\end{figure*}

\begin{figure*}[!htbp]
    \centering
    \includegraphics[width=1.0\linewidth]{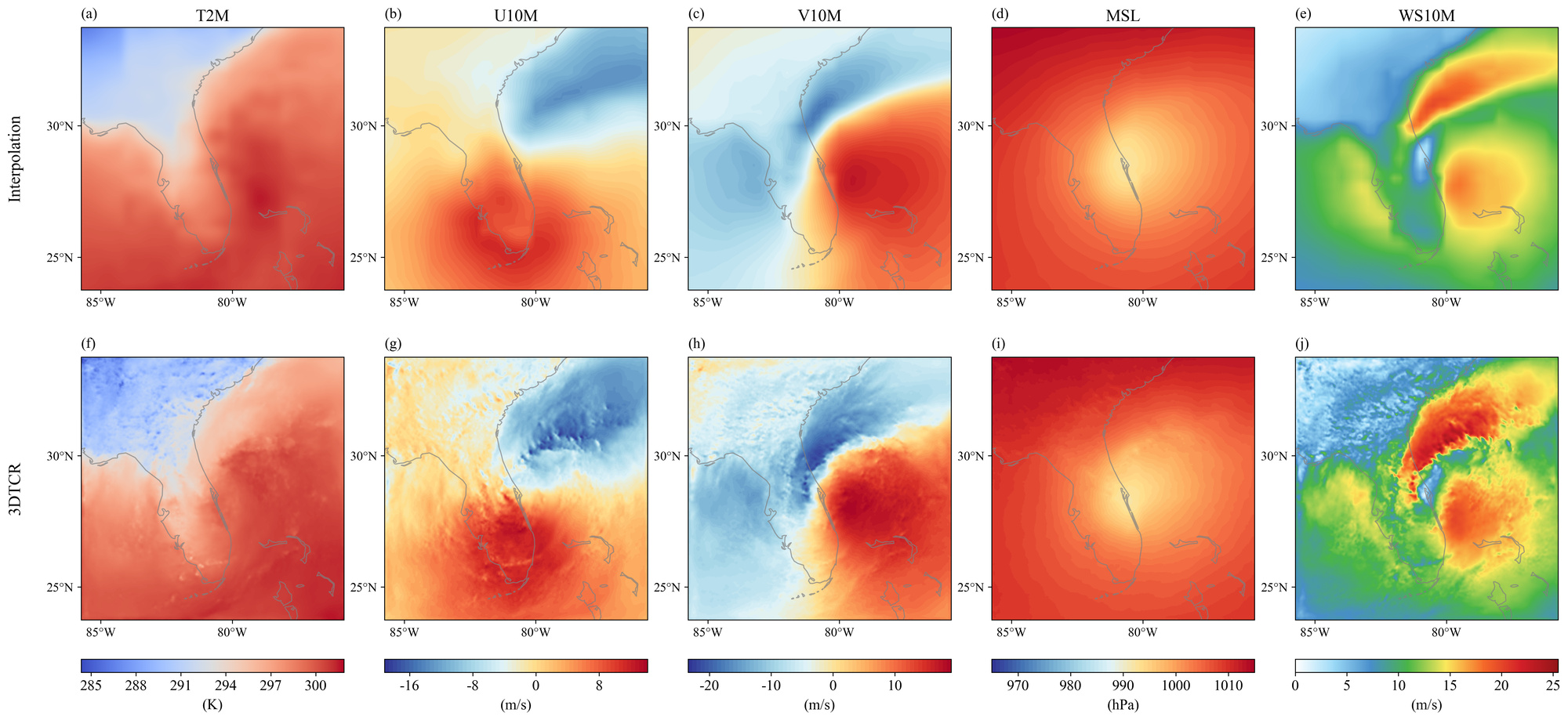}
    \caption{\textbf{Multivariable TC vortex reconstruction for Hurricane MILTON over the North Atlantic at a 120 h forecast lead time.} The forecast was initialized at 1200 UTC 5 October 2024. Columns show 2-m temperature (T2M), 10-m u and v wind components (U10M and V10M), mean sea level pressure (MSL), and 10-m wind speed (WS10M). Rows show bilinearly interpolated FuXi fields and the 3DTCR reconstruction.}
    \label{fig:MILTON_NA_init2024100512_lead20_valid2024101012}
\end{figure*}

As described in the main text, 3DTCR was trained exclusively on WNP TC cases from WRF 3-km simulations spanning 2018--2023. We apply the full model without retraining to two representative cases from basins excluded from training: Tropical Storm JOHN in the eastern Pacific (initialized 0000 UTC 26 September 2024; Supplementary Fig.~\ref{fig:JOHN_EP_init2024092600_lead2_valid2024092612}) and Hurricane MILTON in the North Atlantic (initialized 1200 UTC 5 October 2024, evaluated at 114 h and 120 h lead times; Supplementary Figs.~\ref{fig:MILTON_NA_init2024100512_lead19_valid2024101006} and~\ref{fig:MILTON_NA_init2024100512_lead20_valid2024101012}).

As shown in these supplementary figures, the full 3DTCR model generates localized inner-core structures from the corresponding FuXi inputs, with enhanced near-surface intensity relative to the coarse FuXi fields. This behavior is qualitatively consistent with the WNP examples in the main text. Because no WRF 3-km simulations or equivalent high-resolution reference fields are available for JOHN and MILTON, these cases provide qualitative evidence of transfer and do not establish quantitative cross-basin generalization.

\FloatBarrier

\FloatBarrier
\subsubsection{Spectral and harmonic analysis}

\begin{figure}[!htbp]
    \centering
    \includegraphics[width=\linewidth]{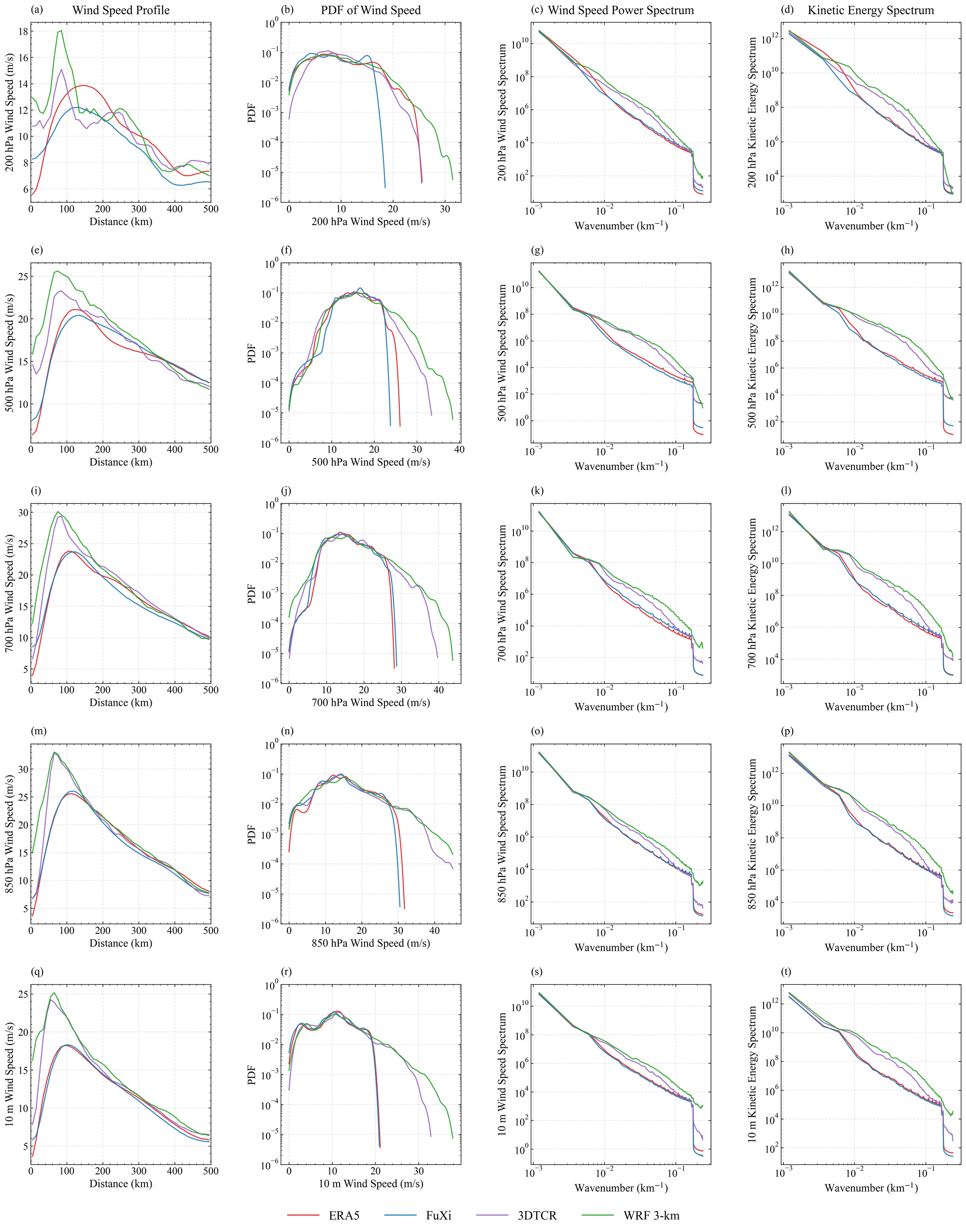}
    \caption{\textbf{Multilevel wind speed distributions and spectra for Typhoon YAGI.}
    From top to bottom, rows show winds at 200, 500, 700 and 850 hPa and at 10 m.
    From left to right, columns show radial wind speed profiles, probability density functions, wind speed power spectra and kinetic energy spectra.
    Results are shown for ERA5 (red), FuXi (blue), 3DTCR (purple) and WRF 3-km (green).
    The wind speed and kinetic energy spectra are plotted on log--log axes.}
    \label{fig:YAGI_comprehensive_multi_level_ws_energy.png}
\end{figure}

\paragraph*{Wind speed distributions and spectra.}
Supplementary Fig.~\ref{fig:YAGI_comprehensive_multi_level_ws_energy.png} repeats for YAGI the multilevel wind speed distributions and spectral diagnostics presented for KONG-REY in the main text. As in KONG-REY, inputs at coarse resolution have narrower distributions and lose energy more rapidly at high wavenumbers. The 3DTCR results show a broader distribution at high wind speeds and retain more variability at small scales, bringing them closer to the WRF 3-km reference. This comparison offers a complementary view of wind structure from the surface to the upper troposphere.

\begin{figure*}[!htbp]
    \centering
    \includegraphics[width=\linewidth]{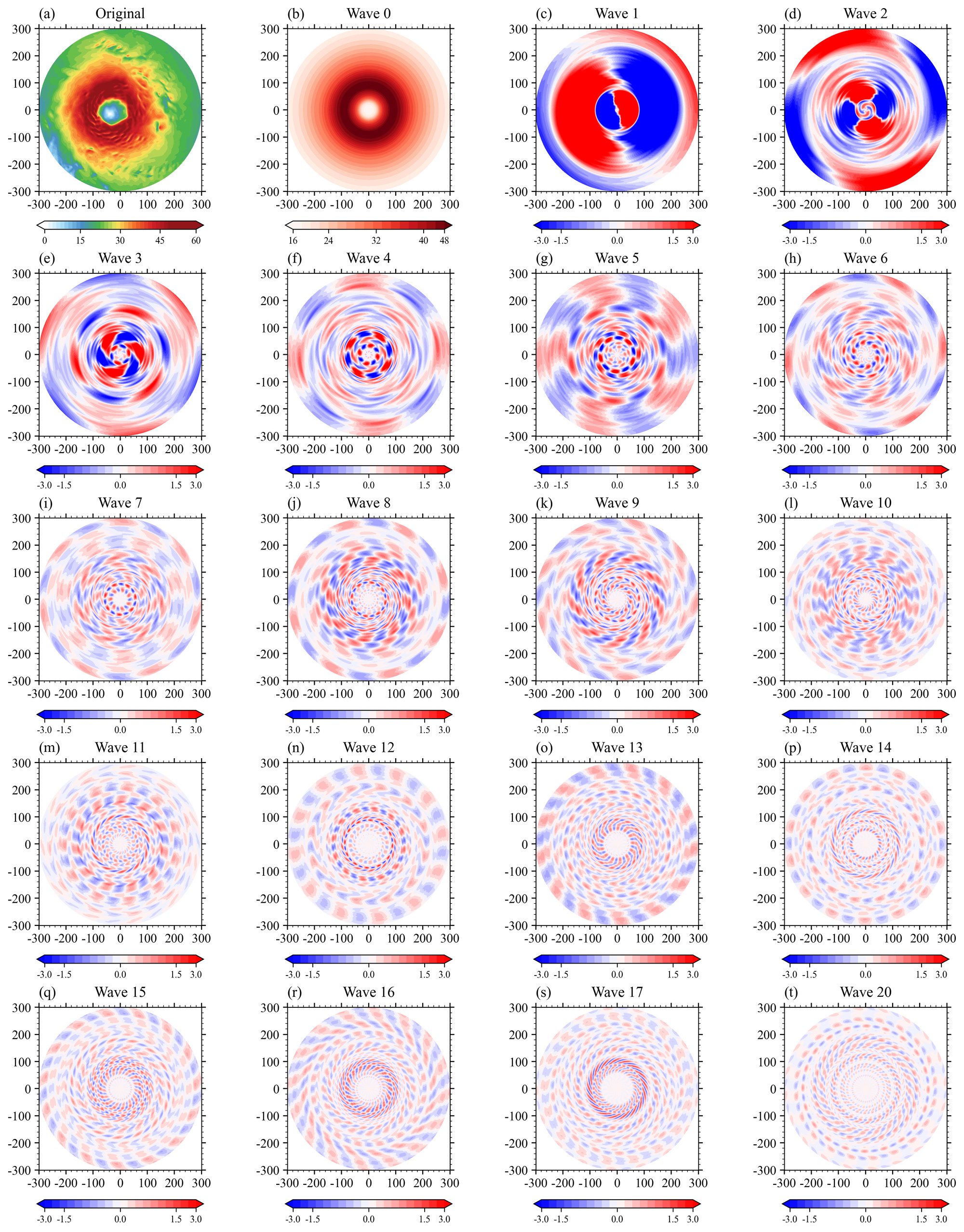}
    \caption{\textbf{Wavenumber decomposition of wind speed for Typhoon KONG-REY from WRF 3-km (0600 UTC on 30 October 2024).} (a) Original wind speed; (b)--(s) wavenumbers 0--17; (t) wavenumber 20. Color shading shows wind speed (m/s). Axes show distance from the center (km).}
    \label{fig:WRF3km_KONG-REY_wave_decomposition_5x4.png}
\end{figure*}

\begin{figure*}[!htbp]
    \centering
    \includegraphics[width=\linewidth]{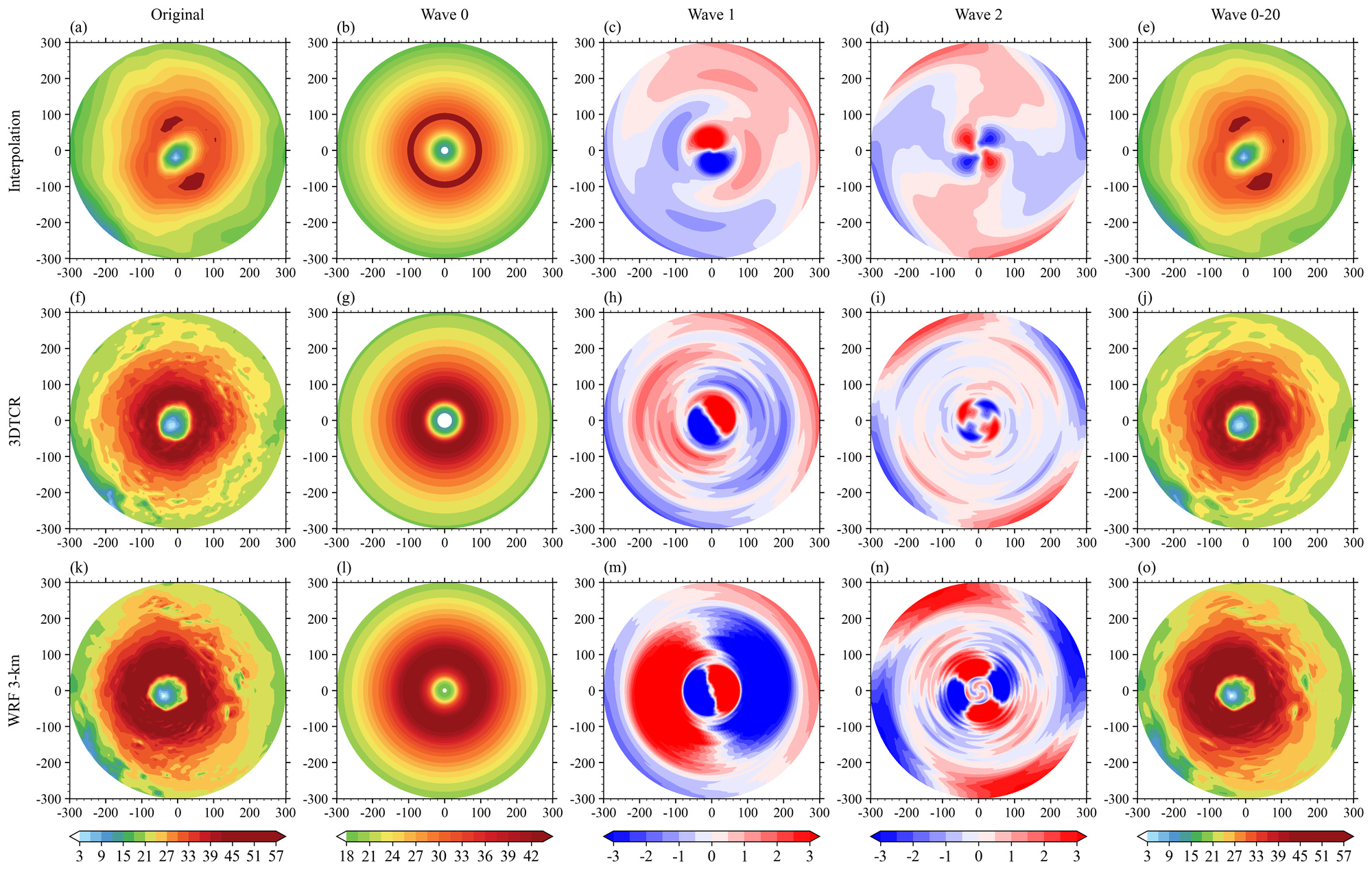}
    \caption{\textbf{Wavenumber decomposition of wind speed for Typhoon KONG-REY (0600 UTC 30 October 2024).} The first through fifth columns show the original wind speed, wavenumber 0 (axisymmetric component), wavenumber 1 (dipole), wavenumber 2 (quadrupole), and the wavenumber 0--20 synthesis. The first row shows interpolation, the second row shows 3DTCR, and the third row shows WRF 3-km. Color shading represents wind speed (m/s), and axes indicate distance from the TC center (km). All data are interpolated to WRF 3-km resolution and transformed to polar coordinates centered on the TC center with a radius of 300 km; a Fourier transform is applied to the azimuthal data at each radial distance to extract the wavenumber components.}
    \label{fig:KONG-REY_wind_speed_wave_decomposition_20.png}
\end{figure*}

\begin{figure*}[!htbp]
    \centering
    \includegraphics[width=\linewidth]{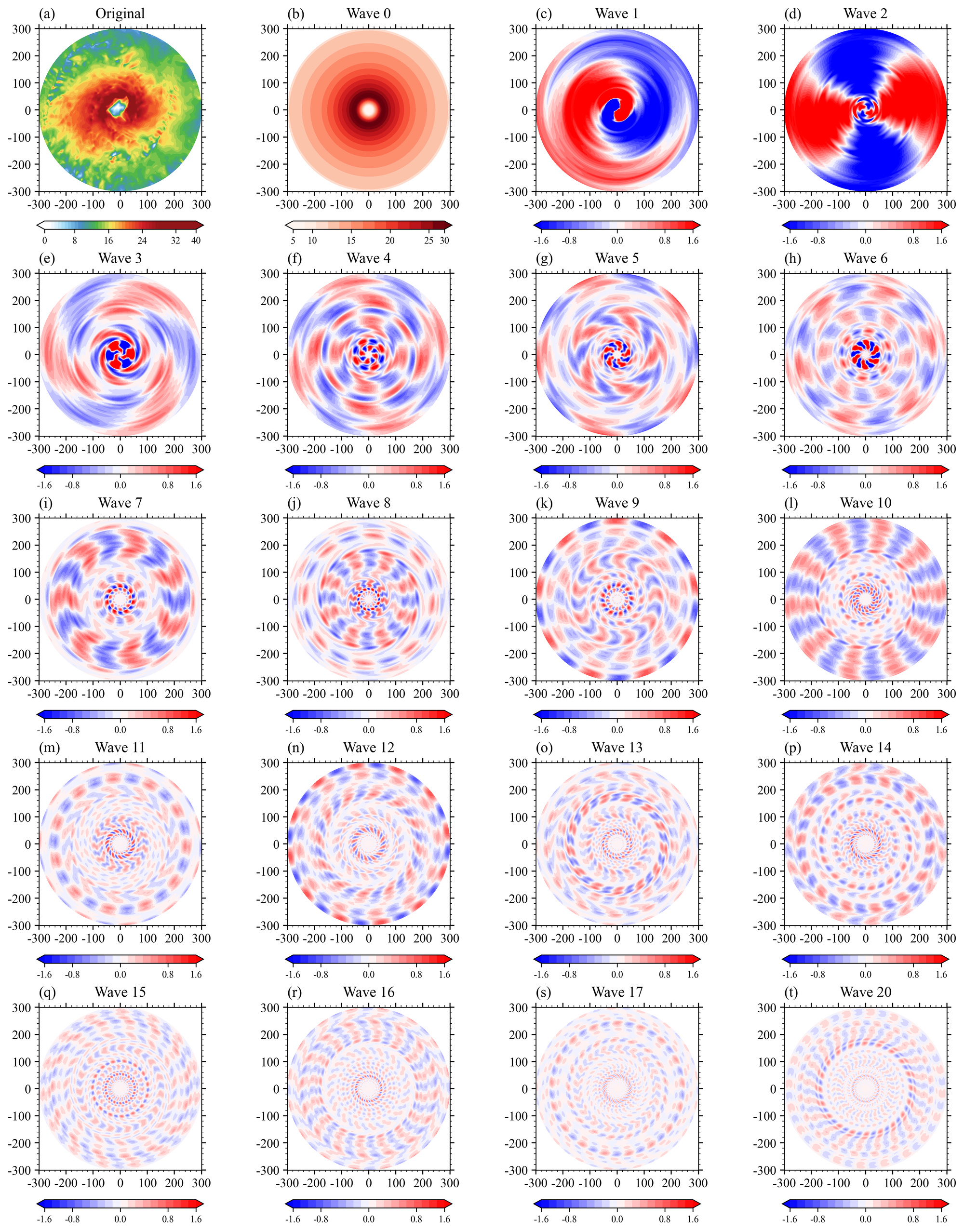}
    \caption{\textbf{Wavenumber decomposition of wind speed for Typhoon YAGI from WRF 3-km (0600 UTC on 4 September 2024).} (a) Original wind speed; (b)--(s) wavenumbers 0--17; (t) wavenumber 20. Color shading shows wind speed (m/s). Axes show distance from center (km).}
    \label{fig:WRF3km_YAGI_wave_decomposition_5x4.png}
\end{figure*}

\begin{figure*}[!htbp]
    \centering
    \includegraphics[width=\linewidth]{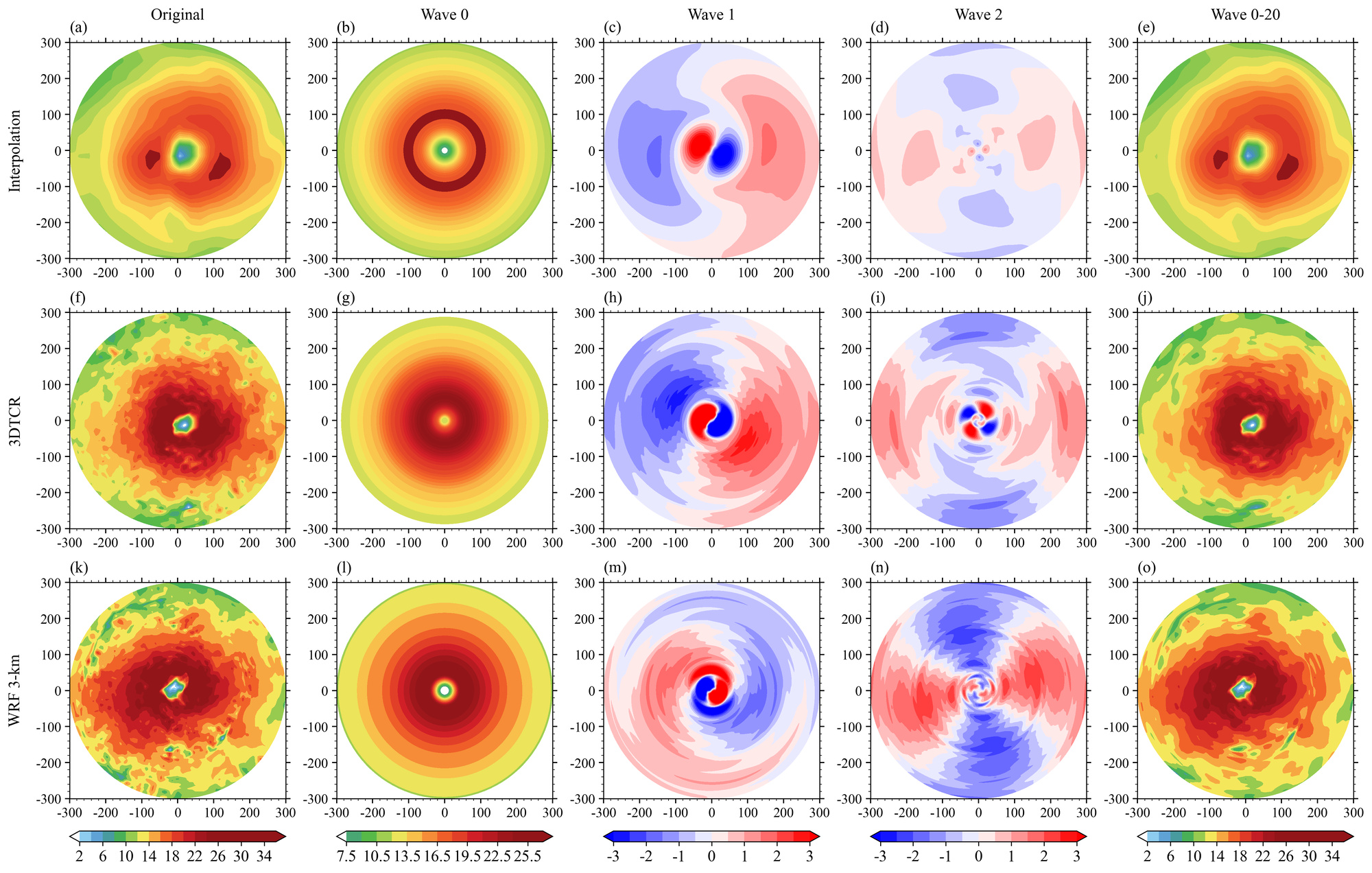}
    \caption{\textbf{Wavenumber decomposition of wind speed for Typhoon YAGI (0600 UTC 4 September 2024).} The first through fifth columns show the original wind speed, wavenumber 0 (axisymmetric component), wavenumber 1 (dipole), wavenumber 2 (quadrupole), and the wavenumber 0--20 synthesis. The first row shows interpolation, the second row shows 3DTCR, and the third row shows WRF 3-km. Color shading represents wind speed (m/s), and axes indicate distance from the TC center (km). All data are interpolated to WRF 3-km resolution and transformed to polar coordinates centered on the TC center with a radius of 300 km; a Fourier transform is applied to the azimuthal data at each radial distance to extract the wavenumber components.}
    \label{fig:YAGI_wind_speed_wave_decomposition_20.png}
\end{figure*}

\FloatBarrier
\paragraph*{Azimuthal harmonic analysis.}
Supplementary Figs.~\ref{fig:WRF3km_KONG-REY_wave_decomposition_5x4.png} and \ref{fig:KONG-REY_wind_speed_wave_decomposition_20.png} present the WRF reference and azimuthal wavenumber components across datasets for KONG-REY. Supplementary Figs.~\ref{fig:WRF3km_YAGI_wave_decomposition_5x4.png} and \ref{fig:YAGI_wind_speed_wave_decomposition_20.png} provide the same analysis for YAGI. Consistent with the main text, the interpolated fields are dominated by wavenumber 0 and contain weak components at higher wavenumbers, whereas 3DTCR strengthens wavenumbers 1--20 toward the WRF 3-km reference. The two cases therefore assess both the symmetric vortex and the asymmetric structures attenuated in the coarse-resolution inputs.

\FloatBarrier
\subsection{Ablation of two-stage training and latent space alignment}

\begin{figure}[!htbp]
    \centering
    \includegraphics[width=1.0\linewidth]{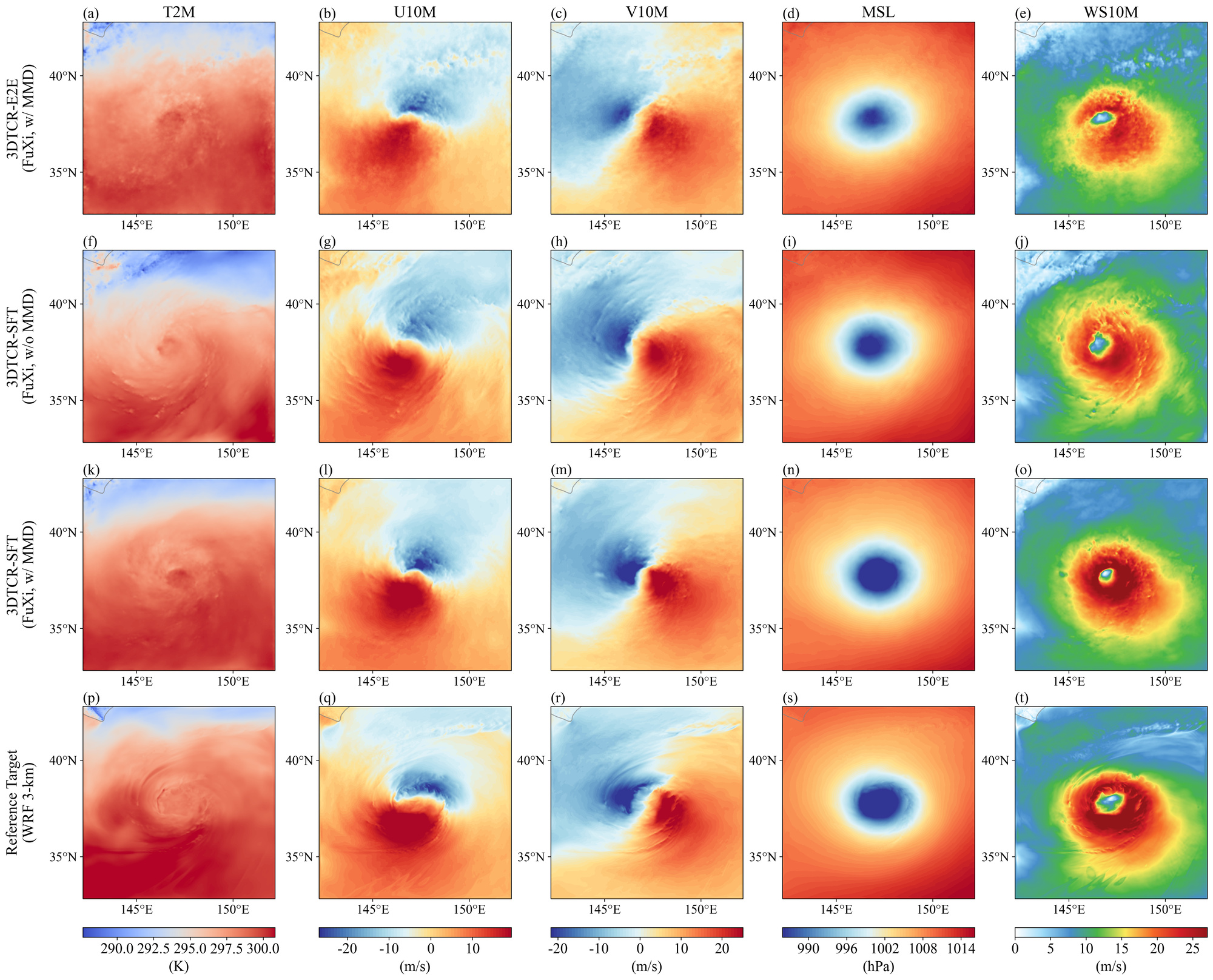}
    \caption{\textbf{Comparison of multivariable TC vortex fields for Typhoon AMPIL across different ablation variants.}
    The forecasts are initialized at 1200 UTC 14 August 2024 with a lead time of 72 h.
    Columns show T2M, U10M, V10M, MSL, and WS10M.
    Rows show 3DTCR-E2E (FuXi, w/ MMD), 3DTCR-SFT (FuXi, w/o MMD), 3DTCR-SFT (FuXi, w/ MMD; full model), and the WRF 3-km reference, from top to bottom.}
    \label{fig:ampil_ablation_multivariable_reconstruction}
\end{figure}

The main text evaluates the effects of two-stage training and MMD-based latent space alignment using threshold-dependent CSI. Supplementary Fig.~\ref{fig:ampil_ablation_multivariable_reconstruction} provides the corresponding spatial comparison for Typhoon AMPIL. It compares the end-to-end model, 3DTCR-E2E (FuXi, w/ MMD); the fine-tuned model without MMD, 3DTCR-SFT (FuXi, w/o MMD); the full model, 3DTCR-SFT (FuXi, w/ MMD); and the WRF 3-km reference.

\FloatBarrier

\FloatBarrier
\subsection{Computational cost of vortex-following reconstruction}
\label{sec:Computational Cost Comparison and the Necessity of vortex-following reconstruction}
Generating large-scale, high-resolution datasets for TC research remains computationally demanding.
Supplementary Table~\ref{tab:computational_cost} summarizes the computational resources and runtime required for 5-day WRF integrations over the WNP or a TC-centered regional domain and for a 5-day 3DTCR reconstruction.

While traditional WRF simulations at moderate resolutions ($0.25^\circ$ and $0.125^\circ$) require manageable integration times (approximately 1 to 6 hours on 32 CPU cores), scaling the resolution to $0.027^\circ$ (approximately 3 km) across the entire basin drastically inflates the grid dimensions to $1852 \times 2592$. Consequently, the computation time exceeds 400 hours for a single 5-day forecast. This computational bottleneck makes it practically infeasible to construct a massive, high-resolution dataset using full-basin dynamical models.

To address this limitation, we focus reconstruction on the region of interest. Because the most destructive TC processes are concentrated within the inner-core region, applying a uniform 3-km grid across the entire WNP basin would cause substantial computational redundancy. Instead, 3DTCR reconstructs a compact $370 \times 370$ grid centered on the TC vortex. As shown in Supplementary Table~\ref{tab:computational_cost}, a full-basin 3-km WRF integration requires approximately 400 h on 32 CPU cores, a TC-centered WRF integration requires approximately 12 h on the same CPU configuration, and 3DTCR requires approximately 3 min on one NVIDIA A100 GPU. Because these values involve different domain sizes and hardware configurations, they compare resource requirements rather than runtime under identical conditions.

\begin{table}[!htp]
  \centering
  \small
  \caption{Computational resource requirements and runtime comparison
  for 5-day WNP and TC-centered tasks.
  All WRF simulations used the same Intel Xeon Platinum 8369B CPU
  platform at 2.90~GHz.}
  \label{tab:computational_cost}

  \begin{tabularx}{\linewidth}{
      p{0.19\linewidth}
      p{0.12\linewidth}
      p{0.16\linewidth}
      X
      p{0.12\linewidth}}
  \toprule
  \textbf{Method}
  & \textbf{Grid spacing}
  & \textbf{Grid size} $(N_{\mathrm{lat}} \times N_{\mathrm{lon}})$
  & \textbf{Computing resource}
  & \textbf{Runtime} \\
  \midrule

  WRF 25-km (WNP)
  & $0.250^\circ$
  & $201 \times 280$
  & 32 CPU cores
  & $\sim$1~h \\[2pt]

  WRF 12-km (WNP)
  & $0.125^\circ$
  & $401 \times 560$
  & 32 CPU cores
  & $\sim$6~h \\[2pt]

  WRF 3-km (WNP)
  & $0.027^\circ$
  & $1852 \times 2592$
  & 32 CPU cores
  & $\sim$400~h \\[2pt]

  WRF 3-km (TC vortex)
  & $0.027^\circ$
  & $370 \times 370$
  & 32 CPU cores
  & $\sim$12~h \\

  \midrule

  3DTCR (TC vortex)
  & $0.027^\circ$
  & $370 \times 370$
  & 1 NVIDIA A100 GPU
  & $\sim$3~min \\

  \bottomrule
  \end{tabularx}
  \end{table}

\clearpage
\renewcommand{\refname}{Supplementary References}
{
    \small
    \putbib[main]
}
\end{bibunit}

\end{document}